\documentclass[lettersize,journal]{IEEEtran}
\usepackage{amsmath,amsfonts}
\usepackage{algorithmic}
\usepackage{algorithm}
\usepackage{array}
\usepackage[caption=false,font=normalsize]{subfig}
\usepackage{textcomp}
\usepackage{stfloats}
\usepackage{url}
\usepackage{verbatim}
\usepackage{graphicx}
\usepackage{cite}
\usepackage[dvipsnames]{xcolor}

\usepackage[pagebackref,breaklinks,colorlinks,citecolor=blue]{hyperref}
\usepackage{hyperref}

\usepackage[normalem]{ulem}
\usepackage{graphicx} 
\usepackage[export]{adjustbox}
\usepackage{multicol}
\usepackage{pifont}
\usepackage{tabularx}
\usepackage{float} 
\usepackage{booktabs} 
\usepackage{enumitem}
\usepackage{multirow}
\usepackage{makecell, bigstrut}
\usepackage{adjustbox}
\usepackage{mdframed}
\usepackage{graphicx}
\usepackage{colortbl}
\usepackage{cancel}
\usepackage[normalem]{ulem}

\usepackage[utf8]{inputenc} 
\usepackage[T1]{fontenc}    
\usepackage{url}            
\usepackage{booktabs}       
\usepackage{amsfonts}       
\usepackage{nicefrac}       
\usepackage{microtype}      

\usepackage{booktabs}
\usepackage{makecell, multirow}
\usepackage{color}

\usepackage{upgreek}
\usepackage{pifont}
\newcommand{\cmark}{\ding{51}}%
\newcommand{\xmark}{\ding{55}}%

\usepackage{xcolor}
\usepackage{xr}

\newcolumntype{P}[1]{>{\centering\arraybackslash}p{#1}}
\newcolumntype{M}[1]{>{\centering\arraybackslash}m{#1}}

\usepackage{booktabs}
\usepackage{multibib}
\usepackage{caption}
\usepackage{multirow}
\newcolumntype{*}{>{\global\let\currentrowstyle\relax}}
\newcolumntype{^}{>{\currentrowstyle}}
\newlength\savewidth
\newcommand{\tablestyle}[2]{\setlength{\tabcolsep}{#1}\renewcommand{\arraystretch}{#2}\centering\footnotesize}

\makeatletter\renewcommand\paragraph{\@startsection{paragraph}{4}{\z@}
  {.5em \@plus1ex \@minus.2ex}{-.5em}{\normalfont\normalsize\bfseries}}\makeatother

\newcolumntype{x}[1]{>{\centering\arraybackslash}p{#1pt}}
\newcolumntype{y}[1]{>{\raggedright\arraybackslash}p{#1pt}}
\newcolumntype{z}[1]{>{\raggedleft\arraybackslash}p{#1pt}}

\newcommand{\ok}[1]{\textcolor{black}{#1}} 

\begin{document}


\title{{\color{black}Exploring the Potential of Multi-Modal AI \\for Driving Hazard Prediction}}


\author{
    Korawat Charoenpitaks\IEEEauthorrefmark{1},
    Van-Quang Nguyen\IEEEauthorrefmark{2},
    Masanori Suganuma\IEEEauthorrefmark{1},
    Masahiro Takahashi\IEEEauthorrefmark{3},
    Ryoma Niihara\IEEEauthorrefmark{3},
    Takayuki Okatani\IEEEauthorrefmark{1}\IEEEauthorrefmark{2}%

    \thanks{\IEEEauthorrefmark{1}Graduate School of Information Sciences, Tohoku University}
    \thanks{\IEEEauthorrefmark{2}RIKEN Center for AIP}
    \thanks{\IEEEauthorrefmark{3}DENSO CORPORATION}
}





\markboth{IEEE TRANSACTIONS ON INTELLIGENT VEHICLES, VOL. , NO. , MONTH 2024}%
{Shell \MakeLowercase{\textit{et al.}}:}


\maketitle
\begin{abstract}
This paper addresses the problem of predicting hazards that drivers may encounter while driving a car. We formulate it as a task of anticipating impending accidents using a single input image captured by car dashcams. Unlike existing approaches to driving hazard prediction that rely on computational simulations or anomaly detection from videos, this study focuses on high-level inference from static images. The problem needs predicting and reasoning about future events based on uncertain observations, which falls under visual abductive reasoning. To enable research in this understudied area, a new dataset named the DHPR (Driving Hazard Prediction and Reasoning) dataset is created. The dataset consists of 15K dashcam images of street scenes, and each image is associated with a tuple containing car speed, a hypothesized hazard description, and visual entities present in the scene. These are annotated by human annotators, who identify risky scenes and provide descriptions of potential accidents that could occur a few seconds later. We present several baseline methods and evaluate their performance on our dataset, identifying remaining issues and discussing future directions. This study contributes to the field by introducing a novel problem formulation and dataset, enabling researchers to explore the potential of multi-modal AI for driving hazard prediction.
\end{abstract}

\begin{IEEEkeywords}
Vision, Language, Reasoning, Traffic Accident Anticipation.
\end{IEEEkeywords}

\section{Introduction}
\begin{figure}[t]
\centering
\includegraphics[width=0.9\linewidth]{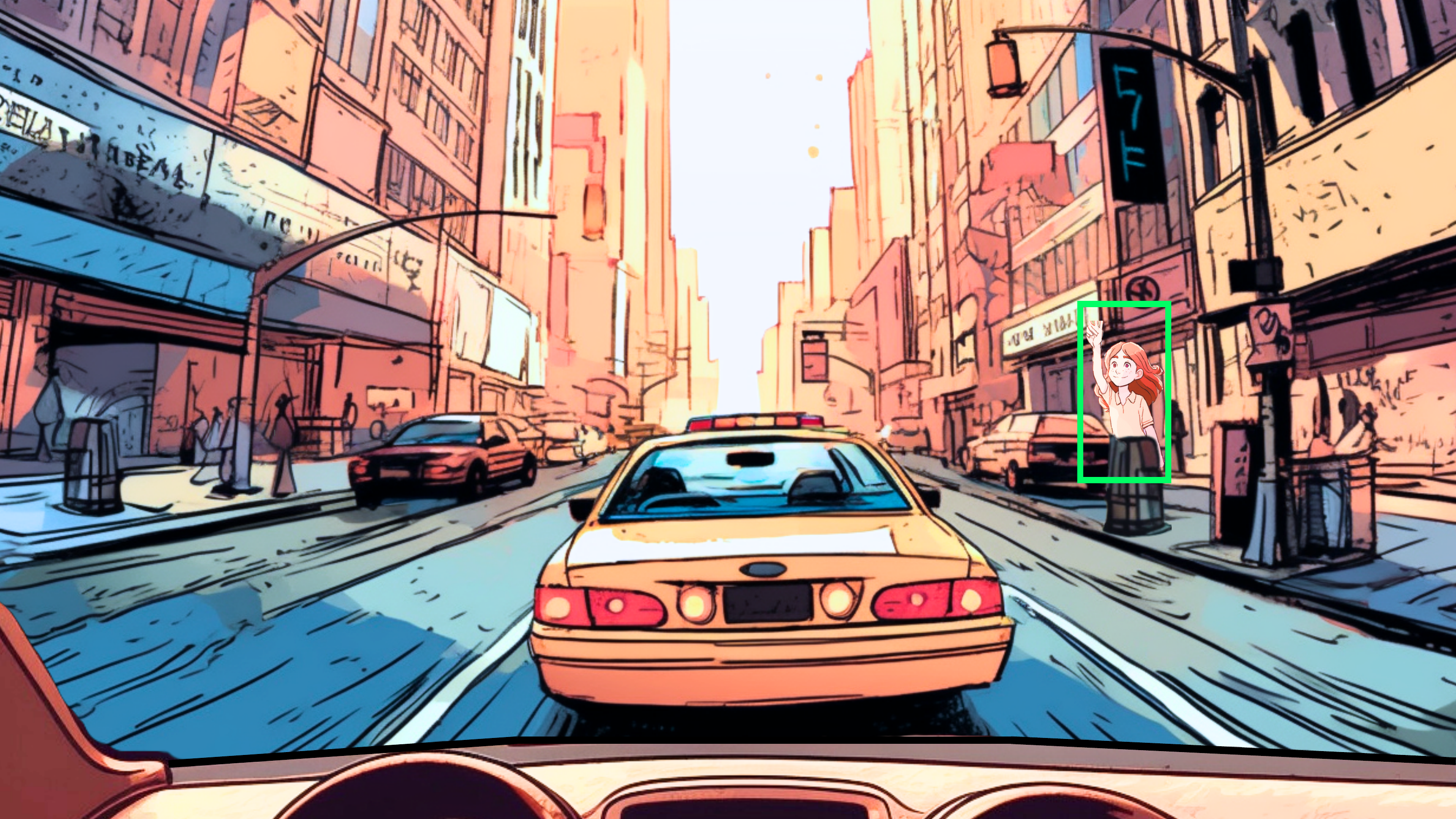}
\caption[Caption for LOF]{Example of driving hazard prediction from a single dashcam image. The pedestrian in the green box may be attempting to flag down a taxi, and the taxi may abruptly stop in front of our car to offer them a ride.}
\vspace*{-3mm}
\label{fig:anticipate_an_impending_accident}
\end{figure}

In this paper, we consider the problem of predicting future hazards that drivers may encounter while driving a car. Specifically, we approach the problem by formulating it as a task of anticipating an impending accident using a single input image of the scene in front of the car. An example input image is shown in Fig.~\ref{fig:anticipate_an_impending_accident},
which shows a taxi driving in front of the car on the same lane, and a pedestrian signalling with their hand. From this image, one possible reason is that the pedestrian may be attempting to flag down the taxi, which could then abruptly halt to offer them a ride. In this scenario, our car behind the taxi may not be able to stop in time, resulting in a collision. This simple example shows that predicting hazards sometimes requires abductive and logical reasoning.

\begin{figure*}[t]
\centering
\includegraphics[width=1.0\linewidth]{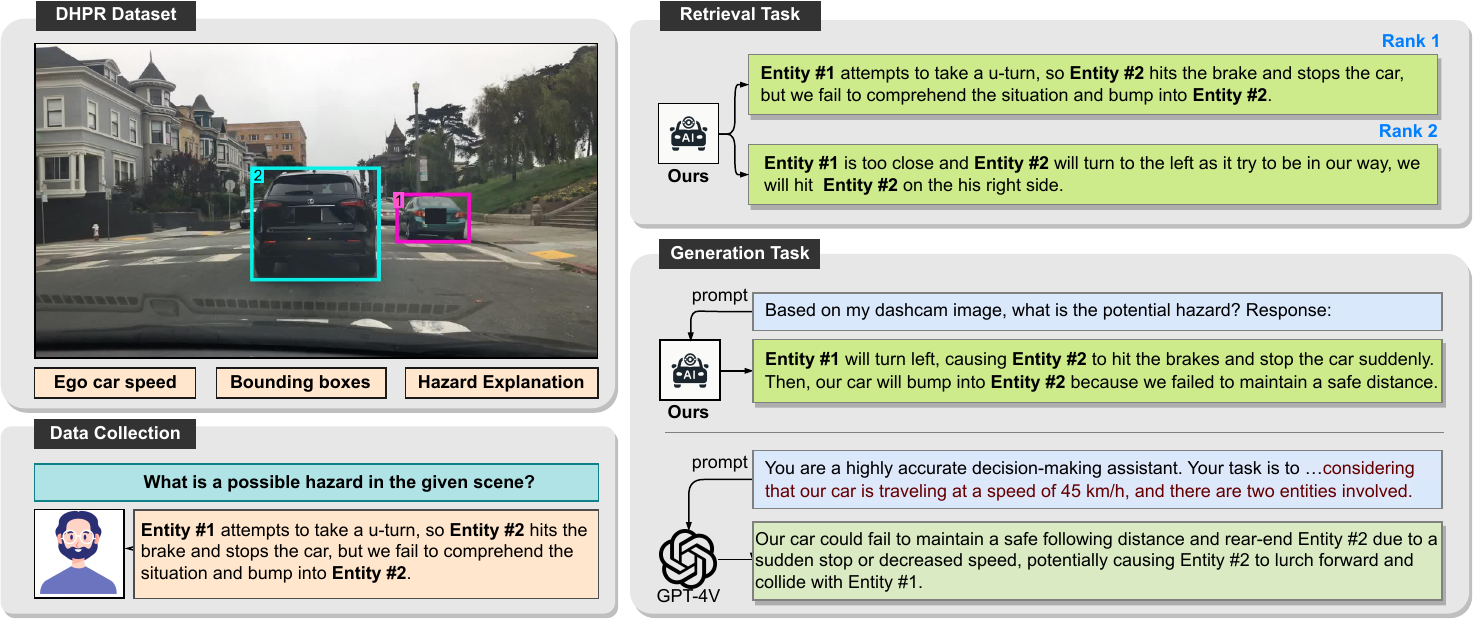}
\caption{ 
Illustration of the DHPR dataset with example annotations (left) and hazard explanations retrieved and generated by our model and GPT-4V (right).}
\label{fig:overview-dataset}
\end{figure*}

Thus, our approach formulates the problem as visual abductive reasoning (VAR) \cite{hessel2022abduction, liang2022visual}
from a single image. As an underlying thought, we are interested in leveraging recent advances in multi-modal AI, such as visual language models (VLMs) \cite{alayrac2022flamingo, li2023blip2, zhu2023minigpt4, dai2023instructblip, liu2023visual, luo2023cheap}. Despite the growing interest in self-driving and driver assistance systems, little attention has been paid to the solution we consider here, to the best of our knowledge.
Existing approaches rely on predicting accidents through computational simulations using physics-based \cite{path_finding_paper} or machine-learning-based models of the environment \cite{xu2022safebench}. For instance, they predict the trajectories of pedestrians and other vehicles. Another approach formulates the problem as detecting anomalies from input videos \cite{yao2022dota, yao2019unsupervised}. However, these methods, which rely only on a low-level understanding of scenes, may have limitations in {\em predicting future events that occur over a relatively long time span}, as demonstrated in the example above. {\color{black} If we expand our focus beyond self-driving and driver assistance systems, there are several studies \cite{deeplearning_paper1, deeplearning_paper2} employing deep learning to predict accidents and analyze factors contributing to accidents within the broader context of traffic systems. However, these studies are beyond the scope of this research.}

{\color{black} The present study is among the first to apply multimodal AI, integrating vision and language, to the prediction of driving hazards {\color{black} in the ego-car context.} Our study seeks to realize in vehicle intelligence the complex reasoning and prediction capabilities that human drivers perform when driving. This line of research is important because human drivers naturally perform complex reasoning tasks, such as understanding the intentions of other road users, including drivers and pedestrians. Current systems in autonomous vehicles and ADAS (advanced driver-assitance system), which mostly rely on simulation-based methods, do not possess these reasoning capabilities. We argue that further research in this area could substantially improve how autonomous driving technologies function, addressing their present limitations. 
It should be noted that our approach does not seek to replace existing simulation-based methods. Instead, it will naturally enhance them by integrating human-like abductive reasoning into vehicle intelligence. 

The advantage of VAR over simulation is its capability for extended forecasting. As the prediction horizon extends, the complexity of forecasting increases with the exponential growth in potential hazards. VAR stands out by integrating logical reasoning into forecasts, aiding in prioritizing events according to their likelihood. Specifically, it evaluates the behaviors of various traffic participants, such as pedestrians and drivers, to anticipate potential incidents. In contrast, simulation-based approach often finds it challenging to handle such predictive tasks efficiently. Figure \ref{fig:anticipate_an_impending_accident} demonstrates the difficulties in predicting a taxi's stopping behavior without specialized algorithms.
}

An important note is that 
{\color{black} in the present study, we use} a single image as input, which may seem less optimal than using a video to predict hazards encountered while driving. There are two reasons {\color{black} simplifying the problem} for our choice. First, human drivers are capable of making accurate judgments even from a static scene image, as demonstrated in the example above. {\color{black} Our study considers this particular type of hazard}. Humans are apparently good at anticipating the types of hazards that may occur and further estimating the likelihood of each one. Second, there are technical challenges involved in dealing with video inputs. Unlike visual inference from a static image (e.g., visual question answering \cite{agrawal2016vqa}), there is currently no established approach in computer vision for performing high-level inference from dynamic scene videos; \ok{see \cite{liang2022visual, hessel2022abduction}} for the current state-of-the-art. While videos contain more information than single images, we believe that there remains much room to explore in using single-image inputs. 

To investigate this understudied approach to driving risk assessment, we present a formulation of the problem and create a dataset for it. Since actual car accidents are infrequent, it is hard to collect a large number of images or videos of real accidents. To cope with this, we utilize existing datasets of accident-free images captured by dashcams, specifically BDD100K (Berkeley DeepDrive) \cite{bdd100k} and ECP (EuroCity Persons) \cite{Braun_2019}; they were originally created for different tasks, e.g., object detection and segmentation. From these datasets, we have human annotators first identify scenes that potentially pose risks, in which an accident could occur a few seconds later. We then ask them to provide descriptions of the hypothesized accidents with mentions of entities (e.g., traffic signs, pedestrians, other cars, etc.) in the scene. 

The proposed dataset, named DHPR (Driving Hazard Prediction and Reasoning), 
is summarized as follows. 
It contains 15K scene images, for each of which a tuple of a car speed, a description of a hypothesized hazard, and visual entities appearing in the image are provided; see Fig.~\ref{fig:overview-dataset}. There are at least one and up to three entities in each scene, each represented by a bounding box with its description. Each entity is referred to as `Entity \#$n$' with $n(=1,2,3)$ in the hazard description. 

{\color{black} Despite certain limitations, DHPR can contribute to the development of AI systems capable of predicting potential hazards in traffic scenes, similar to human drivers. The first limitation is that DHPR only includes single images and vehicle speed as inputs, utilizing a limited portion of the information available in real driving scenarios. The second limitation is that the annotated hazards are hypothesized rather than real. However, these limitations do not undermine the utility of DHPR for two main reasons. First, it is curated by highly skilled annotators, as will be explained later, ensuring a rich collection of realistic hypothesized hazards. Second, although the hazards are restricted to those identifiable from still images and given vehicle speed, the diversity and frequency of these hazards are substantial.}

Based on the dataset, {\color{black}we examine the task of inferring driving hazards using traffic scene images}. This task involves making inferences based on uncertain observations and falls under the category of visual abductive reasoning, which has been the subject of several existing studies, \cite{hessel2022abduction, liang2022visual}. These studies have also introduced 
datasets, such as Sherlock \cite{hessel2022abduction} and VAR \cite{liang2022visual}. 
However, our study differs in several key ways, detailed in Table \ref{tab:datasetcomparison}. While our study specifically targets traffic scenes, we tackle a broader visual reasoning challenge within this context. This involves recognizing multiple objects, understanding their interactions, and engaging in multi-step reasoning. \ok{Aiming to deal with these challenges, we introduce a novel method tailored to this task, which emphasizes efficiently extracting relevant information from both input images and texts.}

It is also worth mentioning that numerous studies on traffic accident anticipation have been conducted, building datasets with similar dashcam imagery (e.g., \cite{yao2019unsupervised,Bao_2020,yao2022dota,DRAMA}). 
However, these studies only provide annotations for closed-set classes of accidents/causation or address relatively uncomplicated traffic scenarios involving singular objects in inference. 
In contrast, our study includes annotations for open-set driving hazards expressed in natural language texts and focuses on more complicated visual reasoning in forecasting potential traffic hazards.



The following section provides a more detailed discussion of related work (Sec.~\ref{sec2}). We then proceed to explain the process of creating the dataset (Sec.~\ref{sec3}). Next, we explore various task designs that can be examined using this dataset (Sec.~\ref{sec4}). The experimental results, which evaluate the performance of \ok{ the proposed method and existing methods for general vision and language tasks in predicting driving hazards,} are presented in Sec.~\ref{sec5}. Finally, we conclude our study in Sec.~\ref{sec6}.

\begin{table*}[t] 
    \centering
        \caption{Comparison of DHPR with existing datasets. {\color{black} A checkmark (\cmark) indicates the presence of the feature, and a crossmark (\xmark) indicates the feature’s absence}
        }    
    \footnotesize
    \begin{adjustbox}{max width=\textwidth}
    \renewcommand{\arraystretch}{1.25}
    \begin{tabular}{
    >{\centering\arraybackslash}M{16mm}
    >{\centering\arraybackslash}M{16mm}
    >{\centering\arraybackslash}M{30mm}
    >{\centering\arraybackslash}M{10mm}
    >{\centering\arraybackslash}M{14mm}
    >{\centering\arraybackslash}M{24mm}
    >{\centering\arraybackslash}M{24mm}
    }
      \toprule
      \textbf{Dataset} & \textbf{Visual \break Inputs} & \textbf{Task} & \textbf{Multiple objects} & \textbf{Multi-step reasoning} & 
      \textbf{Object \break Relationship} & \textbf{Annotation \break Type} 
      \\
      \midrule
      Sherlock \break \cite{hessel2022abduction} & Scene \break images & Abductive reasoning of an interested object & \xmark & \xmark & None 
      & Natural language 
      \\
      
      VAR \break \cite{liang2022visual} & Scene \break images & Abductive reasoning of a missing event & \xmark & \cmark & Event relations  & Natural language 
      \\

      CCD \break \cite{Bao_2020} & Dash-cam videos & Classification \break of a future event & \cmark & \xmark & Trajectory only & Closed-set of classes
      \\

      DoTA \break \cite{yao2022dota} & Dash-cam videos & Classification \break of a future event & \cmark & \xmark & Trajectory only & Closed-set of classes
      \\

      CTA \break \cite{CTA} & Dash-cam videos & Classification \& localization of  accident & \cmark & \cmark &  Event relations & Closed-set of classes 
      \\ 

      TrafficQA \break \cite{TrafficQA} & Videos & Event recognition \& reasoning & \cmark & \cmark & Trajectory only & Natural language 
      \\ 

      DRAMA \break \cite{DRAMA} & Dash-cam videos & Risk localization \& explanation & \xmark & \cmark &  None & Natural language 
      \\ 

      Rank2Tell \break \cite{rank2tell} & Videos \& LiDAR & Identification of objects that affect driving & \cmark & \cmark &  None & Natural language 
      \\ 

      BDD-X \break \cite{BDD-X} & Dash-cam videos & Introspective \break explanation & \xmark & \xmark &  None & Natural language 
      \\ 

      BDD-OIA\break \cite{BDD-OIA} & Dash-cam images & Action reasoning explanation & \cmark & \cmark &  None & Closed-set of classes 
      \\ 

      NuScene-QA \break \cite{nuscene-qa} & Images \& LiDAR & VQA for traffic understanding & \cmark & \cmark &  Spatial position only & Natural language
      \\ 
        
      Talk2Car\break \cite{talk2car} & Dash-cam videos & Object referral in \break natural language & \cmark & \xmark &  None & Natural language 
      \\ 
      
      \rowcolor{gray!10} 
      \textbf{Ours} (\textbf{DHPR}) & Dash-cam images & Abductive reasoning of a future hazard & \cmark & \cmark & Object interactions  & Natural language 
      \\
      \bottomrule
    \end{tabular}
    \renewcommand{\arraystretch}{1.0}
    \end{adjustbox}
    \label{tab:datasetcomparison}
\end{table*}

\section{Related Work} \label{sec2}

\subsection{Reasoning in Traffic Scenes}
Traffic accident anticipation has received significant attention in the fields. We focus here exclusively on studies that utilize a dashboard camera as the primary input source. 

The majority of these studies employ video footage as input and formulate the problem as video anomaly detection, where methods predict the likelihood of an accident occurring within a short time frame based on the input video. While some studies consider supervised settings \cite{Chan2016AnticipatingAI,kataoka2018drive,suzuki2018anticipating,Bao_2020}, the majority consider unsupervised settings, considering the diversity of accidents. Typically, moving objects are first detected in input videos, such as other vehicles, motorbikes, pedestrians, etc., and then their trajectories or future locations are predicted to identify anomalous events; more recent studies focus on modeling of object interactions \cite{herzig2019spatiotemporal,fatima2020globalfeatureaggregation,karim2021adynamicspatial,yao2022dota}. Some studies consider different problem formulations and/or tasks, such as predicting driver's attention in accident scenarios~\cite{fang2023dada}, learning accident anticipation and attention using reinforcement learning~\cite{bao2021drive}, and understanding traffic scenes from multi-sensory inputs by the use of heterogeneous graphs representing entities and their relation in the scene~\cite{Monninger_2023}.

Meanwhile, to establish trust in autonomous driving and driver assistance systems, it is crucial that these systems offer explanations of their reasoning in a format that humans can understand, specifically through natural language explanations. Building on this premise, recent studies have approached the task by framing it as a challenge of image/video question answering~\cite{TrafficQA,nuscene-qa} or generating descriptions about the ego vehicle’s actions~\cite{BDD-X,BDD-OIA} or the objects that may affect ego vehicle’s driving~\cite{DRAMA,rank2tell}, accompanied by relevant datasets.

Our study aligns with the aims of these prior works but places a stronger emphasis on the prediction of future potential hazards. This requires the identification and interpretation of multiple objects and their interactions within a traffic scenario. Although some studies~\cite{TrafficQA,rank2tell} have tackled the prediction of potential risks, there has been a lack of focus on the interactions of multiple objects. Moreover, our dataset stands out by providing comprehensive explanations of potential hazards, which have been extensively annotated to encompass a wide range of accident scenarios, including the underlying reasons for these accidents and their possible outcomes. The comparisons between our dataset and those of previous works are outlined in Table~\ref{tab:datasetcomparison}.

\subsection{Visual Abductive Reasoning}

Abductive reasoning, which involves inferring the most plausible explanation based on partial observations, initially gained attention in the field of NLP \cite{hessel2022abduction, liang2022visual, kadikis-etal-2022-embarrassingly, young-etal-2022-abductionrules}. While language models (LMs) are typically adopted for the task, some studies incorporate relative past or future information as context to cope with the limitation of LMs that are conditioned only on past context \cite{qin-etal-2020-back}. Other researchers have explored ways to enhance abductive reasoning by leveraging additional information. For example, extra event knowledge graphs have been utilized \cite{Du2021LearningEG} for reasoning that requires commonsense or general knowledge, or general knowledge and additional observations are employed to correct invalid abductive reasoning \cite{paul-frank-2021-generating}. However, the performance of abductive reasoning using language models exhibits significant underperformance, particularly in spatial categories such as determining the spatial location of agents and objects \cite{bhagavatula2020abductive}.

Visual abductive reasoning extends the above text-based task to infer a plausible explanation of a scene or events within it based on the scene's image(s). This expansion goes beyond mere visual recognition and enters the realm of the ``beyond visual recognition'' paradigm. 
The machine's ability to perform visual abductive reasoning is tested in general visual scenarios. In a recent study, the task involves captioning and inferring the hypothesis that best explains the visual premise, given an incomplete set of sequential visual events \cite{liang2022visual}.  Another study formulates the problem as {\color{black} identifying visual clues in an image} 
to draw the most plausible inference based on knowledge \cite{hessel2022abduction}. To handle inferences that go beyond the scene itself, the authors employ CLIP, a multi-modal model pre-trained on a large number of image-caption pairs \cite{radford2021learning}. 

\section{The Driving Hazard Prediction and Reasoning (DHPR) Dataset} \label{sec3}

\subsection{Specifications}

The DHPR dataset\footnote{The dataset will be made public upon acceptance.} provides annotations to 14,975 scene images captured by dashcams inside cars running on city streets, sourced from BDD100K (Berkeley Deepdrive) and ECP (EuroCity Persons).
Each image $x$ is annotated with 
\begin{itemize}
    \item Speed $v$: a hypothesized speed $v(\in\mathbb{R})$ of the car
    \item Entities $\{e_n=(e_{\mathrm{bbox},n},e_{\mathrm{desc},n})\}_{n=1,\ldots,N}$: up to three entities $(1\leq N \leq 3)$ leading to a hypothesized hazard, each annotated with a bounding box $e_{\mathrm{bbox},n}$ and a description $e_{\mathrm{desc},n}$ (e.g., `green car on the right side of the road')
    \item Hazard explanation $h$: a natural language explanation $h$ of the hypothesized hazard and how it will happen by utilizing the entities $\{e_n\}_{n=1,\ldots,N}$ involved in the hazard; each entity appears in the format of `Entity \#$n$' with index $n$. 
\end{itemize}
The 14,975 images are divided into train/validation/test splits of 12,975/1,000/1,000, respectively. 

\subsection{Annotation Process} 

{\color{black}We employ Amazon Mechanical Turk (MTurk) to collect the aforementioned annotations \cite{amt}}. To ensure the acquisition of high-quality annotations, we administer an exam resembling the main task to identify competent workers and only qualified individuals are invited to participate in the subsequent annotation process. We employ the following multi-step process to select and annotate images from the two datasets, BDD100K and ECP. Each step is executed independently; generally, different workers perform each step on each image; see the supplementary material for more details.

In the first step, we ask MTurk workers to select images that will be utilized in the subsequent stages, excluding those that are clearly devoid of any hazards. This leads to the choice of 25,000 images from BDD100K and 29,358 images from ECP. For each image, the workers also select the most plausible car speed from the predefined set [10, 30, 50+] (km/h) that corresponds to the given input image. 

In the second step, we assign workers to evaluate if the car could be involved in an accident within a few seconds, assuming it travels at 1.5 times the speed noted in the first step. This speed increase from the initially annotated one is because the original images were captured under normal driving conditions, with no accidents occurring subsequently. Our aim is that this adjustment will aid workers in more effectively assessing accident risks and formulating realistic scenarios. Images judged as safe are excluded, reducing the total count from 54,358 to 20,791.


In the third step, we assign workers to annotate the remaining images by identifying potential hazards involving up to three entities. They draw bounding boxes, describe each entity, and explain the hazard, ensuring the hazard explanation is at least five words and references all entities as `Entity \#$n$'. See examples in Fig.~\ref{fig:overview-dataset}.

Finally, we conduct an additional screening to enhance the quality of the annotations. In this step, we enlist the most qualified workers to evaluate the plausibility of the hazard explanations in each data sample. This process reduces the number of samples from 20,791 to 14,975.

\subsection{Hazard Types} 
\begin{table}[t]
    \caption{Hazard types and their statistics.
    `Retrieval' represents the candidates sampled for retrieval tasks.} 
    \centering

    \resizebox{0.48\textwidth}{!}{%
    \begin{tabular}{
            >{\arraybackslash}p{0.4\columnwidth} 
            >{\centering\arraybackslash}p{0.08\columnwidth}
            >{\centering\arraybackslash}p{0.14\columnwidth}
            >{\centering\arraybackslash}p{0.08\columnwidth}
            >{\centering\arraybackslash}p{0.14\columnwidth}
            }
        \toprule
            \multicolumn{1}{l}{\multirow{2}{*}{\textbf{Hazard Type}}} & \multicolumn{2}{c}{\textbf{Validation Set}} & \multicolumn{2}{c}{\textbf{Test Set}} \\ 
            & \small{Total} & 
            \small{Retrieval} & \small{Total} & 
            \small{Retrieval} 
            \\
        \midrule
            \textbf{Entity Related:} & & & & \\
            \hspace{1mm} Pedestrian & 104 & 10 & 84 & 10  \\
            \hspace{1mm} Stationary Object & 15 & 10 & 61 & 10  \\
            \hspace{1mm} Traffic Signal & 46 & 10 & 41 & 10  \\
            \hspace{1mm} Unusual Condition & 20 & 10 & 16 & 10  \\
        \midrule
            \textbf{Driving Scenario:} & & & & \\
            \hspace{1mm} Speeding \& Braking & 473 & 20 & 413 & 20  \\
            \hspace{1mm} Sideswipe & 52 & 10 & 48 & 10  \\
            \hspace{1mm} Merging Maneuver & 264 & 20 & 305 & 20  \\
            \hspace{1mm} Unexpected Event & 20 & 5 & 23 & 5  \\
            \hspace{1mm} Chain Reaction & 6 & 5 & 9 & 5  \\
        \midrule
            \textbf{Total} & 1000 & 100 & 1000 & 100 \\
        \bottomrule
    \end{tabular}
    }
    \label{tab:hazard_candidates}
\end{table}

To facilitate proper evaluation and detailed analysis of reasoning models, we categorized the hazards annotated as above into multiple types; see Table \ref{tab:hazard_candidates}. Initially, these hazards were divided into two groups based on the entities involved. The first group encompasses hazards related to other vehicles on the road, i.e., cars, buses, and bikes. The second group consists of hazards related to entities other than these vehicles. The latter group, termed ``Entity Related,'' is smaller and further subdivided into four types based on the entity involved, i.e., {\em Pedestrian, Stationary Object, Traffic Indication} (e.g., traffic signal), and {\em Driving Condition.} (i.e., unusual conditions such as sun glare or inclement weather). The former group, labeled as ``Driving Scenario,'' is more numerous and is categorized into the following five types according to the scenario of the hazard: {\em Speeding and Braking}, which frequently result in rear-end collisions; {\em Sideswipe} incidents stemming from improper distance estimation; {\em Merging Maneuvers} that lead to conflicts or misunderstandings about right-of-way; {\em Unexpected Events}, such as sudden stops during turns or mid-motion changes of intent; and {\em Chain Reactions}, where multiple vehicles contribute to a series of accidents. Table \ref{tab:hazard_candidates} shows the statistics for these hazard types in the validation and test subsets. It also displays the number of selected samples for retrieval tasks, which will be explained later.

\section{Task Design and Evaluation}  \label{sec4}
\subsection{Task Definition} \label{sec:taskdef}
Our dataset supports a variety of tasks with varying levels of difficulty. Each sample in our dataset consists of $(x, v, h, \{e_1,\ldots,e_N\})$, where $x$ is an input image, $v$ is the car's speed, $h$ is a hypothesized hazard explanation, and $e_n=(e_{\mathrm{bbox},n},e_{\mathrm{desc},n})$ are the entities involved in the hazard. 

In this study, we consider two main tasks: image/text retrieval and text generation, aiming to conduct a comprehensive evaluation to understand the core aspects of the problem better. The image/text retrieval task involves ranking a set of candidate images or texts based on their relevance to given input texts or images. The text generation task entails creating a natural language explanation $h$ that corresponds to a specific input image $x$. Each task presents its own advantages and challenges, including varying levels of task difficulty and complexity in evaluation.

In addition, we must also consider how to handle visual entities, as different methods impact task difficulty. The most challenging method requires models to autonomously detect and identify entities. A simpler alternative is to provide models with the correct entities already marked as boxes in the input image. We opt for the latter method to minimize ambiguity in hazard prediction. In any given scene, multiple potential hazards may be hypothesized. Specifying the entities helps to narrow down the options available for the models to evaluate.

We address two types of retrieval tasks: image-to-text retrieval (TR) and text-to-image retrieval (IR). In TR, we rank each hazard explanation $h$ from a set $S_h$ based on its relevance to an input image $x$ and its auxiliary data. As explained above, the boxes of involved entities $E=\{e_{\mathrm{bbox},1},\ldots,e_{\mathrm{bbox},N}\}$ are assumed to be provided for each image $x$. The models then compute a score for each $h$ using these inputs as
$s = s(h, x, E)$.
In IR, we rank each pair of an image and its associated boxes $(x,E)$ from a set of candidate image-box pairs $S_{x,E}$ using the same score.





In the generation task, models directly generate a hazard explanation $h$ as natural language text from a given input image $x$ and its associated entity boxes $E$. Typically, these models use large language models (LLMs), which generate the text autoregressive.

\subsection{Evaluation Procedure and Metrics}


In the retrieval tasks, we evaluate the models' performance using a subset of 100 selected samples for each of the validation and test subsets, as detailed in Table \ref{tab:hazard_candidates}. Acknowledging the original data's imbalance in hazard types, we ensure a balanced distribution of these types for a more meaningful evaluation. For TR, we rank the 100 texts (hazard explanations) from them for each image-box pair within the same 100 samples. For IR, we perform a similar ranking but with the roles of the query and the database reversed. To assess the quality of these rankings, we use two metrics: the average rank of the correct entry and Recall@$k$, which measures the frequency of the correct entry appearing within the top $k$ positions. 

In the generation task, we utilize four standard metrics commonly used in image captioning: BLEU-4 \cite{papineni2002bleu}, ROUGE \cite{lin2004rouge}, CIDEr \cite{vedantam2015cider}, and SPIDEr \cite{liu2016optimization}. These metrics assess the similarity between the generated text and its ground truth, which is the annotated hazard explanation. However, they primarily measure formal similarity across various aspects, rather than semantic similarity. To address this, we incorporate GPT-4 \cite{GPT4} to compute a semantic similarity score, which ranges from 0 to 100. We craft a specific prompt for GPT-4 to align the output score closely with our understanding of semantic accuracy. This method assesses spatial and causal relationships and the accuracy of entity references. It is also designed not to penalize the generated text for including additional content that is not present in the ground truth. More detailed information is provided in the supplementary material.
Unlike the retrieval tasks, we use all the samples in the validation/test sets.

\begin{figure*}[t]
\centering
\includegraphics[width=0.85\linewidth]{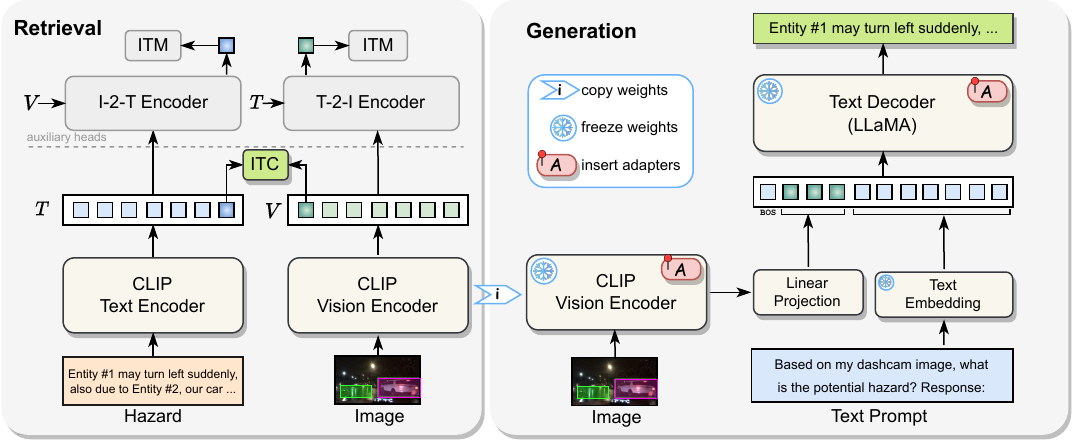}
\caption{The proposed method for the retrieval and generation tasks.}
\label{fig:model}
\end{figure*}

\section{Experiments} \label{sec5}
\subsection{Methods}
\paragraph{How to Input Visual Entities?}

As explained in Sec.~\ref{sec:taskdef}, we assume that the entities involved in hazard prediction are given for each input image $x$; they are given in the form of bounding boxes $E=\{e_{\mathrm{bbox},1},\ldots,e_{\mathrm{bbox},N}\}$. Given that multiple entities can exist within a single image, they are identified in the hazard explanation as `Entity \#$1$’, `Entity \#$2$’, etc. Then, inference models, whether for retrieval or generation, must discern the relationship between these textual references and their corresponding bounding boxes. 

To enable this, we employ an approach to augment the input image $x$ with color-coded bounding boxes, following \cite{hessel2022abduction, yao2022cpt}. Specifically, an opaque color is used to represent a bounding box. As there are up to three entities, we employ a simple color-coding scheme, i.e., using purple, green, and yellow colors to indicate Entity \#$1$, $2$, and $3$, respectively. We employ alpha blending (with 60\% opaqueness) between boxes filled with the above colors and the original image; see the bottom of Fig.~\ref{fig:model} for examples. We will use $\tilde{x}$ to indicate the augmented image in what follows. In this setup, models will learn the color coding scheme through training using the DHPR training set.

\ok{Note that a different method is applied when using GPT4V for zero-shot generation, as will be explained later. While similar color codes are used, the bounding boxes are unfilled instead of filled. No learning process is involved in this case; the colors of the boxes are directly specified in text prompts \cite{SoM}.}

\paragraph{Proposed Method}
\ok{
We introduce a new method designed to serve as a baseline for our task. This method is based on CLIP \cite{radford2021learning}, a well-established method for image and text retrieval in general contexts. CLIP is also applied to VAR \cite{hessel2022abduction}. However, our task, DHPR, demands advanced capabilities, including the recognition of multiple objects (via the color-coding scheme) and identifying their spatial relationship. Thus, our method focuses on enhancing the ability to extract pertinent information from both text and images. 
}


Figure \ref{fig:model} shows the architecture of our model. In CLIP, the input text and image do not interact until the computation of the cosine similarity at the final stage. Text and image embeddings are independently calculated by two separate encoders. Aiming for a higher-level integration of both, we introduce two additional Transformer-based encoders. The first is a text-to-image encoder, designed to attend to the image embeddings from CLIP with the text embeddings from the same, extracting more relevant information. Similarly, we use an inverse image-to-text encoder. 
For the updated text and image embeddings, we compute two image-text matching (ITM) losses, 
{\color{black}each requiring an ITM head on their respective encoders: one for the image-to-text (I-2-T) encoder and one for the text-to-image (T-2-I) encoder}. These losses are combined with the original contrastive loss of CLIP and used for training. 


The retrieval score continues to use the cosine similarity between the embeddings from the CLIP's encoders. We use
the pre-trained CLIP ViT-L/14 for the visual encoder and BERT-base for the text encoder. The added two encoders have a simple design, each comprising two standard Transformer layers.

For the generation task, the vision encoder trained as above is repurposed and integrated with a pre-trained LLM to construct a visual language model (VLM). We employ LLaMA2 7B. Recent studies predominantly employ a blend of a pre-trained LLM and a vision encoder, specifically aligning the output of the vision encoder with the textual input of the LLM. While there exist variations in this methodology, we adopted the strategy outlined in \cite{luo2023cheap}. 
Specifically, we employ multi-modality adapters \cite{luo2023cheap} injected into the Transformer layers of the vision encoder and the LLM. 
We extract three [CLS] embeddings after every eight Transformer layers of the vision encoder and project them into the space of the LLM's input tokens. \ok{A notable difference from \cite{luo2023cheap} is that we use the [BOS] embedding as a routing token.} We append the projected tokens with the tokens of a text prompt, as illustrated in Fig.~\ref{fig:model}. We then train the adapters while freezing the CLIP vision encoder and the LLM for efficient training and to mitigate overfitting. 


\paragraph{Compared Methods}

In addition to the above method, we evaluate several existing methods. The first is the original CLIP, fine-tuned on DHPR, for the retrieval tasks. The cosine similarity between the input image and text provides their relevance score.
Additionally, we evaluate two popular VLMs,
BLIP \cite{li2022blip} and BLIP2 \cite{li2023blip2}, for both retrieval and generation tasks. BLIP employs separate transformers, a ViT for extracting image embeddings and a BERT Transformer for text embeddings. Besides ViT, BLIP2 uses an additional Q-Former for extracting image embeddings. For the retrieval task, we calculate the relevance score using cosine similarity between image and text embeddings from these models. For the generation task, BLIP employs classic BERT as a text decoder, and BLIP2 uses a LLM, OPT 6.7B, as a text decoder.

\ok{
For the generation task, we also evaluate a SOTA VLM,  LLaVA-1.5 \cite{liu2023visual}, which utilizes LLaMA 7B as the underlying LLM. Furthermore, we evaluate GPT-4V's capability to generate hazard explanations in a zero-shot setting. Distinctively, GPT-4V can accurately relate each bounding box in images with input/generated texts, requiring only instructions like `Entity \#$1$ is highlighted by the magenta box' in the prompt, without prior training on the DHPR dataset. To test its maximum performance in a zero-shot scenario, we include the ego car's speed in the prompt for GPT-4V, a piece of information not available to the other models. {\color{black} We utilized a GPT-4V model, gpt-4-vision-preview, through OpenAI API.} Details about the prompt design are available in the supplementary material.
}

\begin{table*}[t]
\begin{minipage}[t]{.48\textwidth}
\centering
\caption{Results for the image retrieval (IR) and text retrieval (TR) tasks on the DHPR test split. The retrieval tasks are evaluated by the average rank and Recall@1. 
For all metrics except the rank metric, higher values indicate better performance.}
\centering
\begin{tabular}{
    >{\raggedright\arraybackslash}M{22mm} 
    >{\centering\arraybackslash}M{11mm}
    >{\centering\arraybackslash}M{11mm}
    >{\centering\arraybackslash}M{11mm}
    >{\centering\arraybackslash}M{11mm}
    }
\toprule
    \multicolumn{1}{c}{\multirow{2}{*}{\textbf{Model}}} &
    \multicolumn{2}{c}{\textbf{IR Task}} &
    \multicolumn{2}{c}{\textbf{TR Task}} 
    \\
    & 
    \textbf{Rank} $\downarrow$ & 
    \textbf{R@1} $\uparrow$ & 
    \textbf{Rank} $\downarrow$ & 
    \textbf{R@1} $\uparrow$ 
    \\
\midrule
    CLIP-ViT-L/14 & 10.8  & 24.1\% & 10.9 & 24.8\% \\
    BLIP-base  & 15.3 & 9.3\% &  15.9  & 8.1\% \\
    BLIP2-OPT-6.7B & 11.5 & 19.1\% & 12.1  & 19.8\% \\
    \rowcolor{gray!10}Ours & \textbf{10.2} & \textbf{24.9\%} & \textbf{10.3} & \textbf{26.3\%} \\
\bottomrule
\label{tab:test_result_retrieval}
\end{tabular}
\end{minipage}%
\hfill
\begin{minipage}[t]{.48\textwidth}
\centering
\caption{Results for the generation task on the DHPR test split, evaluated using BLEU (B4), ROUGE (R), CIDEr (C), SPIDER (S), and the GPT-4 score. Higher values indicate better performance. For GPT-4V$^\dagger$, we perform a zero-shot evaluation on the test split; see text for details.}
\begin{tabular}{
    >{\raggedright\arraybackslash}M{20mm} 
    >{\centering\arraybackslash}M{8mm}
    >{\centering\arraybackslash}M{8mm}
    >{\centering\arraybackslash}M{8mm}
    >{\centering\arraybackslash}M{8mm}
    >{\centering\arraybackslash}M{10mm}
    }
\toprule
    \multicolumn{1}{c}{\multirow{2}{*}{\textbf{Model}}} & 
    \multicolumn{5}{c}{\textbf{Generation Task}} 
    \\
    & \multirow{1}{*}{\textbf{B4} $\uparrow$} & \multirow{1}{*}{\textbf{R} $\uparrow$} & \multirow{1}{*}{\textbf{C} $\uparrow$} & \multirow{1}{*}{\textbf{S} $\uparrow$} & \multirow{1}{*}{\textbf{GPT-4} $\uparrow$}
    \\
\midrule
    BLIP-base & 12.6 & 32.9 & 34.9 & 30.3 & 39.34 \\
    BLIP2-OPT-6.7B & \textbf{18.7} & \textbf{42.7} & 38.9 & 35.4 & 50.52 \\
    LLaVA-1.5-7B & 14.9 & 36.9 & 34.5 & 30.9 & 56.2 \\
    {GPT-4V$^\dagger$} & 0.3 & 19.0 & 0.9 & 7.2 & 50.0 \\
    \rowcolor{gray!10}
    Ours & 16.9 & 39.5 & \textbf{49.1} & \textbf{39.6} & \textbf{58.5} \\
\bottomrule
\label{tab:test_result_generation}
\end{tabular}
\end{minipage}
\vspace*{-5mm}
\end{table*}

\subsection{Training}

All models mentioned (but GPT-4V) are fine-tuned on the DHPR training set, following their respective training protocols. For the proposed method, we train (or fine-tune) the retrieval model using the image-text contrastive loss (ITC) \cite{radford2021learning} and image-text matching losses (ITM) \cite{lu2019vilbert} from two auxiliary encoders, as explained earlier. We train it in {\color{purple}15} epochs.
For the generation task, we employ the cross entropy loss as the training objective and train the model in 20 epochs following the setting in \cite{luo2023cheap}. {\color{black} All our experiments are conducted on 4 A100 GPUs (40GB each), taking approximately 80 minutes for generation tasks and about 60 minutes for retrieval tasks.}


\paragraph{Entity {Shuffle} 
Augmentation}
While a hypothesized hazard explanation can contain multiple visual entities, their order in the explanation is arbitrary, e.g., `Entity \#$1$' may appear after `Entity \#$2$' etc in the text. As explained earlier, we assign a color to each index ($n=1,2,3$), and this assignment is fixed throughout the experiments, i.e., {purple} = `Entity \#$1$,' {green} = `Entity \#$2$,' and {yellow}= `Entity \#$3$.' To facilitate the models to learn this color coding scheme, we augment each training sample by randomly shuffling the indices of entities that appear in the explanation, while we keep the color coding unchanged.

\subsection{Quantitative Results and Discussions}

{\color{black}Table \ref{tab:test_result_retrieval} presents the results of the compared methods for the retrieval tasks, while Table \ref{tab:test_result_generation} details the results for the generation tasks}. From the retrieval task results, we observe that both CLIP and our model—a variant that extends CLIP—surpass BLIP and BLIP2 in performance. This superiority is likely attributed to CLIP's comprehensive pre-training on a wide range of image-caption pairs. Additionally, our model, enhanced with dual auxiliary encoders, achieves the highest performance. 


The experimental results for the generation task present several observations. Firstly, the VLMs, such as BLIP2, LLaVA, and ours, attain fairly good performance. It is largely attributed to their use of advanced LLMs, highlighting the critical role of LLMs in the generation task.


\ok{
Secondly, in the comparison between LLaVA and our model, both utilizing the state-of-the-art language model LLaMA-2, it is evident that our model outperforms LLaVA across all metrics. This outcome is unexpected, especially considering that LLaVA (version 1.5) employs visual instruction tuning with a variety of datasets, leading to top-tier results in general vision and language tasks. The likely reason for LLaVA's underperformance may be its approach to processing image information. Our task demands specific capabilities, such as accurately identifying and understanding entities within color-highlighted boxes. These skills are closely linked to the method of extracting information from images, which depends on the model’s architecture and learning techniques. The fact that LLaVA freezes its visual encoder and uses a simple transformation to align its outputs with text tokens could be a significant factor in its lower performance.
}

The third observation concerns the evaluation results using GPT-4. It is seen that the GPT-4 scores of the VLMs, excluding GPT-4V as an inference model, generally align with traditional metrics. However, unlike traditional metrics that focus primarily on formal similarities, we can expect that the GPT-4 score provides a more nuanced measure, capturing the closeness in meaning. The scores 56-58\% for LLaVA and ours indicate a commendable level of performance in capturing true meaning, though there is still considerable room for further improvement. It is noteworthy that our model outperformed others in this respect.

The final observation is about the performance of GPT-4V as an inference model. GPT-4V operates in a zero-shot manner, meaning it has not learned what the correct text should be. As a result, the text produced by GPT-4V often significantly deviates in form from the expected correct text, usually providing overly detailed explanations. This tendency leads to its lower performance on the traditional metrics. However, GPT-4V yields a fairly high GPT-4 score. Assuming that the GPT-4 score accurately assesses the correctness of meaning, GPT-4V's zero-shot generation performance
is impressive and may open new avenues for future research. 

\begin{figure*}[t]
  \centering
  \subfloat[]
  {\includegraphics[width=0.45\textwidth, keepaspectratio]{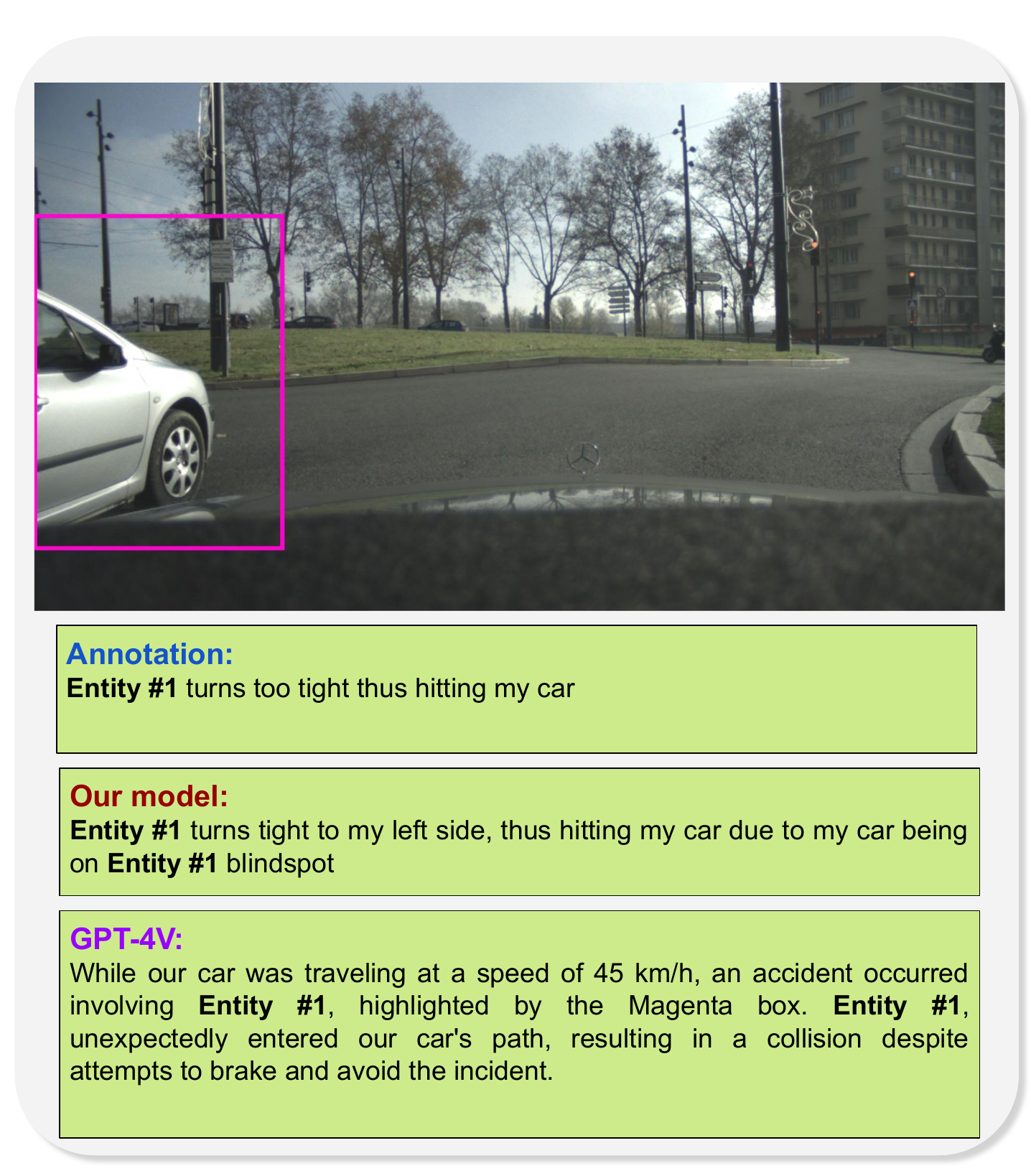}} 
  \subfloat[]
  {\includegraphics[width=0.45\textwidth, keepaspectratio]{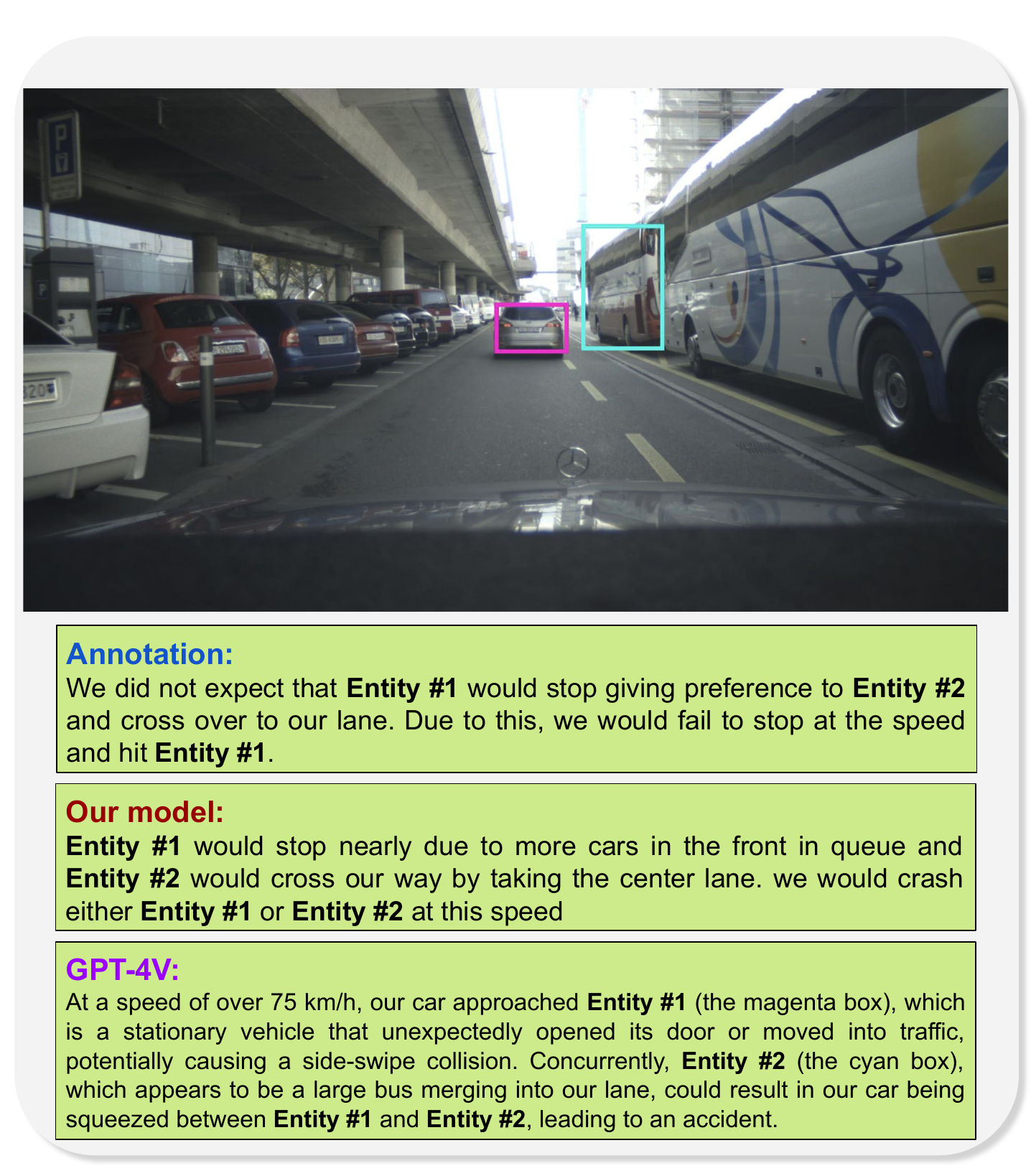}} 
  \caption{Examples of hazard explanations generated by our baseline model and GPT-4V.}
  \label{fig:hazard_explanation_qualitative}
\end{figure*}

\subsection{Qualitative Results}
We showcase hazard explanations produced by our baseline model and GPT-4V in Fig.~\ref{fig:hazard_explanation_qualitative}. 
Figure.~\ref{fig:hazard_explanation_qualitative} (a) illustrates that our model is capable of generating hazard explanations that are not only more accurate but also more semantically aligned with those provided by human annotators compared to GPT-4V.

In Fig.~\ref{fig:hazard_explanation_qualitative} (b), a scenario is presented that poses a greater challenge, necessitating the recognition of interactions among multiple objects for accurate hazard prediction. In this example, both models struggle to correctly infer the causal relationships among the involved objects. Specifically, our model does not accurately identify that the large bus (Entity \#2) is in the process of changing lanes, leading to Entity \#1 coming to a stop. GPT-4V exhibits a similar limitation in recognizing the interplay between the objects. Additional examples of hazard explanations generated by the models can be found in the supplementary materials.

\begin{table*}[t]
\centering

\caption{Results of ablation tests on the retrieval task (a-c) and the generation task (d-f)}.

\subfloat[
\textbf{Vision encoder.} Stronger CLIP vision encoder yields better retrieval scores.
\label{tab:ablation:a}]{
\centering
\begin{minipage}{0.29\linewidth}{\begin{center}
\tablestyle{2pt}{1.1}
\begin{tabular}{y{48}x{36}x{36}}
\toprule
Vision encoder & Rank $\downarrow$ & R@1 $\uparrow$ \\
\midrule
ViT-B/16 & 10.8 & 23.7\% \\
ViT-L/14 & \textbf{10.3} & \textbf{25.6}\% \\
\bottomrule
& & \\
\end{tabular}
\end{center}}\end{minipage}
}
\hspace{2em}
\subfloat[
\textbf{Auxiliary encoder}. Adding auxiliary losses has gains over merely contrastive loss.
\label{tab:ablation:b}]{
\centering
\begin{minipage}{0.29\linewidth}{\begin{center}
\tablestyle{4pt}{1.1}
\begin{tabular}{y{64}x{36}x{36}}
\toprule
Auxiliary losses & Rank $\downarrow$ & R@1 $\uparrow$ \\
\midrule
w/o aux. losses & {10.8} & {24.9}\% \\
w/\phantom{o}  aux. losses & \textbf{10.3} & \textbf{25.6}\% \\
\bottomrule
& & \\
\end{tabular}
\end{center}}\end{minipage}
}
\hspace{2em}
\subfloat[
\textbf{Shuffling entities}. Entity shuffle augmentation enhances the model's retrieval performance.
\label{tab:ablation:c}]{
\centering
\begin{minipage}{0.29\linewidth}{\begin{center}
\tablestyle{1pt}{1.1}
\begin{tabular}{y{48}x{36}x{36}}
\toprule
Shuffle entities  & Rank $\downarrow$ & R@1 $\uparrow$ \\
\midrule
No & 10.8 & 24.3\% \\
Yes & \textbf{10.3} & \textbf{25.6}\% \\
\bottomrule
& & \\
\end{tabular}
\end{center}}\end{minipage}
}
\\
\centering
\vspace{1em}
\subfloat[
\textbf{Text decoder}. Using a better LLM as the decoder yields better hazard explanations.
\label{tab:ablation:d}
]{
\begin{minipage}{0.29\linewidth}{\begin{center}
\tablestyle{2pt}{1.1}
\begin{tabular}{y{48}x{36}x{36}}
\toprule
Text decoder & B4 $\uparrow$ & CIDEr $\uparrow$ \\
\midrule
LLaMA-1 7B & 14.7 & 46.4 \\
LLaMA-2 7B & \textbf{16.9} & \textbf{49.1} \\
\bottomrule
&& \\
\end{tabular}
\end{center}}\end{minipage}
}
\hspace{2em}
\subfloat[
\textbf{Visual features}. Inputting 3 CLS tokens to the text decoder yields the best performance.
\label{tab:ablation:e}]{
\centering
\begin{minipage}{0.29\linewidth}{\begin{center}
\tablestyle{4pt}{1.1}
\begin{tabular}{y{64}x{36}x{36}}
\toprule
Visual features & B4 $\uparrow$ & CIDEr $\uparrow$ \\
\midrule
1 CLS token & 15.8 & 47.2 \\
3 CLS tokens & \textbf{16.9} & \textbf{49.1} \\
6 CLS tokens & 16.1 & 48.4 \\
\bottomrule
\end{tabular}
\end{center}}\end{minipage}
}
\hspace{2em}
\subfloat[
\textbf{Routing token}. Using only BOS as routing tokens in the adapters is better.
\label{tab:ablation:f}]{
\centering
\begin{minipage}{0.29\linewidth}{\begin{center}
\tablestyle{2pt}{1.1}
\begin{tabular}{y{48}x{36}x{36}}
\toprule
Routing token & B4 $\uparrow$ & CIDEr $\uparrow$ \\
\midrule
Indicator & 15.8 & 47.7 \\
BOS token & \textbf{16.9} & \textbf{49.1} \\
\bottomrule
&& \\
\end{tabular}
\end{center}}\end{minipage}
}

\vspace*{-3mm}
\label{tab:ablations}
\end{table*}

\subsection{Ablation Studies}

Our proposed method is composed of multiple components and configurations. We conduct ablation tests to validate their effectiveness. Table \ref{tab:ablations} shows the results. \ok{More ablation studies are provided in the supplementary material.}

In the retrieval task (Tables \ref{tab:ablation:a} - \ref{tab:ablation:c}), we find that superior vision encoders (ViT-L/14) significantly enhance performance over their less powerful counterparts (ViT-B/16), as evidenced by improved average rank and R@1 scores. The addition of auxiliary image-text matching losses further boosts these metrics, affirming their importance in fine-tuning when compared with the CLIP baseline that solely relies on contrastive loss. Lastly, using entity shuffle augmentation leads to retrieval improvement, showing that it enhances the model's proficiency in associating color codes with entity names.

For the generation task (Tables \ref{tab:ablation:d} - \ref{tab:ablation:f}), the choice of text decoder emerges as a pivotal factor. The advanced LLaMA 2 model outperforms its predecessor (LLaMA 1), leading to higher BLEU-4 and CIDEr scores. The optimal use of three CLS tokens from the vision encoder as inputs to the text decoder is also established, showing superior results over other configurations. Additionally, the use of a BOS token as a routing token in adapters proves to be more effective than an indicator token proposed by \cite{liu2023visual}.


\section{Conclusion and Discussions} \label{sec6}
We have introduced a new approach to predicting driving hazards that utilizes recent advancements in multi-modal AI, to enhance methodologies for driver assistance and autonomous driving. Our focus is on predicting and reasoning about driving hazards using scene images captured by dashcams. We formulate this as a task of visual abductive reasoning.

To assess the feasibility and effectiveness of the approach, we curated a new dataset called DHPR (Driving Hazard Prediction and Reasoning), featuring around 15,000 dashcam-captured scene images, annotated through crowdsourcing with details like car speed, hazard explanations, and visual elements marked by bounding boxes and text. The dataset was used to create specific tasks and evaluate model performances, including a CLIP-based baseline, VLMs, and GPT-4V, on image-to-text and text-to-image retrieval tasks and a text generation task. The results affirmed the approach's feasibility and efficacy, offering insights for future research.

{\color{black} 
This paper serves as a preliminary exploration of using multimodal AI for abductive reasoning in car driving, mimicking human behavior. Due to technical constraints and data acquisition challenges, the current method relies solely on still images and vehicle speed for inference, limiting its ability to address all real-world driving hazards. Despite these limitations, our results affirm the framework's utility.

Future research will extend the framework to include video input and a broader set of vehicle data, enhancing hazard identification capabilities. Notably, our framework can adapt to incorporate additional vehicle information simply by altering the initial prompts fed into our models. For instance, introducing instructions like `As we turn right' allows the model to integrate steering data to anticipate hazards contextually, which will be explored further in subsequent studies.



}




\section*{Acknowledgments}
{\color{black}This work was partly supported by JSPS KAKENHI Grant Number 20H05952 and 23H00482.}




\clearpage
\section{Biography Section}
\begin{IEEEbiography}[{\includegraphics[width=1in,height=1.25in,clip,keepaspectratio,trim={21cm} {18cm} {19cm} {12cm}]{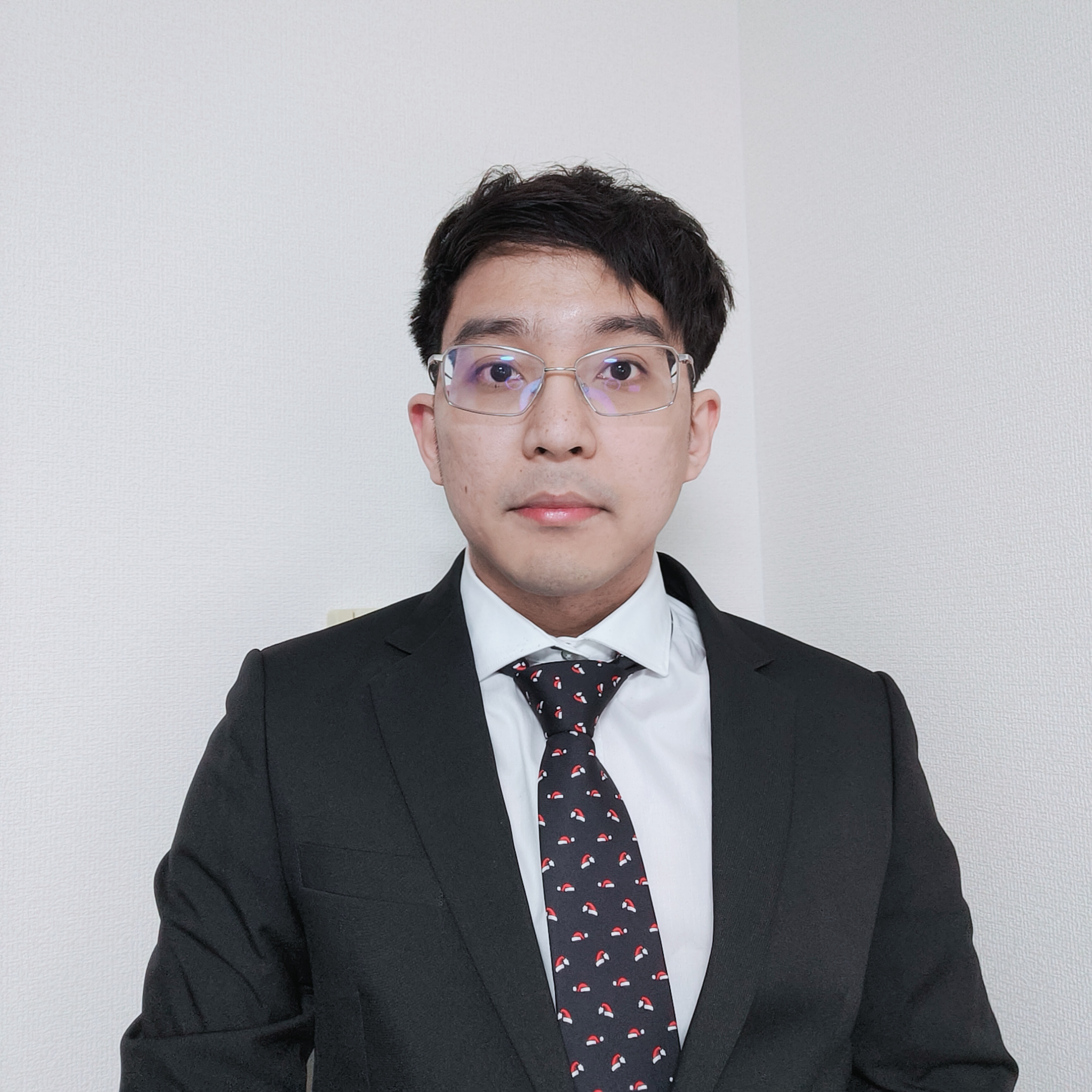}}]%
{Mr. Korawat Charoenpitaks}
is currently pursuing a Ph.D. degree at the Graduate School of Information Sciences, Tohoku University. His research interests are in the field of computer vision and natural language processing.
\end{IEEEbiography}
\begin{IEEEbiography}[{\includegraphics[width=1in,height=1.25in,clip,keepaspectratio,trim={16cm} {18cm} {14cm} {0cm}]{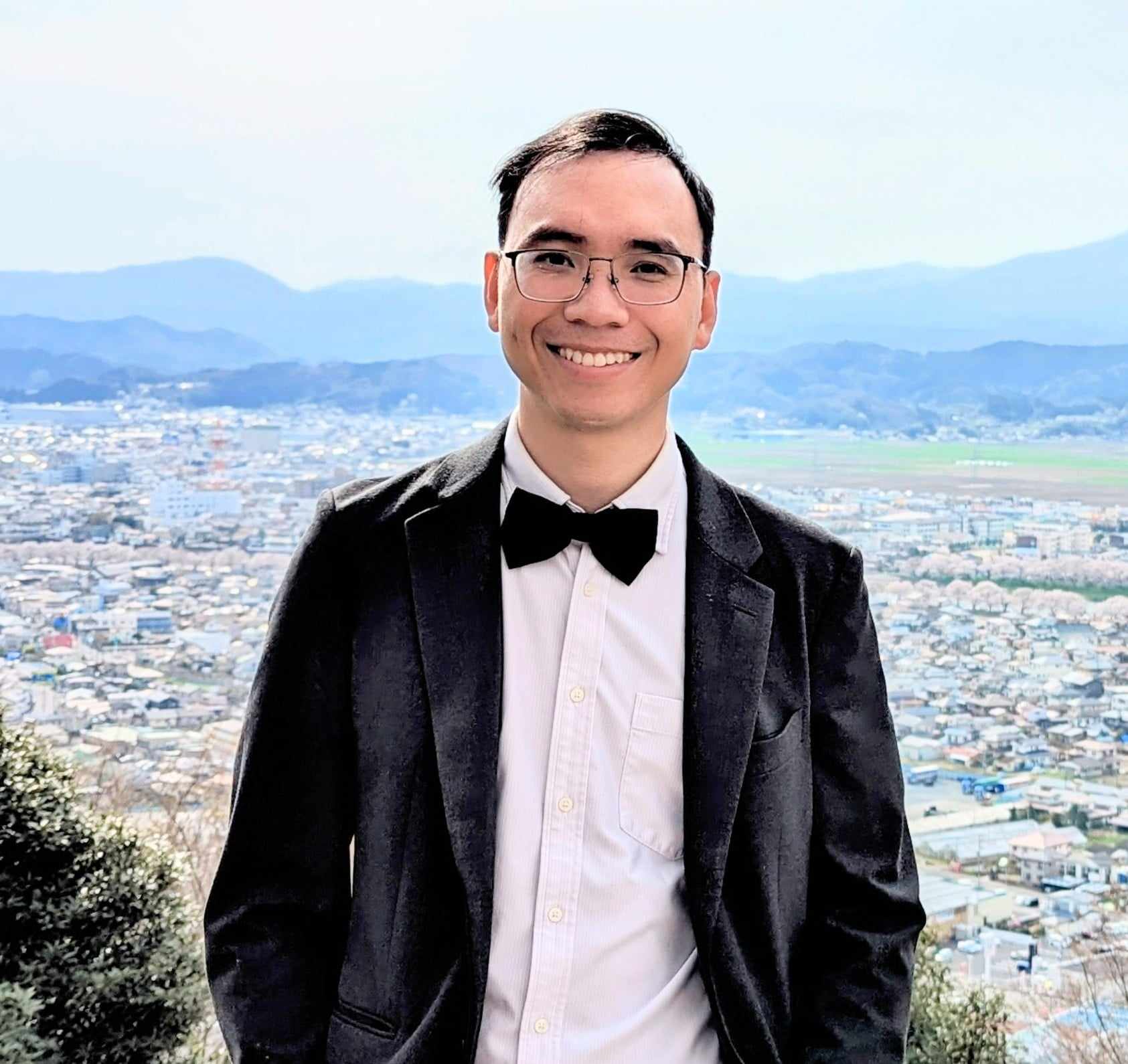}}]%
{Dr. Van-Quang Nguyen}
received the Ph.D. degree from Graduate School of Information Sciences, Tohoku University, in 2022. He is currently the postdoctoral researcher at RIKEN AIP, Japan. His research interests are in the intersection of computer vision and natural language processing.
\end{IEEEbiography}
\begin{IEEEbiography}[{\includegraphics[width=1in,height=1.25in,clip,keepaspectratio,trim={0cm} {0cm} {0cm} {0cm}]{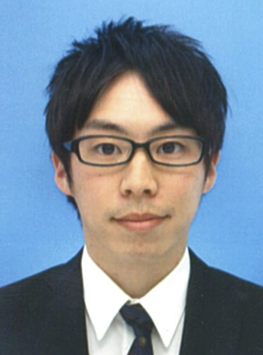}}]%
{Dr. Masanori Suganuma}
received the Ph.D. degree from Graduate School of Environment and Information Sciences, Yokohama National University, in 2017. He is currently an Assistant Professor at Tohoku University. His research interests are in the field of computer vision and machine learning.
\end{IEEEbiography}
\begin{IEEEbiography}[{\includegraphics[width=1in,height=1.25in,clip,keepaspectratio,trim={4cm} {3cm} {4cm} {1cm}]{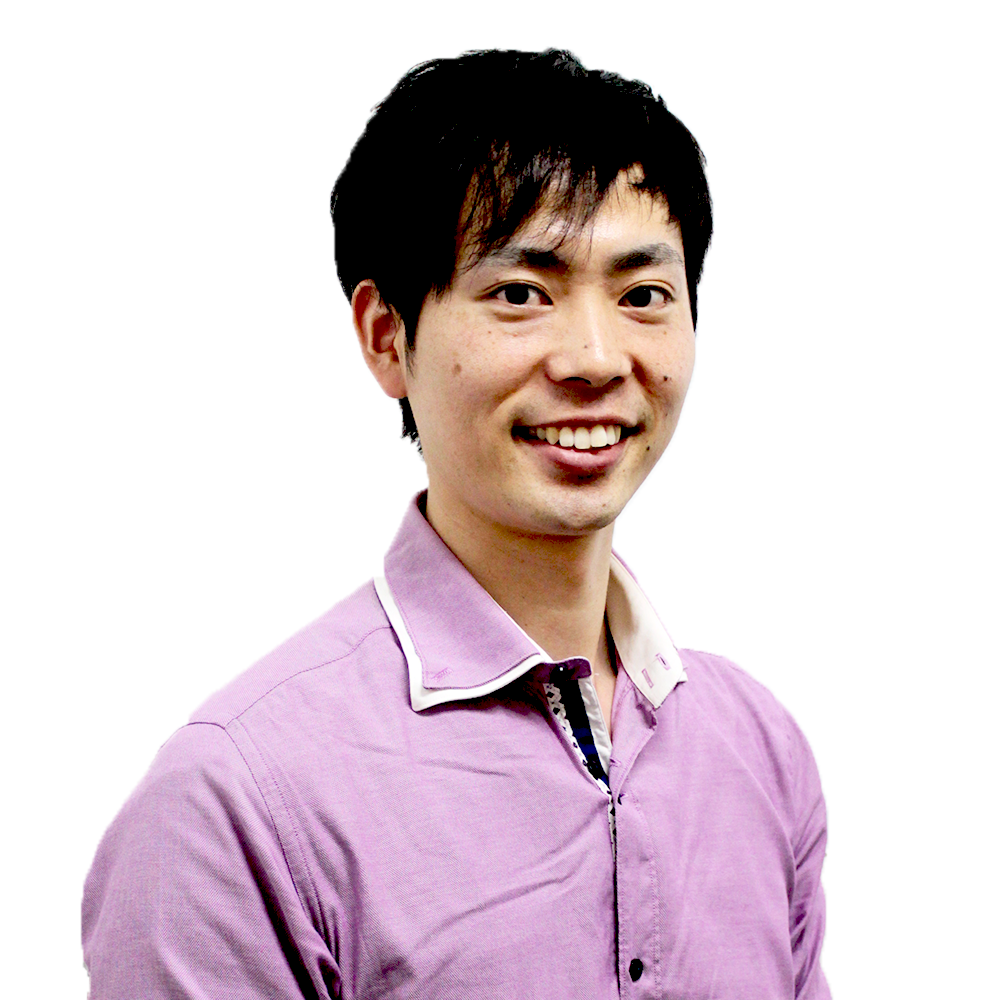}}]%
{Dr. Masahiro Takahashi} received his Ph.D. in science (Physics) from the Graduate School of Natural Science and Technology at Okayama University in 2010. He is currently working as a Project Assistant Manager at DENSO CORPORATION. His current research interests lie in the field of machine learning.
\end{IEEEbiography}
\begin{IEEEbiography}[{\includegraphics[width=1in,height=1.25in,clip,keepaspectratio,trim={6cm} {0cm} {6cm} {2.5cm}]{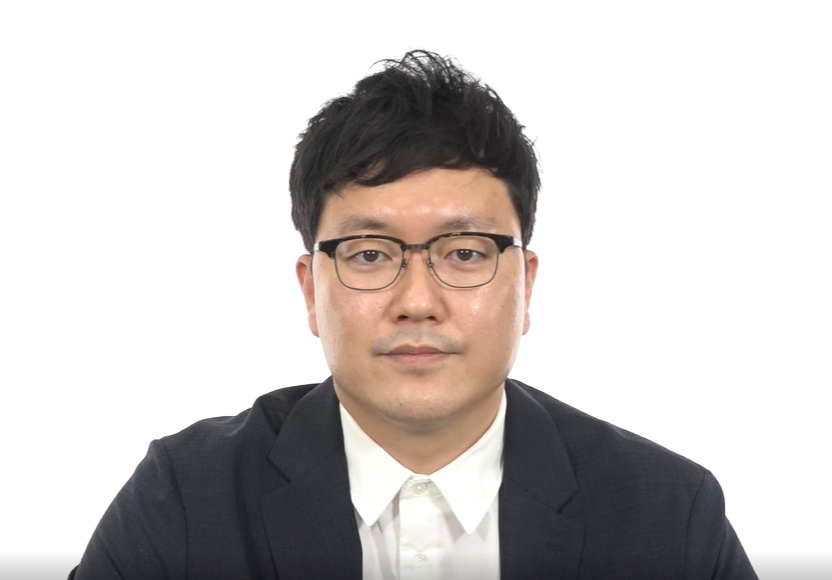}}]%
{Mr. Ryoma Niihara} received the master degree from Kyushu Institute of Technology in 2009. He is currently a manager at DENSO CORPORATION. His research interests are in the field of machine learning and autonomous vehicles.
\end{IEEEbiography}
\begin{IEEEbiography}[{\includegraphics[width=1in,height=1.25in,clip,keepaspectratio,trim={0.5cm} {0.5cm} {0.5cm} {0.5cm}]{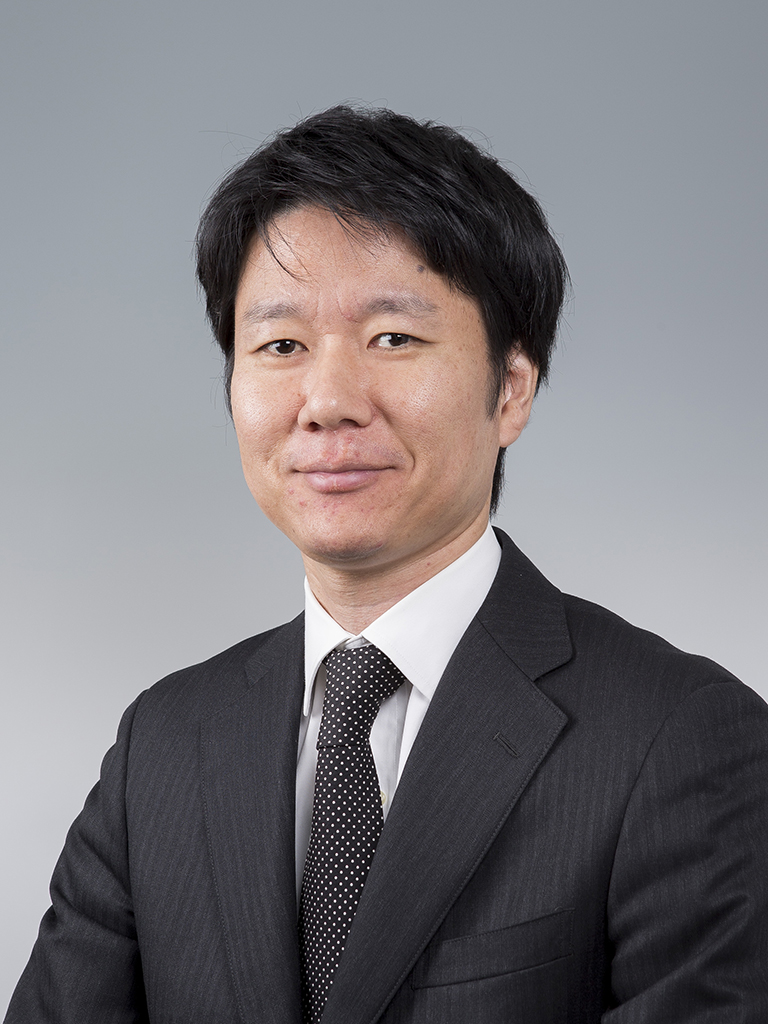}}]%
{Dr. Takayuki Okatani} earned his B.Eng., M.Sc., and Ph.D. degrees in Mathematical Engineering and Information Physics from the Graduate School of Engineering at the University of Tokyo in 1994, 1996, and 1999, respectively. He currently serves as a Professor in the area of computer vision at Tohoku University. In addition, he heads the Infrastructure Management Robotics Team at the RIKEN Center for Advanced Intelligence Project. With over 100 publications in peer-reviewed journals and conference proceedings, his work encompasses computer vision, deep learning, and multi-modal AI. He is an active member of several professional societies, including the IEEE Computer Society, the Information Processing Society of Japan, the Institute of Electronics, Information and Communication Engineers, and the Society of Instrument and Control Engineers.
\end{IEEEbiography}
\end{document}


\clearpage
\setcounter{page}{1}
\title{\Large{Exploring the Potential of Multi-Modal AI \\for Driving Hazard Prediction \\ Supplementary Material}}
\maketitle

\section{Access to the DHPR Dataset}

\subsection{URLs}

The Driving Hazard Prediction and Reasoning (DHPR) dataset 
will be publicly available at:
\begin{itemize}
\item \url{https://github.com/DHPR-dataset/DHPR-dataset} 
\item \url{https://huggingface.co/datasets/DHPR/Driving-Hazard-Prediction-and-Reasoning}
\end{itemize}
This dataset includes annotations paired with images sourced from two external datasets, {\color{black} BDD100K \cite{bdd100k} and ECP \cite{Braun_2019}}. 
Users can access the images from their original sources at:
\begin{itemize}
\item \url{https://bdd-data.berkeley.edu/}
\item \url{https://eurocity-dataset.tudelft.nl/eval/overview/statistics}
\end{itemize}
Additionally, we have developed a website to facilitate the exploration of the dataset. This website will also be accessible to the public at:
\begin{itemize}
    \item \url{https://huggingface.co/spaces/DHPR/Demo}
\end{itemize}

\subsection{Notes on Availability and Maintenance of the Data}

The DHPR dataset created in this study will be openly available for access from the URL given above. We plan to establish an evaluation server and leaderboard in the future. Any updates pertaining to the dataset will be communicated through the aforementioned repository, ensuring that users have access to the most up-to-date information.

\subsection{Ethical and Responsible Use}

The present study complies with the ethical standards for responsible research practice.
Our dataset is built upon images of two existing datasets, ECP and BDD100K. 
It is compliant with GDPR for ECP \cite{Braun_2019} and other data-related regulations for BDD100K \cite{bdd100k}. We protected the anonymity of personal information by blurring identifiable details in the images used in both the main paper and this supplementary material. The datasets are sourced following the licensing regime of each dataset. 

\section{Statistics of the Dataset}


Table \ref{tab:comprehensive_dataset_analysis} provides a detailed statistical breakdown of the dataset. It first shows the source of each image, either from the ECP or BDD dataset for the training, validation, and test sets, respectively. The table 
offers further granularity for each set by showing the speed of our car, the average length of hazard descriptions, hazard type, and the common position words used. It also includes the hazard types, which are already shown in Table 2 (in the main paper). 

Almost all of the annotated hazards pertain to the risk of the ego-vehicle potentially colliding with an object. Table \ref{tab:analysis-dataset-split} details the various entities that the self-car may collide with within the hazard explanations for each data split.




\begin{table}[]
\centering
\caption{Types of entities that the self-car is described as hitting in the hazard explanations.}
\label{tab:analysis-dataset-split} \small
\begin{tabular} 
{
>{\centering\arraybackslash}p{1.3cm}
>{\centering\arraybackslash}p{1.3cm}
>{\centering\arraybackslash}p{1.3cm}
>{\centering\arraybackslash}p{1.3cm}
} 
\toprule
\textbf{Split} & \textbf{Car} & \textbf{Pedestrian \& Bike} & \textbf{Others}  \\ 
\midrule
validation & 854 & 125 & 19 \\ 
test & 844 & 110 & 40 \\ 
\bottomrule
\end{tabular}
\end{table}

\begin{table*}[]
    \caption{Detailed Dataset Statistics}
    \centering
        \begin{adjustbox}{max width=0.9\textwidth}
        \begin{tabular}{
                >{\arraybackslash}p{65mm}
                >{\centering\arraybackslash}p{22mm}
                >{\centering\arraybackslash}p{22mm}
                >{\centering\arraybackslash}p{22mm}
                }
            \toprule
                \textbf{Data Attributes} & \textbf{Training Set} & \textbf{Validation Set} & \textbf{Test Set} \\
            \midrule
                \textbf{Data Sample} & 12975 & 1000 & 1000 \\
            \midrule
                \textbf{Image sourced from:} & & & \\
                \indent\hspace{0.5cm} ECP dataset & 6968 & 470 & 482 \\
                \indent\hspace{0.5cm} BDD dataset & 6007 & 530 & 518 \\
            \midrule
                \textbf{Our car speed:} & & & \\
                \indent\hspace{0.5cm} 15 km/h & 5260 & 418 & 402 \\
                \indent\hspace{0.5cm} 45 km/h & 4649 & 380 & 388 \\
                \indent\hspace{0.5cm} 75+ km/h & 3066 & 202 & 210 \\
            \midrule
                \textbf{Hazard:} & & & \\
                \indent\hspace{0.5cm} Avg. Length Tokens & 25.9 & 22.7 & 23.2 \\
                \indent\hspace{0.5cm} Unique Words & 3645 & 1216 & 1265 \\
                \indent\hspace{0.5cm} Less than 5 Occurrence Words (\%) & 62.8\% & 67.7\% & 68.0\% \\
                \indent\hspace{0.5cm} More than 10 Occurrence Words (\%) & 26.0\% & 19.7\% & 18.3\% \\
                \indent\hspace{0.5cm} \textbf{Position words:} & & &  \\
                \indent\hspace{1cm} Left & 1481 & 118 & 125 \\
                \indent\hspace{1cm} Right & 1572 & 97 & 120 \\
                \indent\hspace{1cm} Front & 1842 & 95 & 101 \\
                \indent\hspace{1cm} Side & 1528 & 92 & 114 \\
                \indent\hspace{1cm} Back & 1136 & 102 & 113 \\
                \indent\hspace{0.5cm} \textbf{Accident-related words:} & & & \\
                \indent\hspace{1cm} Hit & 7623 & 604 & 573 \\
                \indent\hspace{1cm} Crash & 1619 & 94 & 106 \\
                \indent\hspace{1cm} Collide & 759 & 52 & 63 \\
                \indent\hspace{1cm} Clip & 221 & 50 & 42 \\
            \midrule
                \textbf{Hazard Type:} & & & \\
                \indent\hspace{0.5cm}\textbf{Entity Related:} & & & \\
                \indent\hspace{1cm}\hspace{1mm} Pedestrian & - & 104 & 84 \\
                \indent\hspace{1cm}\hspace{1mm} Stationary Object & - & 15 & 61 \\
                \indent\hspace{1cm}\hspace{1mm} Traffic Signal & - & 46 & 41 \\
                \indent\hspace{1cm}\hspace{1mm} Unusual Condition & - & 20 & 16 \\
                \indent\hspace{0.5cm}\textbf{Driving Scenario:} & & & \\
                \indent\hspace{1cm}\hspace{1mm} Speeding \& Braking & - & 473 & 413 \\
                \indent\hspace{1cm}\hspace{1mm} Sideswipe & - & 52 & 48 \\
                \indent\hspace{1cm}\hspace{1mm} Merging Maneuver & - & 264 & 305 \\
                \indent\hspace{1cm}\hspace{1mm} Unexpected Event & - & 20 & 23 \\
                \indent\hspace{1cm}\hspace{1mm} Chain Reaction & - & 6 & 9 \\
            \bottomrule
            \centering
            \end{tabular}
    \end{adjustbox}
    \label{tab:comprehensive_dataset_analysis}
\end{table*}

\section{Using GPT-4 for Evaluating Generated Texts}

As stated in our main paper, our experiments used OpenAI's GPT-4 (not the GPT-4V(ision) used for inference) as one of the evaluation metrics for the generation task. We explain this method in detail here. To evaluate a model, we measure how similar the hazard explanation texts generated by the model are to the correct texts, i.e., the annotated hazard explanations. Our aim is to supplement conventional metrics, which can measure formal similarity but not semantic similarity. Here, GPT-4 refers to the {\tt gpt-4-1106-preview} model and was used via the API provided by OpenAI.

\begin{figure*}[t]
\centering
\includegraphics[width=1.0\linewidth]{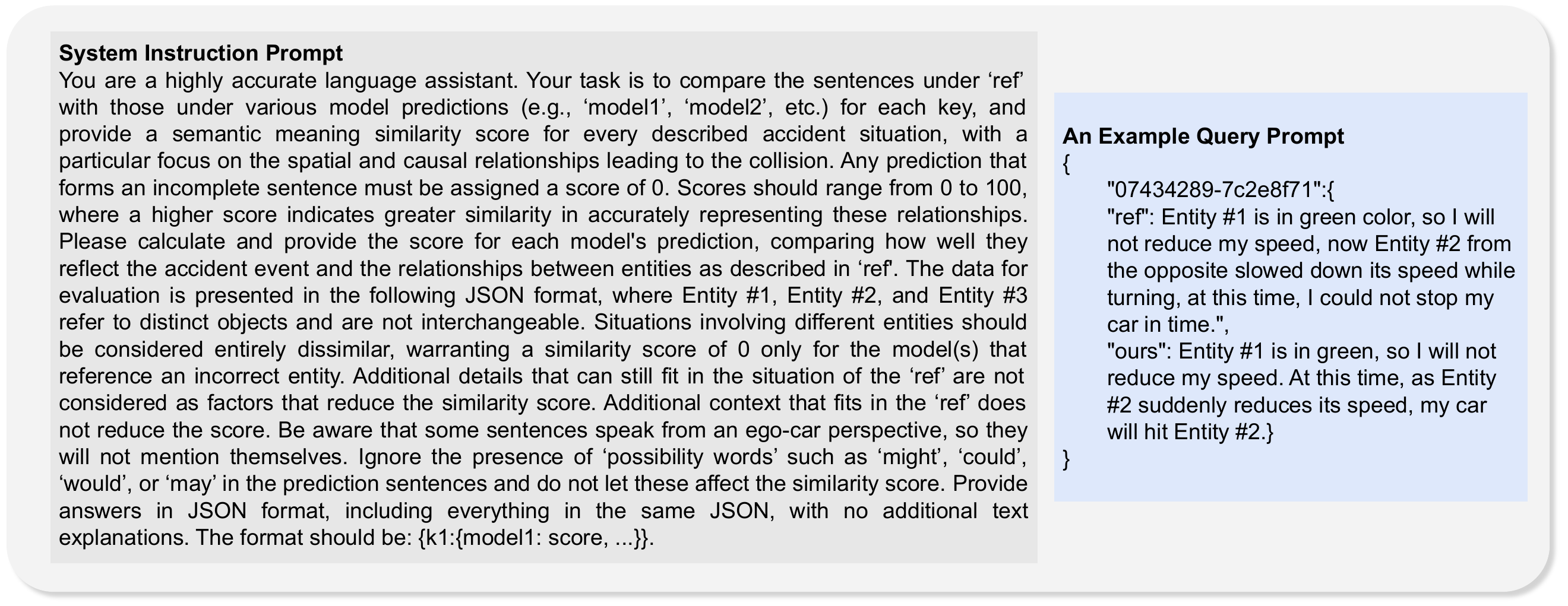}
\caption{Our prompt for GPT-4 to evaluate the similarity between a model-generated hazard explanation text and its corresponding ground truth text. This prompt includes a system instruction, followed by query prompts. Each query prompt contains the generated text alongside the ground truth text for comparison.}
\vspace*{-3mm}
\label{fig:gpt4_eval_instruction_prompt}
\end{figure*}

Specifically, when given the correct and generated texts, we input a prompt containing these into GPT-4, and have it respond with a score reflecting their similarity, as shown in Fig.~\ref{fig:gpt4_eval_instruction_prompt}. The prompt is constructed by placing a task instruction at the beginning, followed by the correct and generated texts as a query. The construction of the prompt, particularly the task instruction, is crucial to making the evaluation meaningful. We focused on several aspects, described below. 

Firstly, we set the output score range as 0-100. A perfect semantic match with the correct answer yields 100, while no commonality (including incomplete generated texts) results in 0. Intermediate values should ideally represent semantic similarity aligned with human intuition.

Secondly, given the nature of the hazard prediction task, it's essential that the generated text correctly refers to the entities indicated by colored boxes in images. This accuracy should be rigorously evaluated.

\begin{figure*}[b]
\centering
\includegraphics[width=1.0\linewidth]{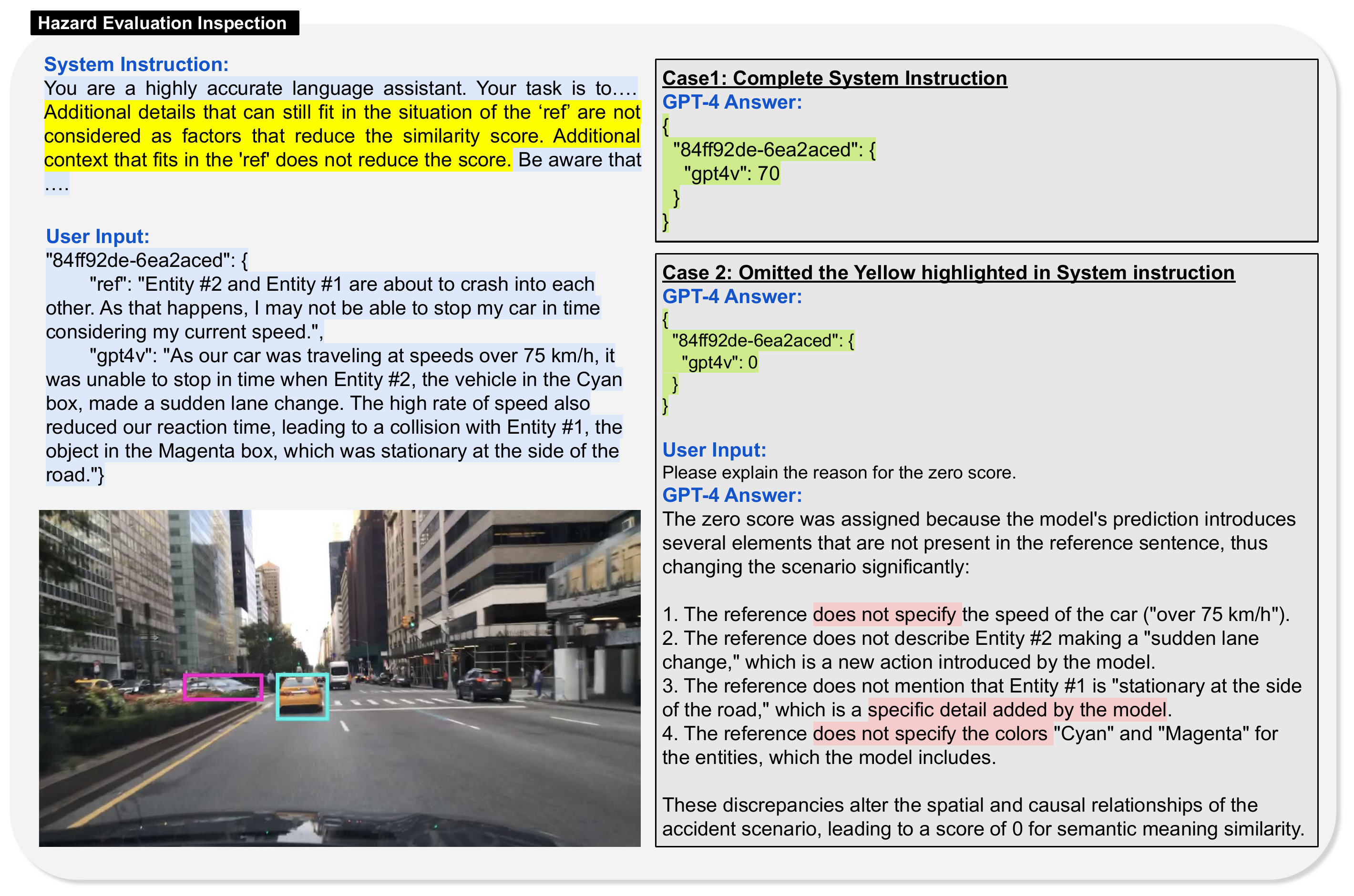}
\caption{Example of GPT-4 evaluation of a generated text from GPT-4V with and without the policy of unpenalizing for extra contents.}
\vspace*{-3mm}
\label{fig:gpt4_eval_inspection}
\end{figure*}

However, as this task involves hypothetical hazard reasoning, there's often no single correct answer. Therefore, we only require the generated text to mention the same content as the correct text, without penalizing for extra content not included in the correct text. This policy significantly affects the evaluation of GPT-4V as an inference model, as it tends to generate texts with additional content not in the correct answer. This is reasonable as GPT-4V infers in a zero-shot manner (without learning from DHPR training data), as described in Sec.~{\color{red} 4.2} of the main paper and Sec.~\ref{sec:gpt-4v} in this supplementary material. 
Figure \ref{fig:gpt4_eval_inspection} illustrates how the presence or absence of this policy changes the evaluation of the same generated text from GPT-4V. 

The input prompt is designed with these considerations in mind; see Fig.~\ref{fig:gpt4_eval_instruction_prompt}.
\newpage

For the sake of efficiency in evaluation processing through the API, we included 25 pairs of correct and generated texts, corresponding to 25 scenes (images), in a single prompt. Furthermore, to ensure the reproducibility of the evaluation, we set the temperature of GPT-4 to 0.




\section{Hazard Explanation Generation by GPT-4V}
\label{sec:gpt-4v}
In the experiments for the generation task, we evaluated GPT-4V along with other models. Here, we detail how GPT-4V executed these tasks. Unlike traditional text-only models, GPT-4V processes both text and images as input. As reported in \cite{yang2023dawn}, it can perform a variety of image understanding tasks, and is currently considered one of the most advanced VLMs. For the experiments, the gpt-4-vision-preview model was utilized via the OpenAI API.

\begin{figure}[t]
\centering
\includegraphics[width=1.0\linewidth]{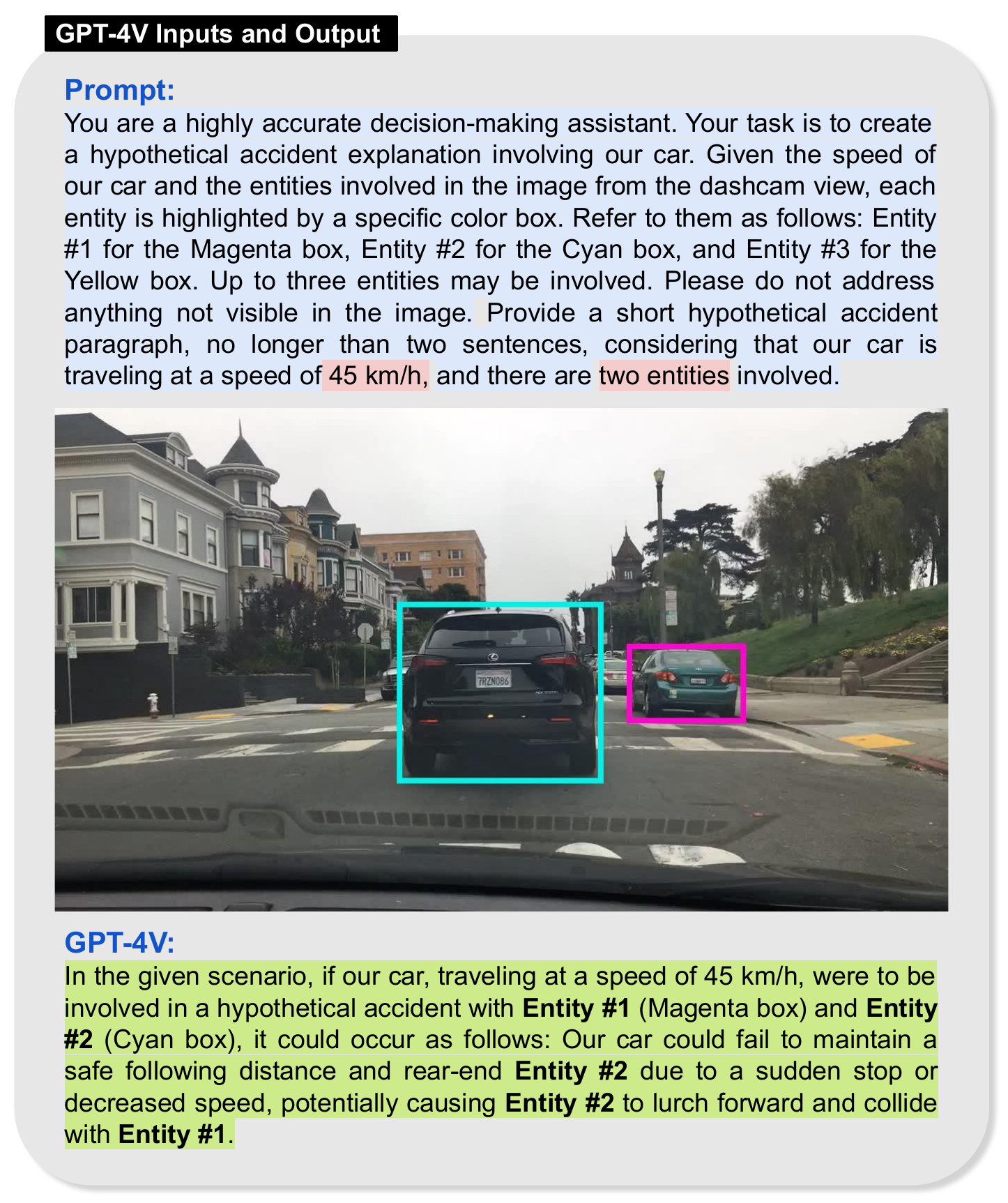}
\caption{Example of input text and image prompts for GPT-4V and output text. 
}
\vspace*{-3mm}
\label{fig:gpt4v_input_template}
\end{figure}

For each scene, we input two inputs to GPT-4V: an image and a text prompt. Examples of these inputs, along with their corresponding output results, are illustrated in Fig.~\ref{fig:gpt4v_input_template}.

As shown in the figure, the input image includes color-coded bounding boxes (BBs), explicitly specifying entities in the scene involved in a hypothesized hazard. Specifically, Entity \#$1$, \#$2$, and \#$3$ correspond to magenta, cyan, and yellow,
respectively. In the text prompt (explained later), GPT-4V is instructed to refer to these entities in the generated text using the format of `Entity \#$n$', corresponding to the color-coded BBs in the image.

It is important to note that while the method of indicating entities with colored BBs is common across all models compared, we adopted a different approach for GPT-4V to coloring these boxes than the other models. Unlike other models for which each box is filled with an opaque
color, we simply colors the outline of the BBs since this is a proven method for GPT-4V \cite{SoM}. Even without training using images with embedded color-coded BBs, GPT-4V works sufficiently well by simply having each input text prompt explain which entities in the image correspond to Entity \#$1$-$3$.

The other input, the text prompt, consists of a template that is fixed regardless of the input scene and contextual information, which changes according to the input scene, embedded in the template; see Fig.~\ref{fig:gpt4v_input_template}. The template provides the task instructions, while the context information includes two pieces of information: the ego-vehicle's speed and the number of entities involved in the hazard to predict, both of which are included in the DHPR dataset.

\begin{figure*}[t]
\centering
\includegraphics[width=1.0\linewidth]{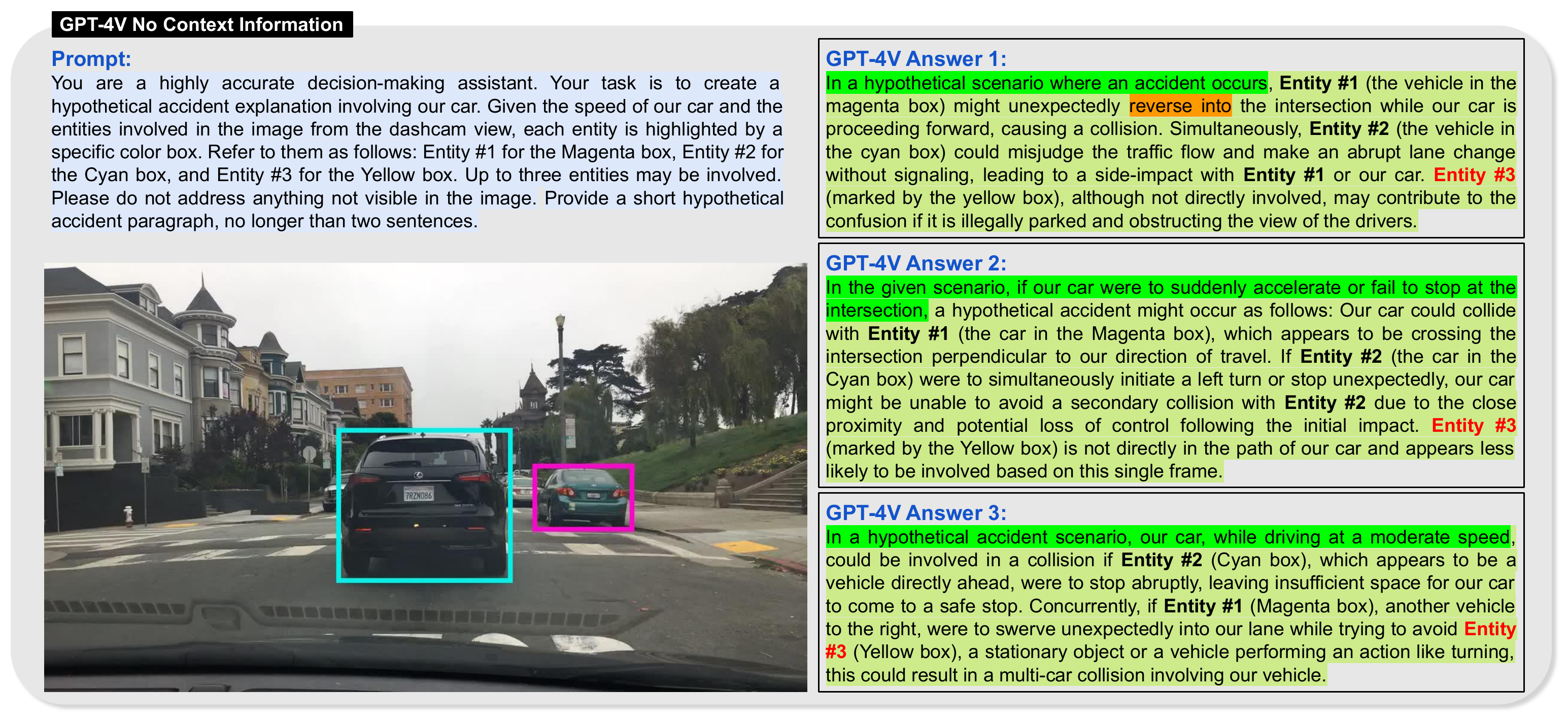}
\caption{
Example of generated texts from GPT-4V when no context information is provided (i.e., ego-vehicle speed and the number of involved entities).
}
\vspace*{-3mm}
\label{fig:gpt4v_no_context}
\end{figure*}

Note that this information is not provided (i.e., not included in the prompt) to models other than GPT-4V. This approach for GPT-4V, partly due to its zero-shot inference nature, is intended to prevent its outputs from exceeding the intended scope of the dataset. Figure \ref{fig:gpt4v_no_context} illustrates output examples in the absence of these context details—three different outputs for the same input. As these examples demonstrate, without specifying the ego-vehicle's speed, GPT-4V contemplates various possibilities for the vehicle's behavior, such as driving in reverse or unexpected acceleration. Similarly, not specifying the number of entities can lead GPT-4V to fabricate non-existent entities based on the color coding information in the instruction template, always referring to the maximum number of entities.

Examples of generated hazard explanations are presented in Sec.~{\color{red} H}.

\clearpage

{\color{black}
\section{Details of Experimental Setup}

\subsection{More Details of the Proposed Model}
\label{sec:ours}

The proposed baseline model consists of a text encoder, a vision encoder, and an LLM, as explained in Sec.~{\color{black} 5.1} of the main paper. 

For the text and vision encoders, we utilize the pre-trained model sourced from  OpenAI's CLIP repository\footnote{\url{https://github.com/openai/CLIP}}. We augment each encoder by incorporating two additional Transformer layers, which are stacked on top of the existing structure. These added layers adhere to the standard Transformer architecture having self-attention and cross-attention mechanisms. In terms of configuration, each layer is designed with a token dimensionality of 512 and is equipped with eight attention heads. Furthermore, we have implemented a dropout rate of 0.1. To enhance positional awareness, we employ relative position embeddings \cite{shaw2018self}, with a maximum distance of 128 and a total of 32 buckets. 

For the LLM, we utilize the LLaMA-2 7B model. We follow the configuration specified in Luo et al.'s research on visual instruction tuning with LLMs \cite{luo2023cheap}.

Our model incorporates two loss functions to train the retrieval tasks: the Image-Text Contrastive (ITC) loss and the Image-Text Matching (ITM) loss. For the ITC loss, we follow the original CLIP \cite{radford2021learning}. Regarding the ITM loss, we intentionally introduce a mismatched pair in half of the image-text pairs during training. 
In our model's architecture, there are two auxiliary encoders: the Image-to-Text (I-2-T) and Text-to-Image (T-2-I) encoders. The ITM loss is calculated from their outputs as follows.
Each of the two base encoders generates a [CLS] token.
The ITM encoders transform the corresponding class token into a binary logit, for which the binary cross-entropy loss is calculated. 

After completing training on retrieval tasks, we proceed to fine-tune our model for generation tasks. As outlined in Sec.~5.1 of the main paper, we repurpose the trained vision encoder and integrate it with the LLM. This phase of training utilizes the standard objective of next token prediction and employs the DHPR training set. In this setup, the model receives a preprocessed image and a text prompt as inputs, and generates hazard explanation text as its output. We follow the configuration parameters detailed in \cite{luo2023cheap} for our model's fine-tuning\footnote{\url{https://github.com/luogen1996/LaVIN}}. This includes setting an effective batch size of 32 and running a total of 20 training epochs. In accordance with their guidance, we configure the adapters within both the vision encoder and the LLM to have a dimensionality of 8. Notably, during the training phase, we exclusively optimize these adapter parameters.






\subsection{Training of Compared Models}

For the retrieval tasks, we apply a uniform training procedure across all models, including CLIP, BLIP, BLIP2, and our own model. We utilized the public repository for each model. Each model begins training with weights from its corresponding pretrained version and undergoes 15 epochs of weight updates. For the baseline, CLIP employs the ITC loss. BLIP adds the ITM loss; see also Sec.~\ref{sec:ours}.

BLIP2 utilizes the Image-Grounded Text Generation (ITG) loss within its Quantum-Former (Q-Former) architecture, in addition to the ITC and ITM losses. For training BLIP2, we adopt the stage-1 training method from the original BLIP2 study by Li et al. (2023) \cite{li2023blip2}, using the DHPR training set. Thus, the ITG loss involves prompting the Q-Former to generate hazard explanation texts based on input images, which are then compared to their ground truths.


For the generation task, we adhere to standard fine-tuning procedures, specifically tailored to each model. In alignment with their respective original studies, we fine-tune\footnote{\url{https://github.com/salesforce/LAVIS/tree/main}} the BLIP model for 5 epochs \cite{li2022blip} and the BLIP2 model for 5 epochs \cite{li2023blip2}. However, we deviate from the recommended protocol for LLaVA-1.5\footnote{\url{https://github.com/haotian-liu/LLaVA}}, opting to fine-tune it for 12 epochs, as opposed to the suggested single epoch \cite{llava1_5}. This adjustment is based on our findings that the standard setting for LLaVA-1.5 resulted in subpar performance, whereas extending the fine-tuning to 12 epochs – a strategy derived from LLaVA-1.0 – significantly improves outcomes.
We utilize the same loss of next token prediction for the training; see also Sec.~\ref{sec:ours} for more details.

\subsection{Image Preprocessing}

In line with standard practices, we preprocess input images for all models under comparison as follows. For models other than those using ViT-L/14, images are first resized to a square dimension of $224+16$, and then randomly cropped to $224\times 224$. In contrast, for baseline models employing ViT-L/14, images are resized to a square dimension of 
$336+16$ and subsequently cropped to 
$336\times 336$. Additionally, we apply color jitter augmentation with settings of 0.5 for brightness, 0.3 for hue, and 0.3 for saturation, prior to accentuating regions containing entities of interest. Notably, we avoid using horizontal flip augmentation to preserve the spatial integrity of the images.

{\color{black}
\section{Ablation Test with Image/Text Inputs}

As outlined in the main paper, the DHPR dataset allows us to design various tasks, each with a distinct level of difficulty. Selecting the text/image retrieval tasks as a testbed, we conducted further experiments to explore the impact of different input formats. In particular, we varied the format of each of a scene image and a hazard explanation to assess their effects on retrieval performance. In all the experiments, we used the {\color{black} proposed} baseline. 

\paragraph{Image Input} 

To test the effects of different formats of image inputs, we considered four types of format: position only, `no entity', `no context', and `only context', as illustrated in Fig.~\ref{fig:ablation}. We trained and tested the model using each format.

The results are presented in Table \ref{tab:ablation_result}. Notably, when only the positions of entities were shown (`position only'), a significant decline in performance occurred, with the average rank dropping to over 16.7. This outcome is reasonable as the model lacks visibility of the visual entities or context. Furthermore, excluding any direct specification of entities (`no entity') also led to considerably poorer performance. This result highlights the necessity for the models to accurately identify the entities present in the given hazard explanations to accurately estimate the similarity between an input image and an explanation.

\begin{table}[]
    \caption{Ablation results for the retrieval tasks with different input data formats. The proposed model for retrieval tasks is utilized in the experiments.}
    \centering
    \begin{adjustbox}{width=\columnwidth}
    \begin{tabular}{
            >{\arraybackslash}p{30mm}
            >{\centering\arraybackslash}p{12mm}
            >{\centering\arraybackslash}p{12mm}
            >{\centering\arraybackslash}p{12mm}
            >{\centering\arraybackslash}p{12mm}
            }
        \toprule
            \multicolumn{1}{l}{\multirow{2}{*}{\textbf{Model}}} & \multicolumn{2}{c}{\textbf{IR Task}} & \multicolumn{2}{c}{\textbf{TR Task}} \\
            &  \textbf{Rank} $\downarrow$ & \textbf{R@1} $\uparrow$ & \textbf{Rank} $\downarrow$ & \textbf{R@1} $\uparrow$ \\
        \midrule
            Baseline & 10.2 & 24.9\% & 10.3 & 26.3\% \\
        \midrule
            Image: position only & 16.7 & 12.9\% & 16.9 & 11.7\%  \\
            Image: no entities & 20.2 & 11.3\% & 18.2 & 12.2\%  \\
            Image: no context & 12.0 & 22.9\% & 11.8   & 20.2\% \\
            Image: only context & 12.6 & 19.7\% & 12.4  & 18.1\% \\
        \midrule
            Comprehensive text (train only) & 14.3 & 18.9\% & 14.2  & 17.0\% \\
            Comprehensive text (train \& test) / \textbf{Oracle} & 6.5 & 56.8\% & 6.6  & 58.2\% \\
        \bottomrule
        \end{tabular}
    \end{adjustbox}
    \label{tab:ablation_result}
\end{table}

Interestingly, when the context was omitted from the input images (labeled as `no context'), we observed relatively improved results. The retrieval ranks varied between 11.8 and 12.0, and the Recall at 1 (R@1) exceeded 20\%. However, these results were marginally but noticeably inferior to those of the baseline. In contrast, using only the contextual information (denoted as `only context') led to significantly poorer outcomes compared to the removal of context. These findings highlight the critical role of both visual entities and their surrounding context in making accurate predictions.

\paragraph{Text inputs}
\label{subsec:comprehensive-hazard}

In terms of text input formats, we add the descriptions of visual entities to hazard explanations. It's important to note that these descriptions have not been utilized in our previous experiments, although DHPR includes them as part of its annotations. Specifically, we enhance the hazard explanation for each sample by incorporating the descriptions of all visual entities into the corresponding section of the explanation. This involves replacing the first occurrence of ``Entity \#$n$'' with ``Entity \#$n$, <its description>.'' For example, the following explanation
\begin{itemize}
\item[] 
``Entity \#$1$ decides to go behind of Entity \#$2$ to cross street misjudges my speed, can't stop in time and hits Entity \#$1$''
\end{itemize}
changes into
\begin{itemize}
\item[] 
``Entity \#$1$, cyclist on right side by sidewalk,  decides to go behind of Entity \#$2$, white car in front of my car, to cross street misjudges my speed, can't stop in time and hits Entity \#$1$.''
\end{itemize}

\begin{figure*}[t]
		\centering
		\includegraphics[width=0.8\linewidth]{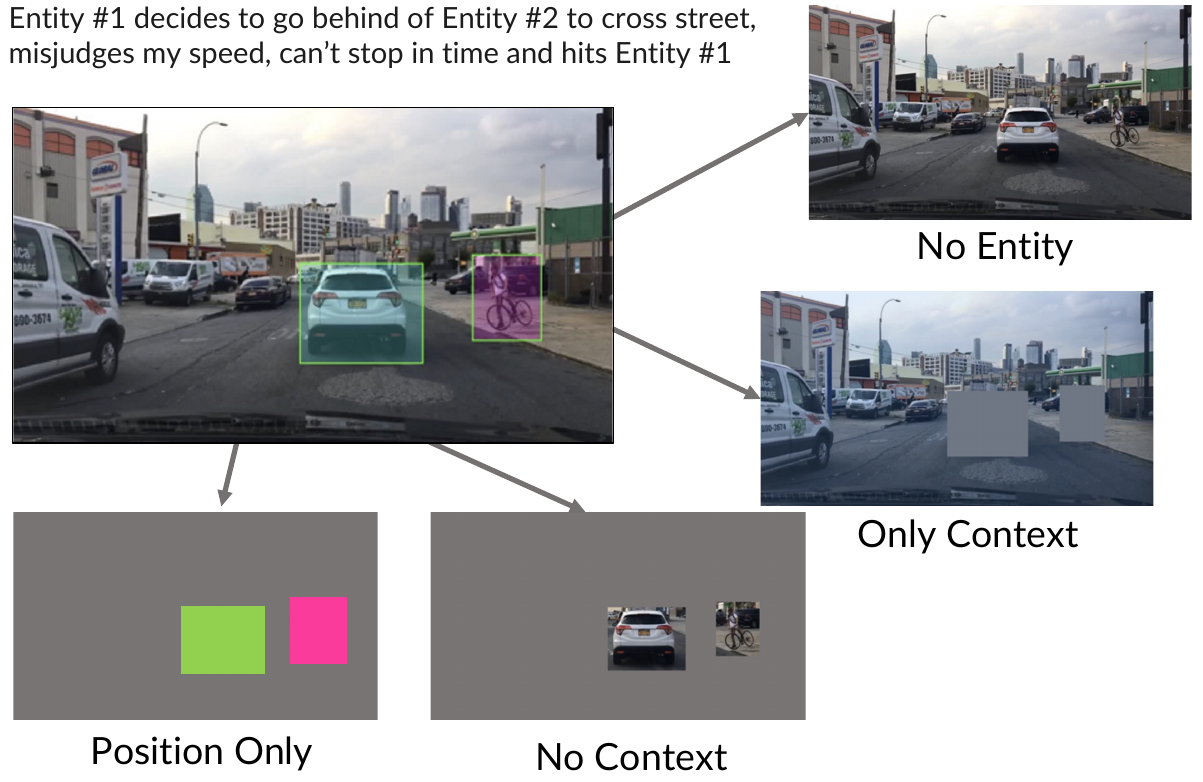}
    \caption{Illustrations of our image input ablations, which are conducted by drawing in pixel-space directly, following \cite{hessel2022abduction}.}
    \label{fig:ablation}
\end{figure*}

We employed two experiments for training and testing the model. The first experiment involved training the model using the comprehensive format of explanations mentioned above and testing it on the original format of explanations (without the descriptions). This experiment aimed to improve the model's performance on the test split while maintaining the same experimental setting as before.

The second experiment involved testing the same model on the comprehensive explanation format, which includes the additional descriptions. This experiment aimed to evaluate the impact of incorporating these descriptions. However, it is important to note that obtaining the descriptions requires inference and is not freely available. Therefore, we consider this case as an "oracle" scenario.

The lower block of Table \ref{tab:ablation_result} presents the results. When evaluating the model trained on comprehensive explanations on the original explanations without the entities' descriptions, it performs significantly worse than the baseline. This decline in performance can be attributed to the disparity in the format between the training and test data, indicating that the intended aim was not achieved. However, when the same model is evaluated on the test data using the comprehensive format, 
there is a notable enhancement in its performance, {\color{black} an average reduction of 6.5 in its ranking.}
We attribute this improvement mainly to the model's ability to associate the added entities' descriptions with the contents of the images. It is why we termed the setting `oracle.'



\section{Qualitative Analysis on the Retrieval Tasks}
{\color{black}

We present here multiple results for the text retrieval task using our proposed baseline model. Figure \ref{fig:resuls-correct1} displays two instances of successful retrievals. In both cases, the ground-truth explanation appears first in the rankings, denoting accurate retrieval. Notably, in Fig.~\ref{fig:resuls-correct1}(b), the top-ranked results are semantically similar to the ground-truth explanation positioned at the top rank.



\begin{figure*}[]
  \centering
  \subfloat[]
  {\includegraphics[width=0.5\textwidth]{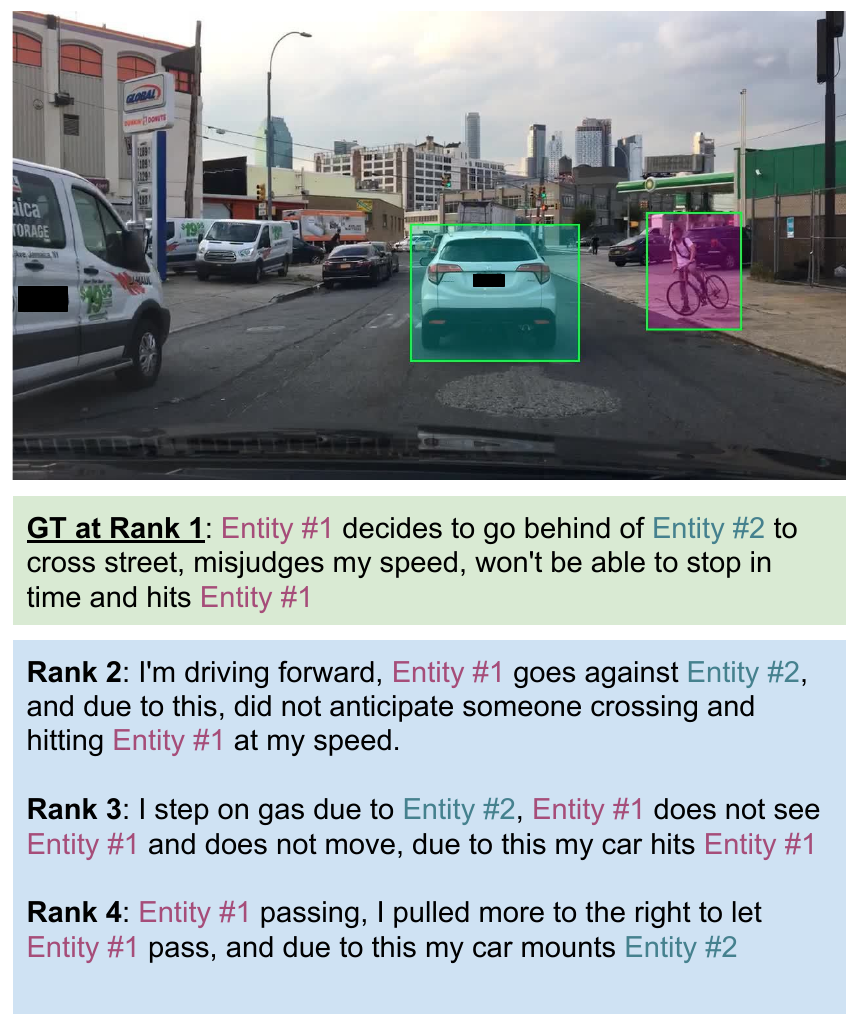}} 
  \hfill
  \subfloat[]
  {\includegraphics[width=0.5\textwidth]{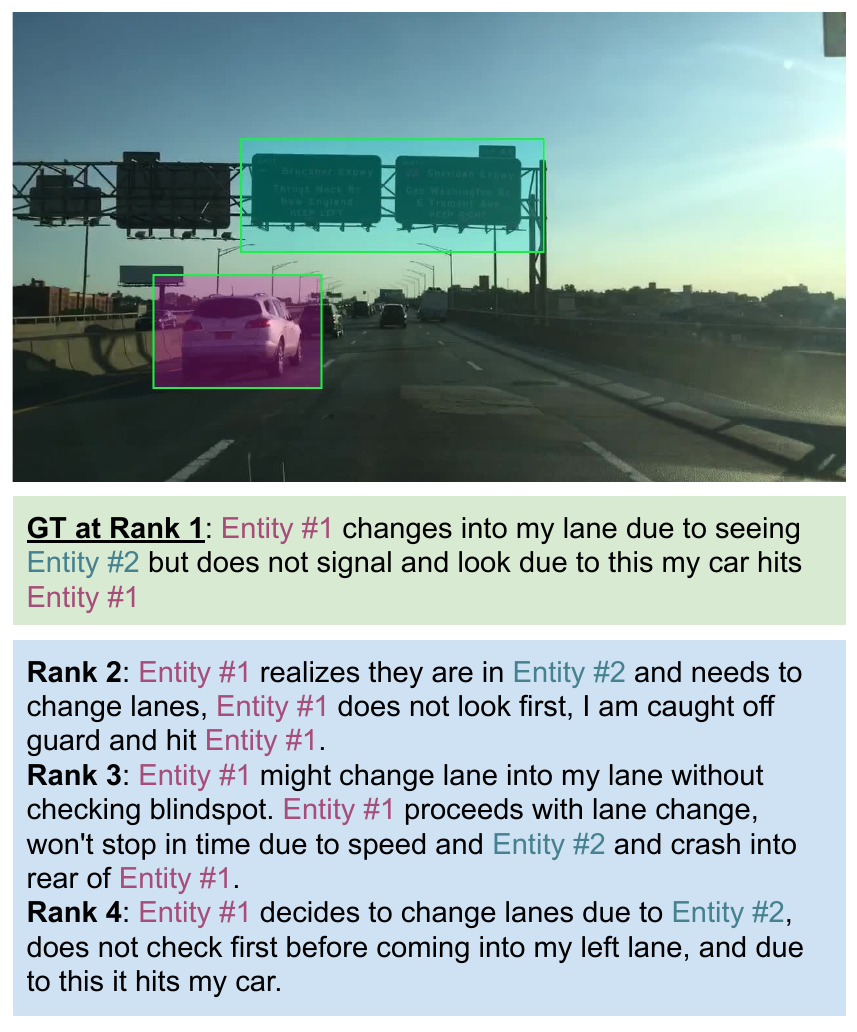}} 
  \caption{Example of image-to-text retrieval by our best-performing model, including the annotated hazard (GT) and its rank, alongside the other top three candidates. Each candidate rank is indicated as \textbf{Rank n}.}
  \label{fig:resuls-correct1}
\end{figure*}


Figure \ref{fig:resuls-interesting1} shows challenging examples of the text retrieval task. In the scene shown in Fig.~\ref{fig:resuls-interesting1}(a), the ground-truth explanation expects our car to turn left and collide with Entity \#$1$. The explanations ranked 1st and 2nd propose that our car continues straight while Entity \#$1$ makes a left turn, resulting in a collision. Both explanations present valid hazard hypotheses. Figure \ref{fig:resuls-interesting1}(b) showcases an example where the ground-truth explanation is ranked very low, specifically at the 25th position. This discrepancy might be due to the term `red lights' being used to refer to a specific concept, the reverse ramp of a bus, rather than its typical meaning of a traffic signal. Nonetheless, the explanations ranked 1st, 2nd, and 3rd convey meanings that are highly similar to the ground-truth explanation: the self-car cannot stop and collides with the bus.

\begin{figure*}[]
  \centering
  \subfloat[] 
  {\includegraphics[width=0.5\textwidth]{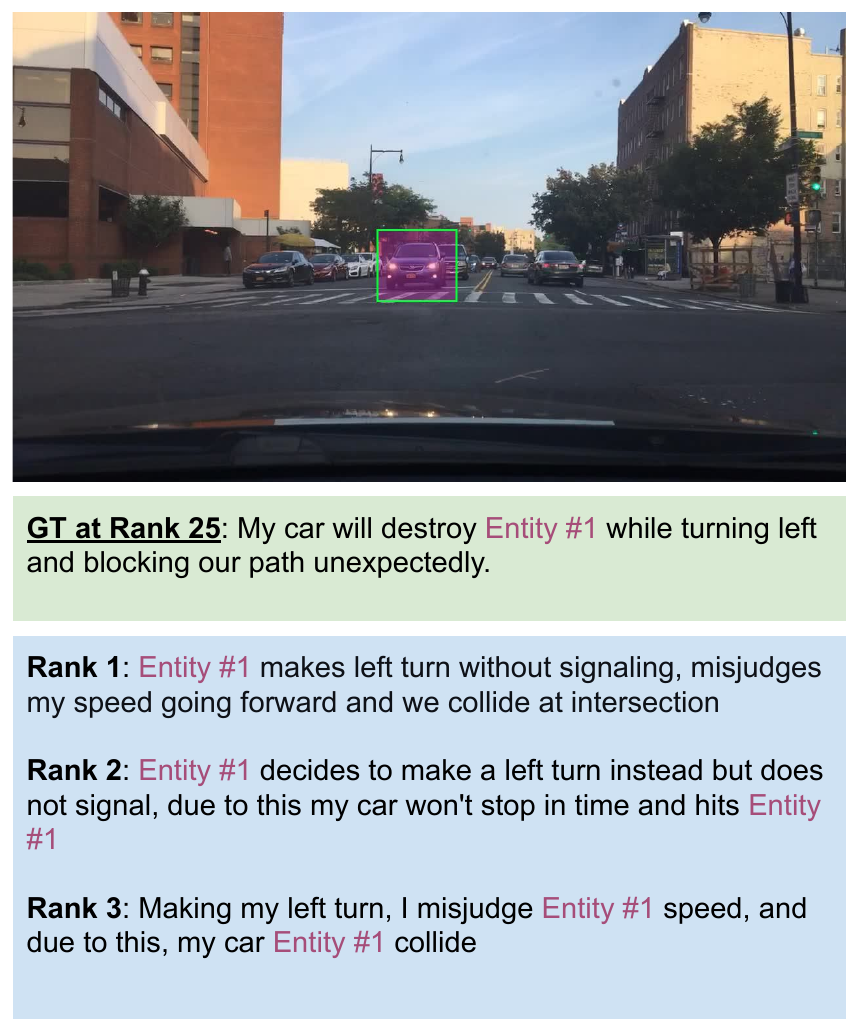}} 
  \hfill
  \subfloat[] 
  {\includegraphics[width=0.5\textwidth]{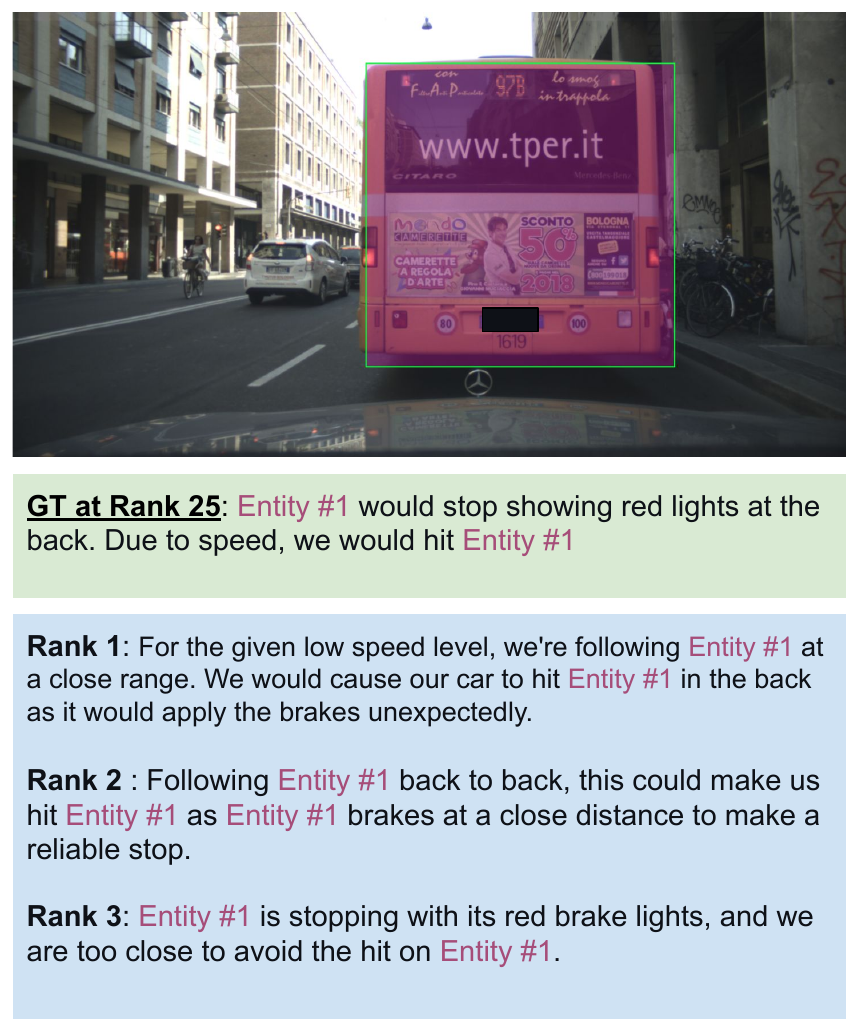}} 
  \caption{Examples of \textbf{challenging cases} of image-to-text retrieval by our best-performing model, including the annotated hazard (GT) and its rank, alongside the other top three candidates. Each candidate rank is indicated as \textbf{Rank n}.}
  \label{fig:resuls-interesting1}
\end{figure*}

\begin{figure*}[]
  \centering
  \subfloat[] 
  {\includegraphics[width=0.5\textwidth]{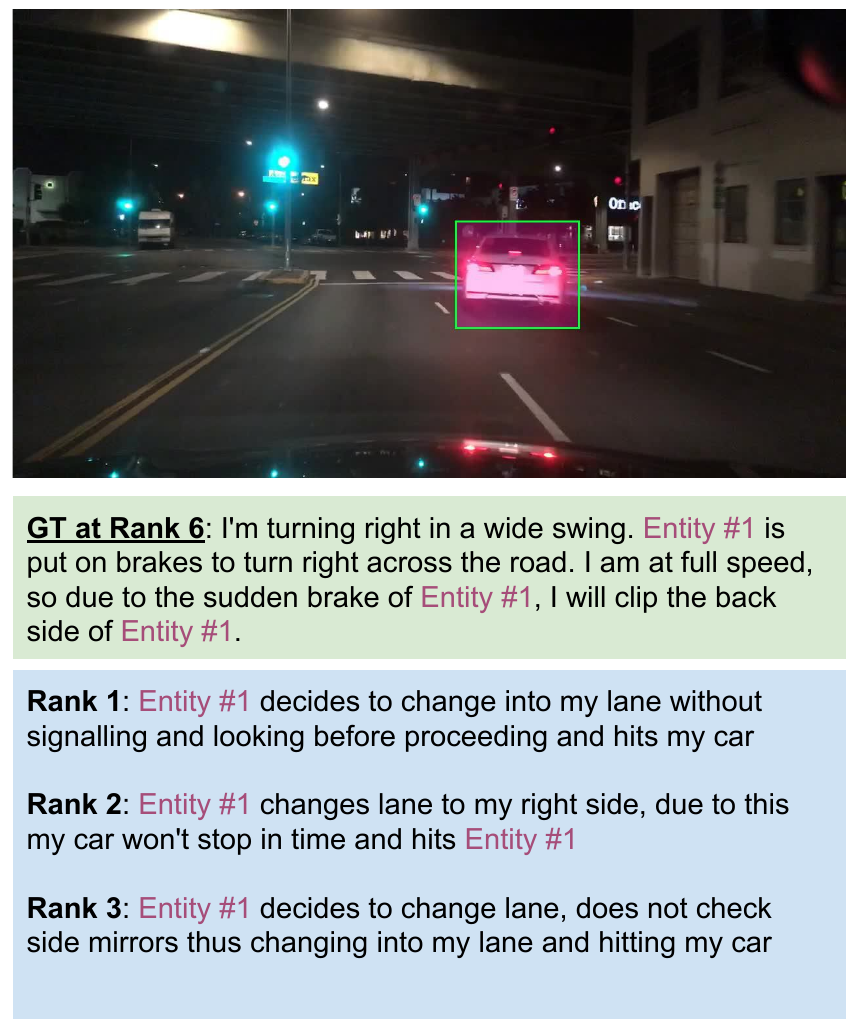}}
  \hfill
  \subfloat[] 
  {\includegraphics[width=0.5\textwidth]{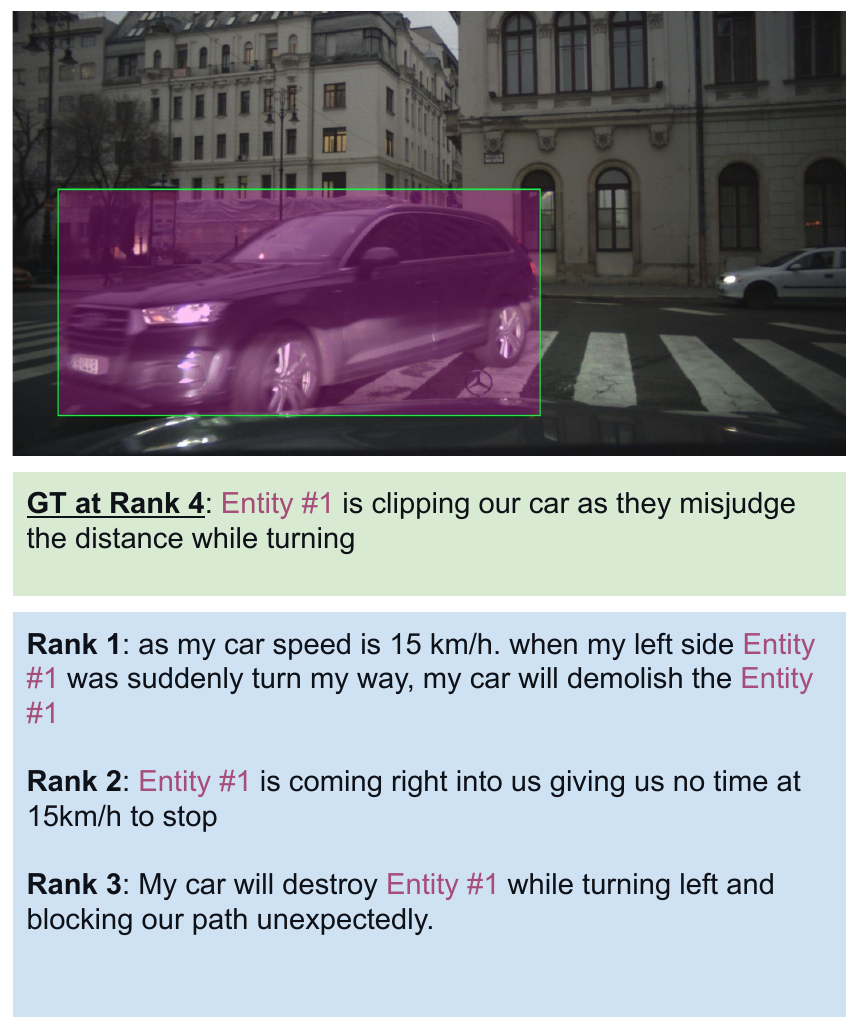}} 
  \caption{Examples of \textbf{challenging cases} of image-to-text retrieval by our best-performing model, including the annotated hazard (GT) and its rank, alongside the other top three candidates. Each candidate rank is indicated as \textbf{Rank n}.}
  \label{fig:resuls-challenge1}
\end{figure*}

We found from the results that our model tends to favor shorter, similar versions of hazard explanations, leading to an inaccurate ranking of longer and more complex sentences. A typical example is shown in Fig.~\ref{fig:resuls-challenge1}(a), where the ground-truth explanation was assigned only the 6th rank.


Furthermore, we encountered different types of challenges that may arise from using a single image instead of a video or multiple frames. In Fig.~\ref{fig:resuls-challenge1}(b), the car in front of ours appears to be turning left towards our left side. However, our model seems to assume that the car is moving backwards and turning right in front of our car. Consequently, it retrieved hazard explanations that predict the convergence of both cars into the same lane, resulting in a collision.

\section{Qualitative Analysis on the Generation Task}

We present multiple examples of hazard explanation texts produced by our baseline model and GPT-4V. Remember, our model has undergone fine-tuning using the DHPR training set, whereas GPT-4V has not and relies solely on zero-shot inference. However, GPT-4V is provided with extra context information, such as the speed of the ego-vehicle and the total number of entities involved in the hazard.

Figure \ref{fig:hazard_speeding} presents two examples pertaining to the hazard type {\em Speeding \& Braking}. Both our model and GPT-4V generate hazard explanations that closely match the annotated example. Our model consistently follows a specific sentence structure, as outlined in the instructions. In contrast, while GPT-4V produces sensible explanations, it deviates from the instructed format. This deviation results in a failure to explicitly mention one of the two entities shown in Fig.~\ref{fig:hazard_speeding}(b).



Figure \ref{fig:hazard_merging} displays examples for the {\em Merging Maneuver} hazard type. Our model generates explanations that are similar in content to the annotated ones, but they are overly simplified and could apply to a wide variety of situations. GPT-4V, on the other hand, incorrectly identifies the traffic sign in example (a) and the movement direction of the referenced vehicle in example (b). Moreover, it typically starts with descriptions of the specified entities before moving on to the reasoning aspect.

In the {\em Pedestrian} hazard type examples shown in Fig.~\ref{fig:hazard_pedestrian1}, our model produces hazard explanations that align semantically with the annotated examples but, similar to previous instances, are overly simplistic. In contrast, GPT-4V offers detailed descriptions of the entities involved and bases its hazard predictions on these details. However, 
in example (a), the incorrect interpretation of the Entity\#$1$'s action leads to the prediction of an unplausible hazard.



\begin{figure*}[t]
  \centering
  \subfloat[]
  {\includegraphics[width=0.48\textwidth, keepaspectratio]{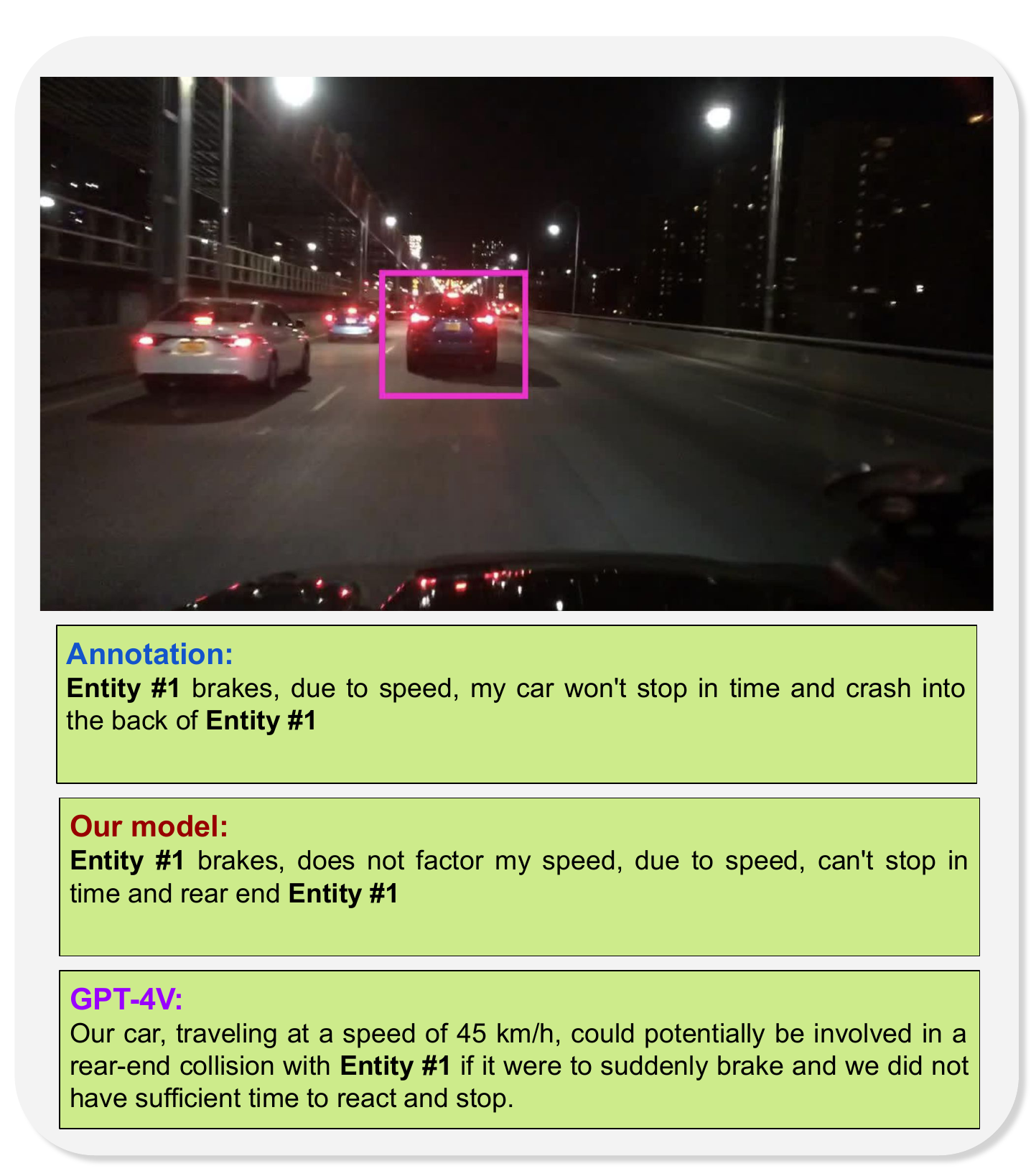}} 
  \subfloat[]
  {\includegraphics[width=0.48\textwidth, keepaspectratio]{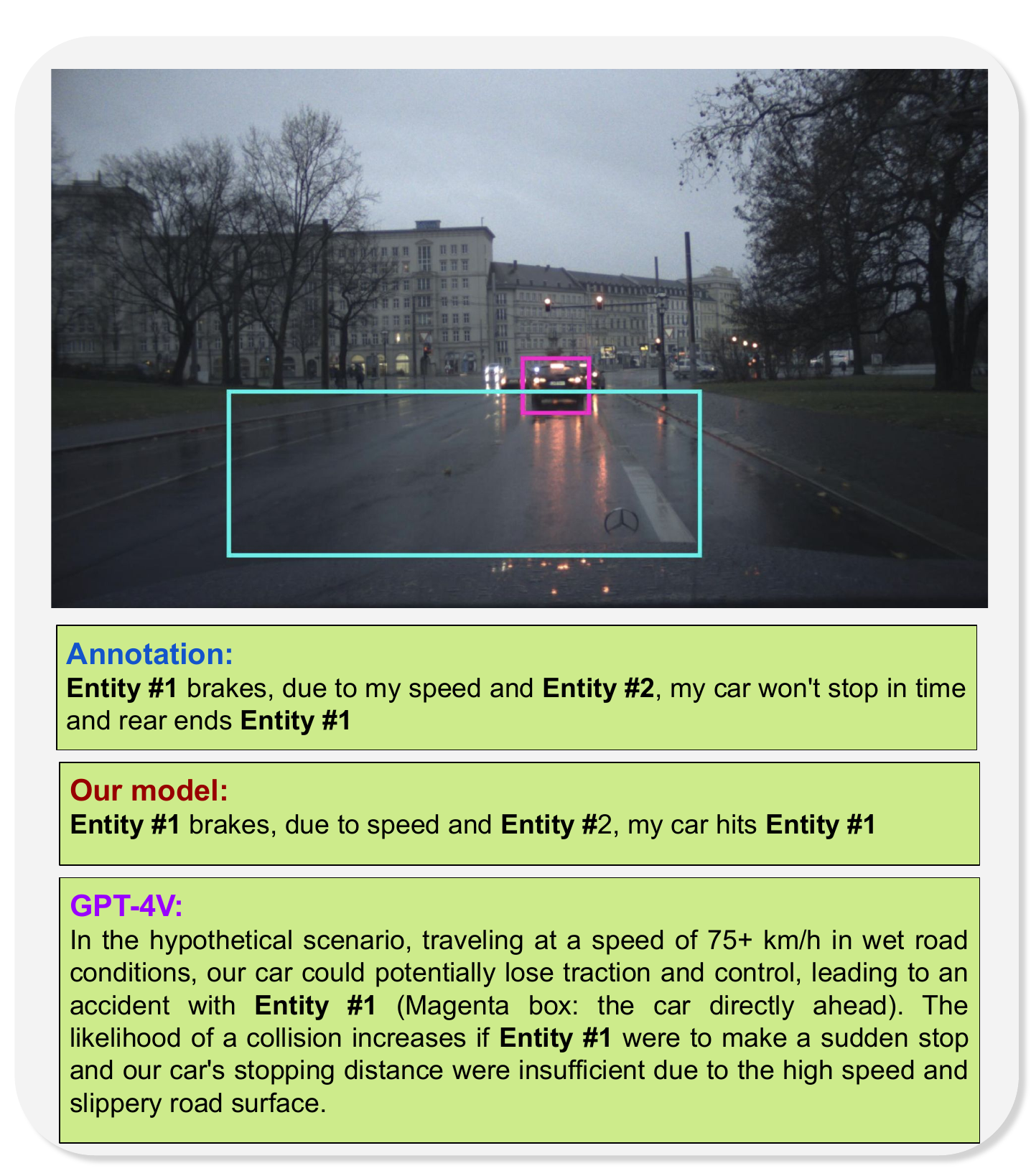}} 
  \caption{Hazard explanation texts generated by our baseline model and GPT-4V: {\em Speeding \& Braking}.}
  \label{fig:hazard_speeding}
\end{figure*}

\begin{figure*}[t]
  \centering
  \subfloat[]
  {\includegraphics[width=0.48\textwidth, keepaspectratio]{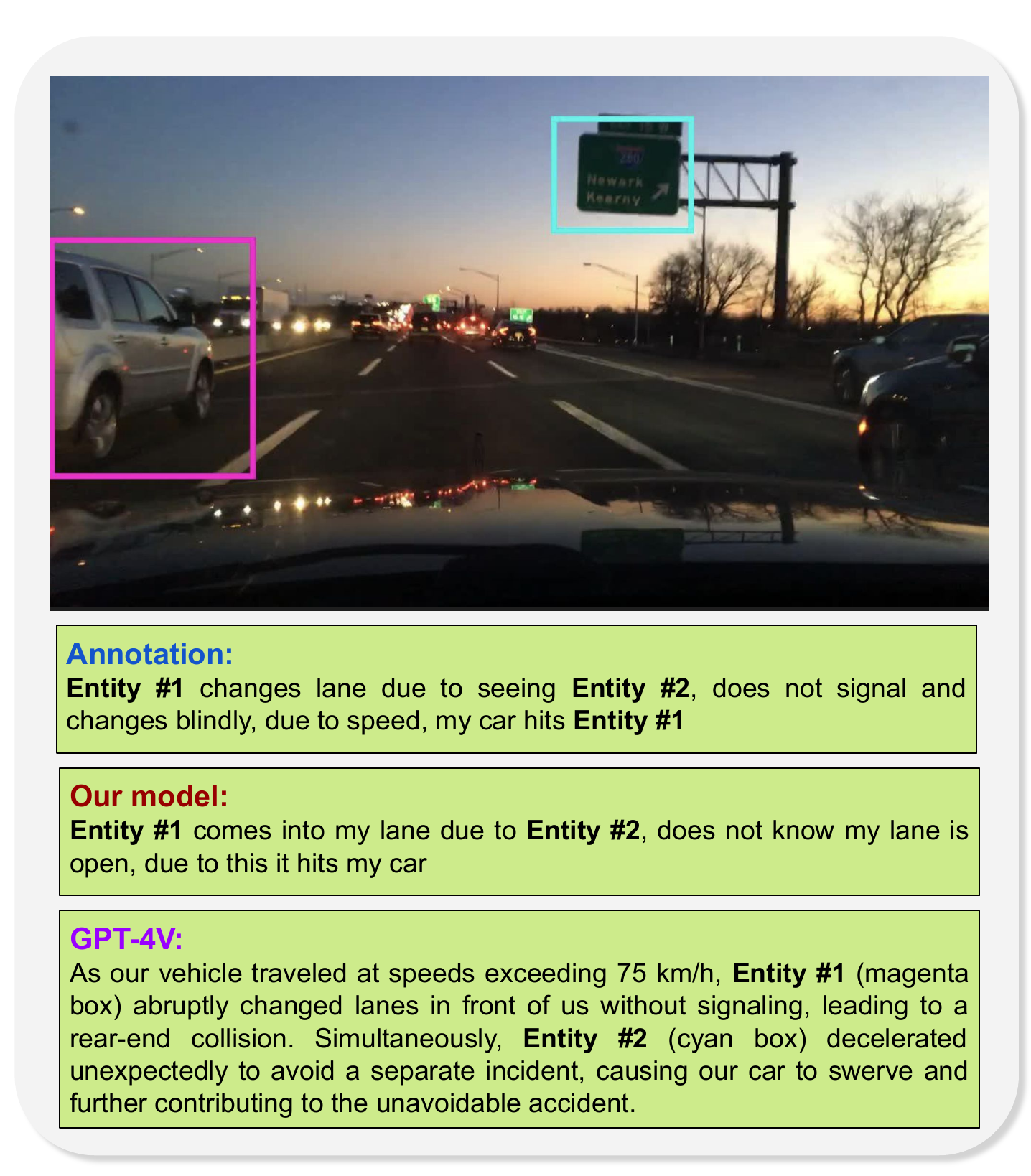}} 
  \subfloat[]
  {\includegraphics[width=0.48\textwidth, keepaspectratio]{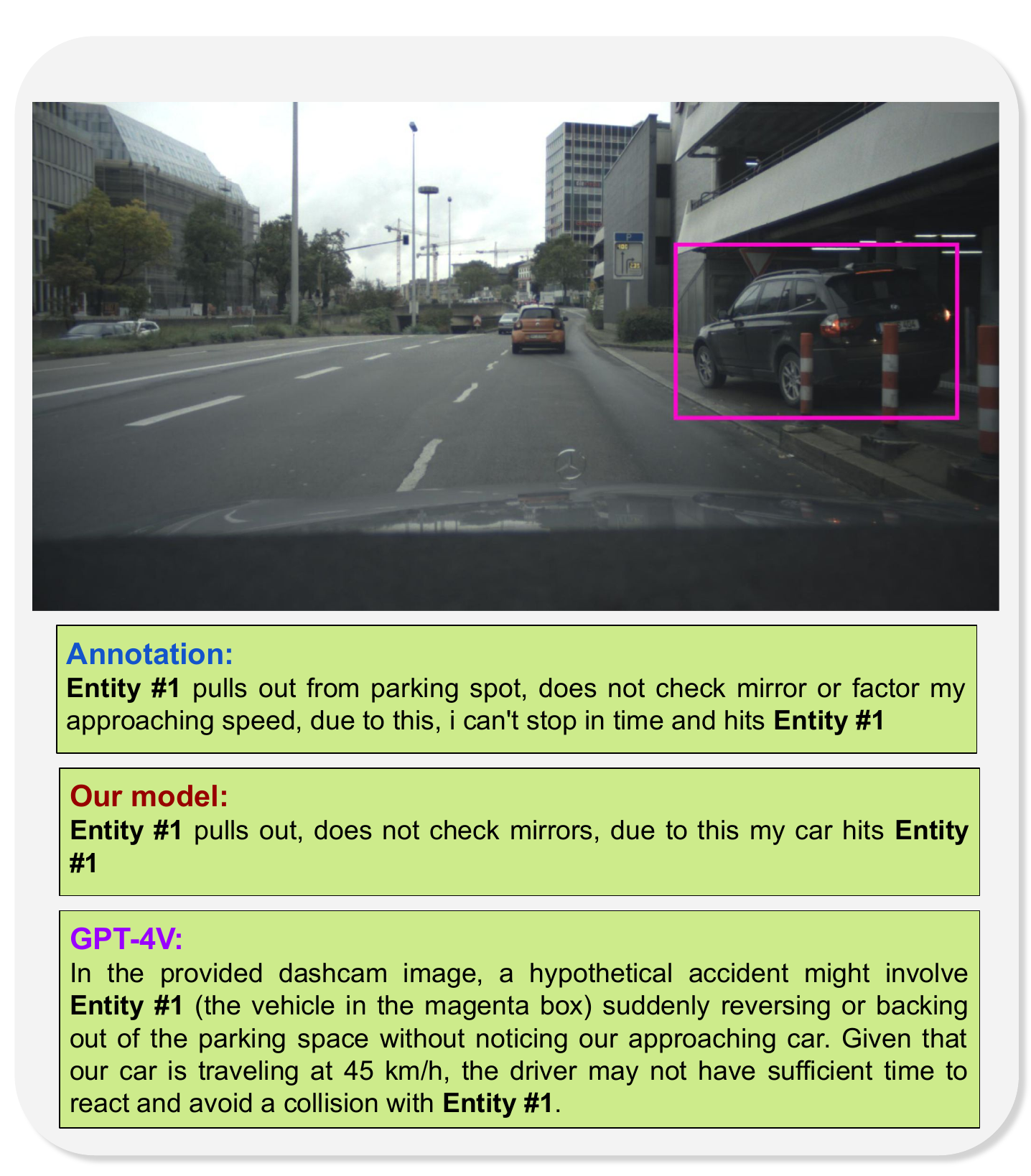}} 
  \caption{Hazard explanation texts generated by our baseline model and GPT-4V: {\em Merging Maneuver}.}
  \label{fig:hazard_merging}
\end{figure*}




\begin{figure*}[t]
  \centering
  \subfloat[]
  {\includegraphics[width=0.48\textwidth, keepaspectratio]{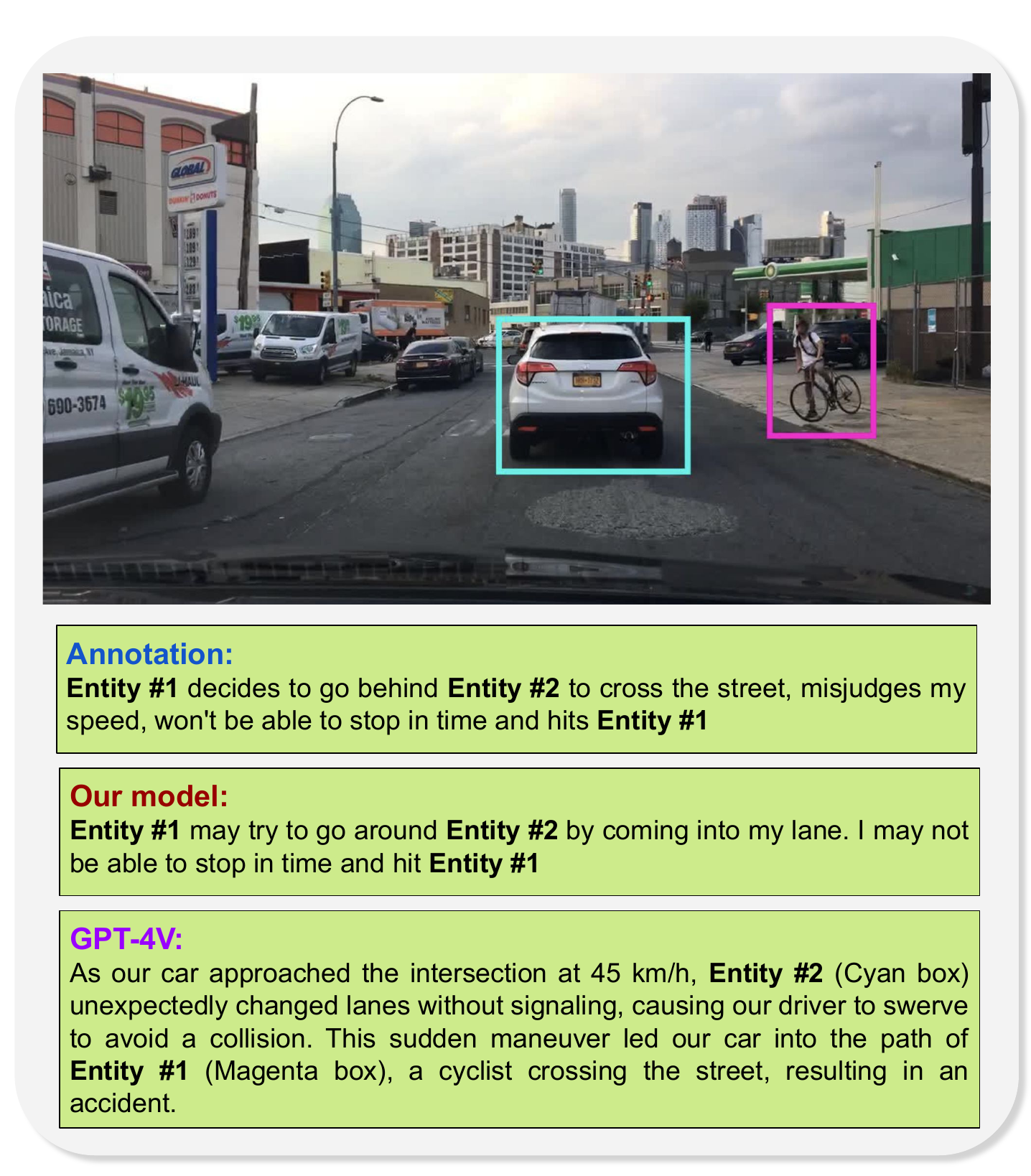}} 
  \subfloat[]
  {\includegraphics[width=0.48\textwidth, keepaspectratio]{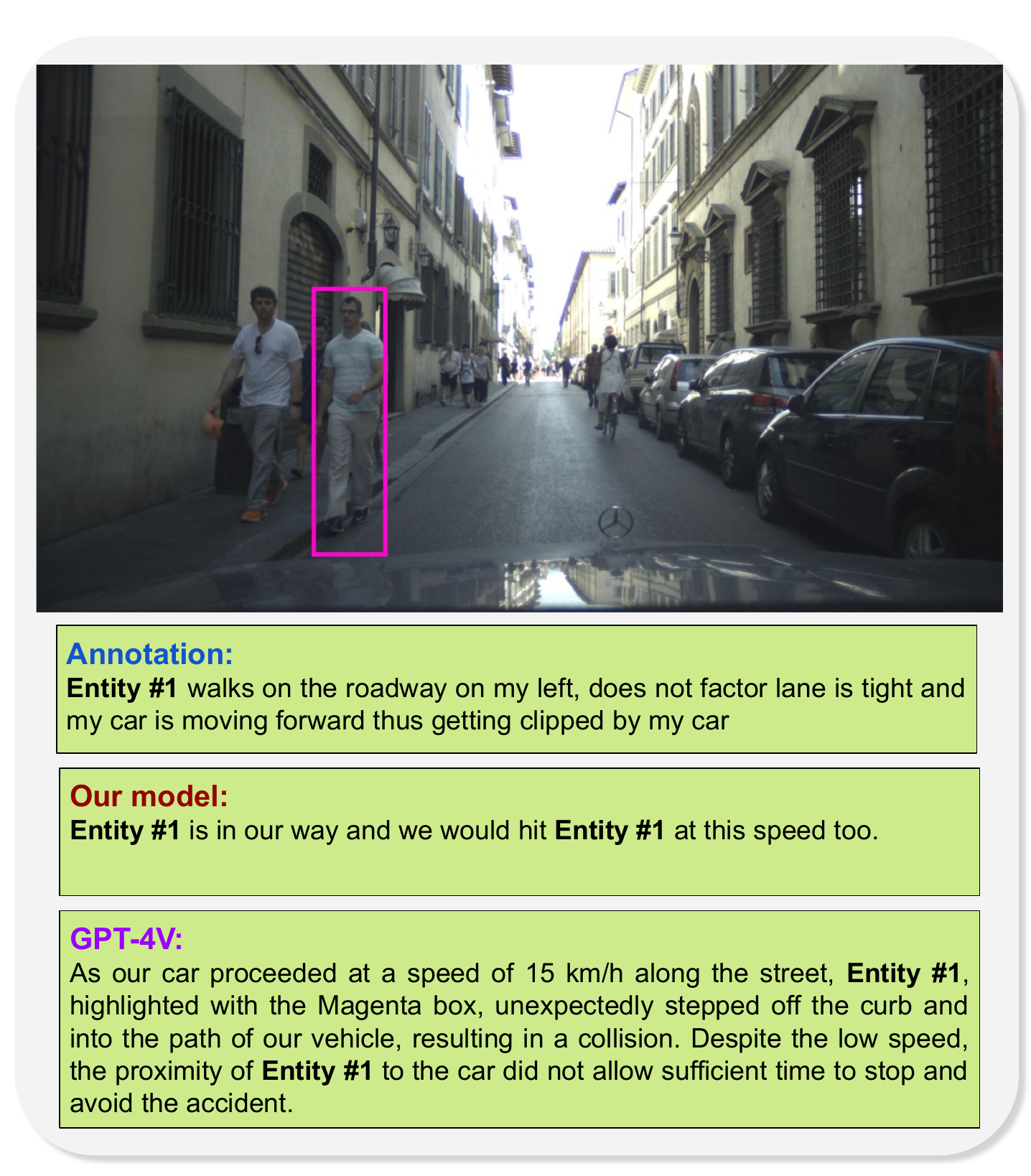}} 
  \caption{Hazard explanation texts generated by our baseline model and GPT-4V: {\em Pedestrian}.}
  \label{fig:hazard_pedestrian1}
\end{figure*}

\begin{figure*}[t]
  \centering
  \subfloat[]
  {\includegraphics[width=0.48\textwidth, keepaspectratio]{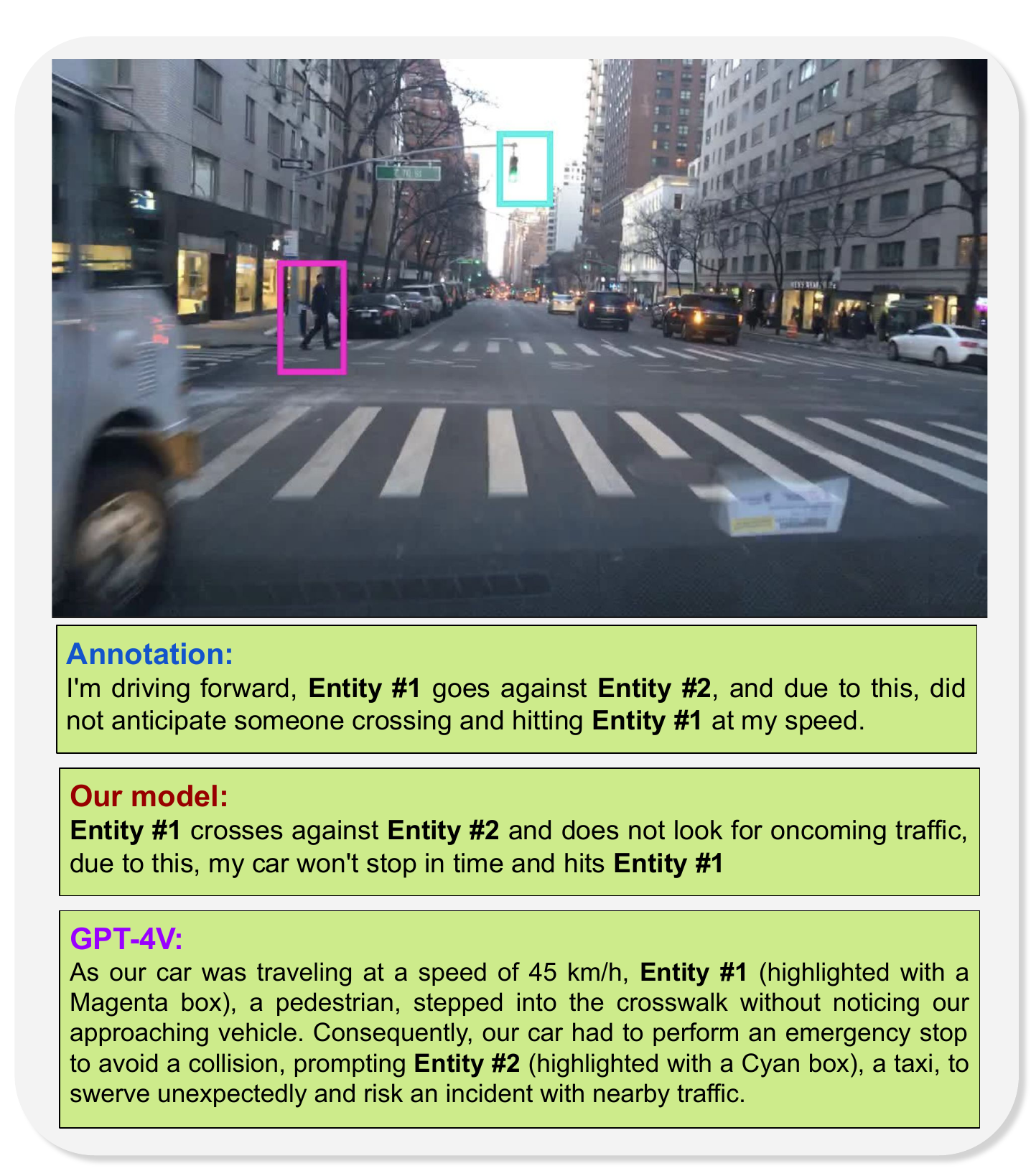}} 
  \subfloat[]
  {\includegraphics[width=0.48\textwidth, keepaspectratio]{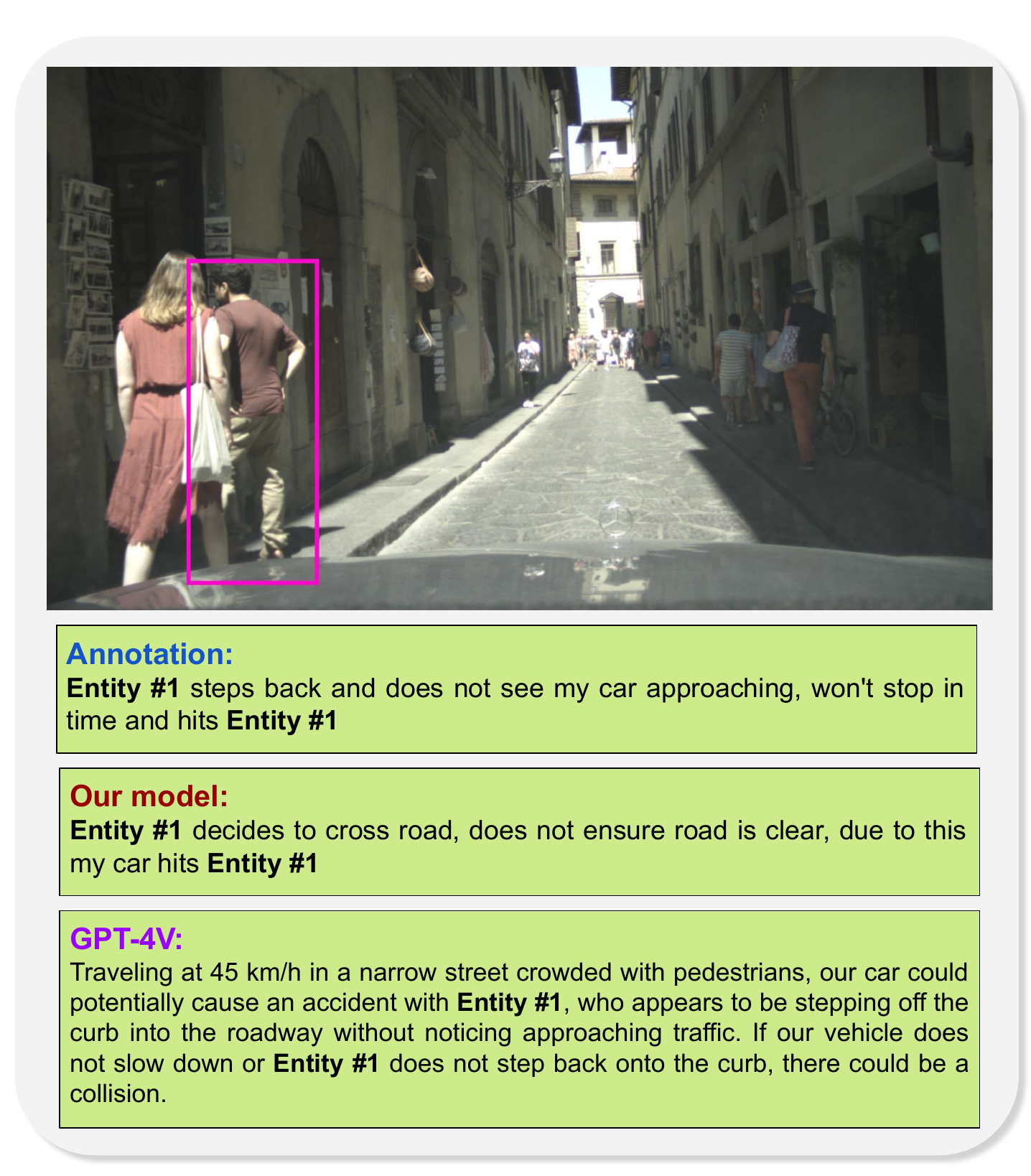}} 
  \caption{Hazard explanation texts generated by our baseline model and GPT-4V: {\em Pedestrian}.}
  \label{fig:hazard_pedestrian2}
\end{figure*}

Figure \ref{fig:hazard_pedestrian2} presents additional examples of the {\em Pedestrian} hazard type. Our model accurately explains the hazard in example (a), but in example (b), it incorrectly predicts the pedestrian's motion direction. It is implausible that a pedestrian not facing the road would walk into the street. In contrast, GPT-4V correctly identifies the hazard in example (b), recognizing that the pedestrian is about to step off the curb, and its explanation aligns with the annotation. However, in example (a), GPT-4V incorrectly identifies a traffic sign as a taxi when recognizing Entity\#$2$.






\begin{figure*}[t]
  \centering
  \subfloat[]
  {\includegraphics[width=0.48\textwidth, keepaspectratio]{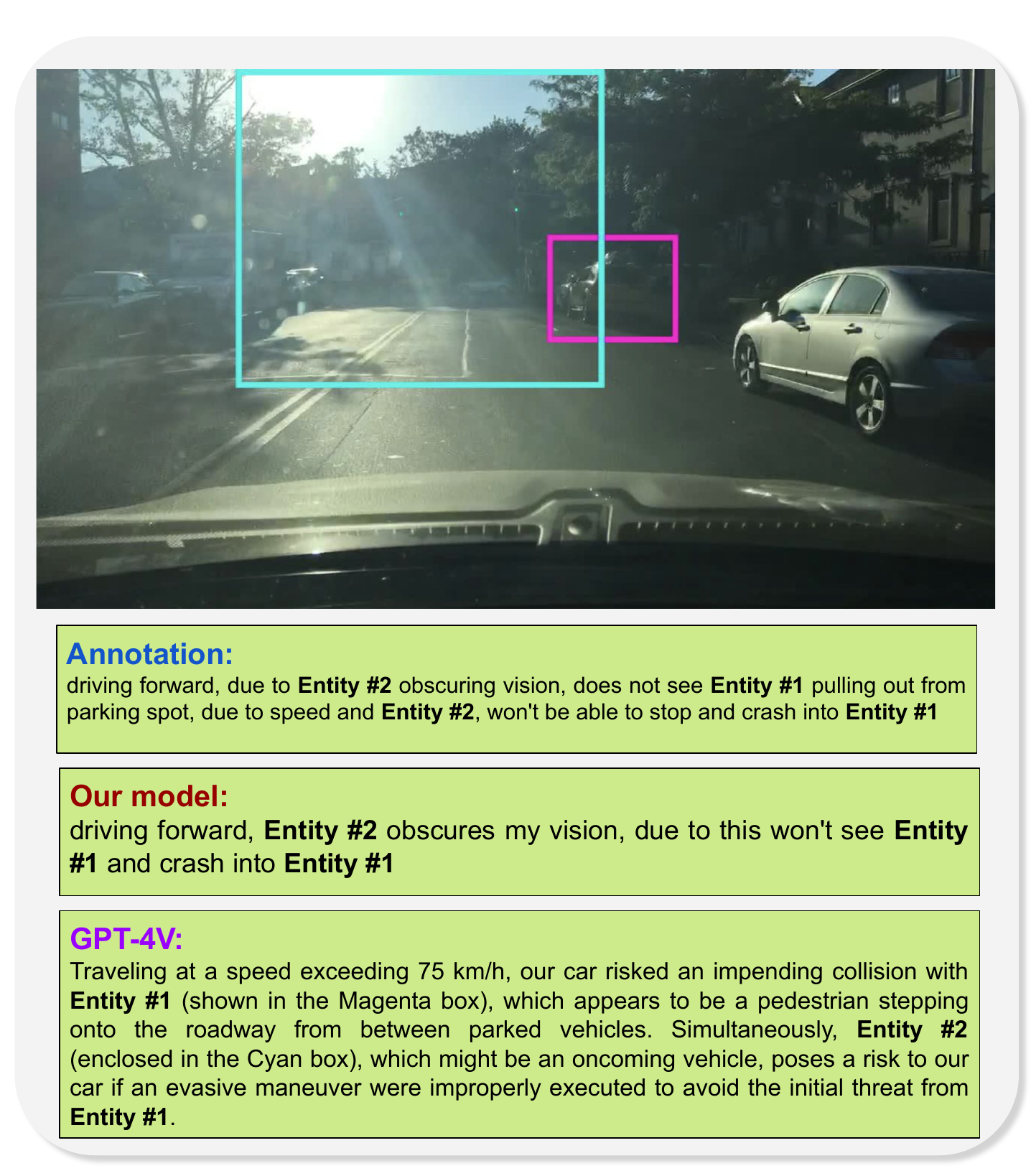}} 
  \subfloat[]
  {\includegraphics[width=0.48\textwidth, keepaspectratio]{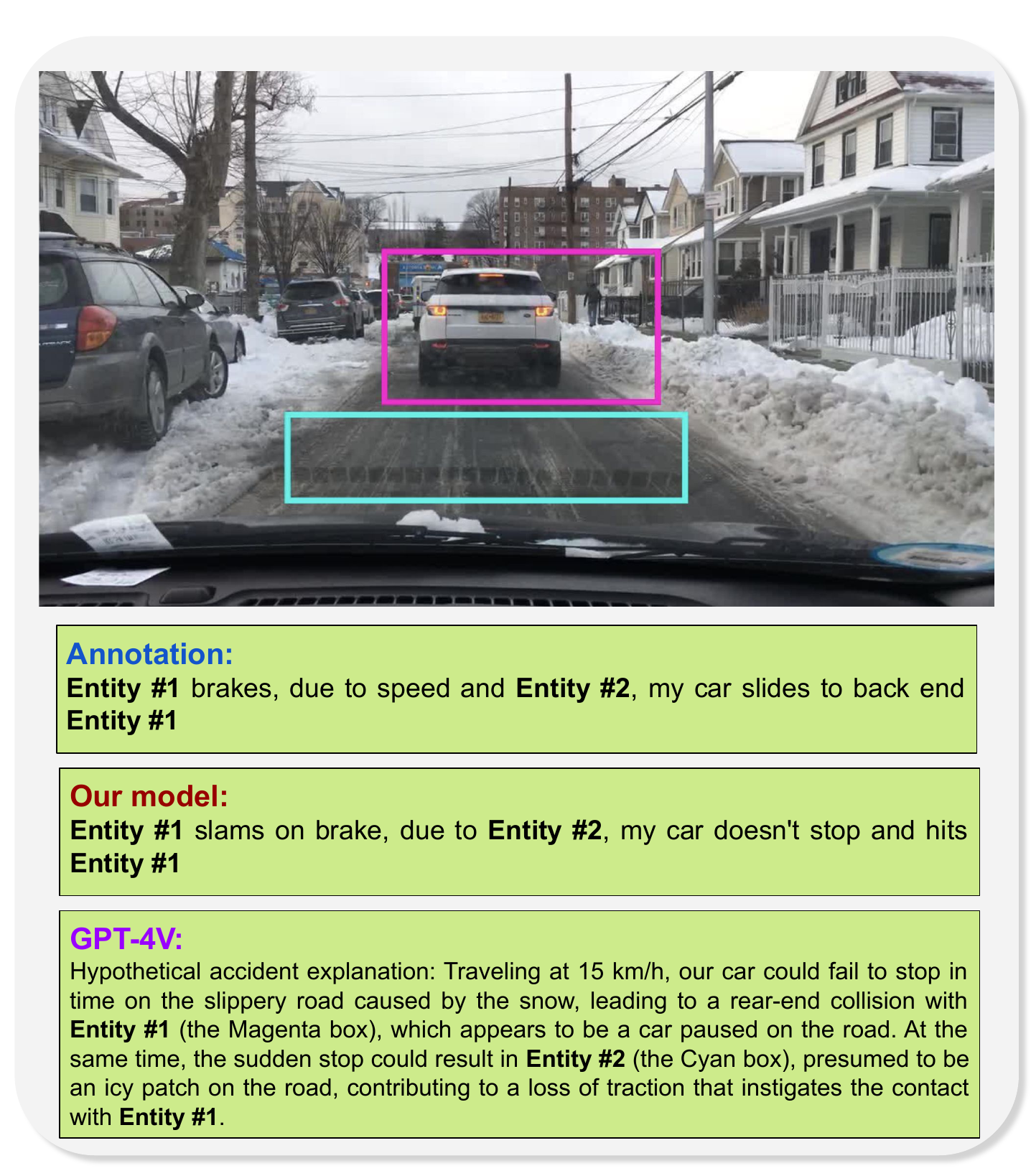}} 
  \caption{Hazard explanation texts generated by our baseline model and GPT-4V: {\em Unusual Condition}.}
  \label{fig:hazard_unusual}
\end{figure*}

In examples involving the {\em Unusual Condition} hazard type shown in Fig.~\ref{fig:hazard_unusual}, our model delivers effective explanations, demonstrating a high degree of semantic similarity to the annotations. GPT-4V also performs well by accurately identifying the relevant entities and offering hazard explanations that closely match the annotations.



\begin{figure*}[t]
  \centering
  \subfloat[]
  {\includegraphics[width=0.48\textwidth, keepaspectratio]{figures/generation_examples/example_12.pdf}} 
  \subfloat[]
  {\includegraphics[width=0.48\textwidth, keepaspectratio]{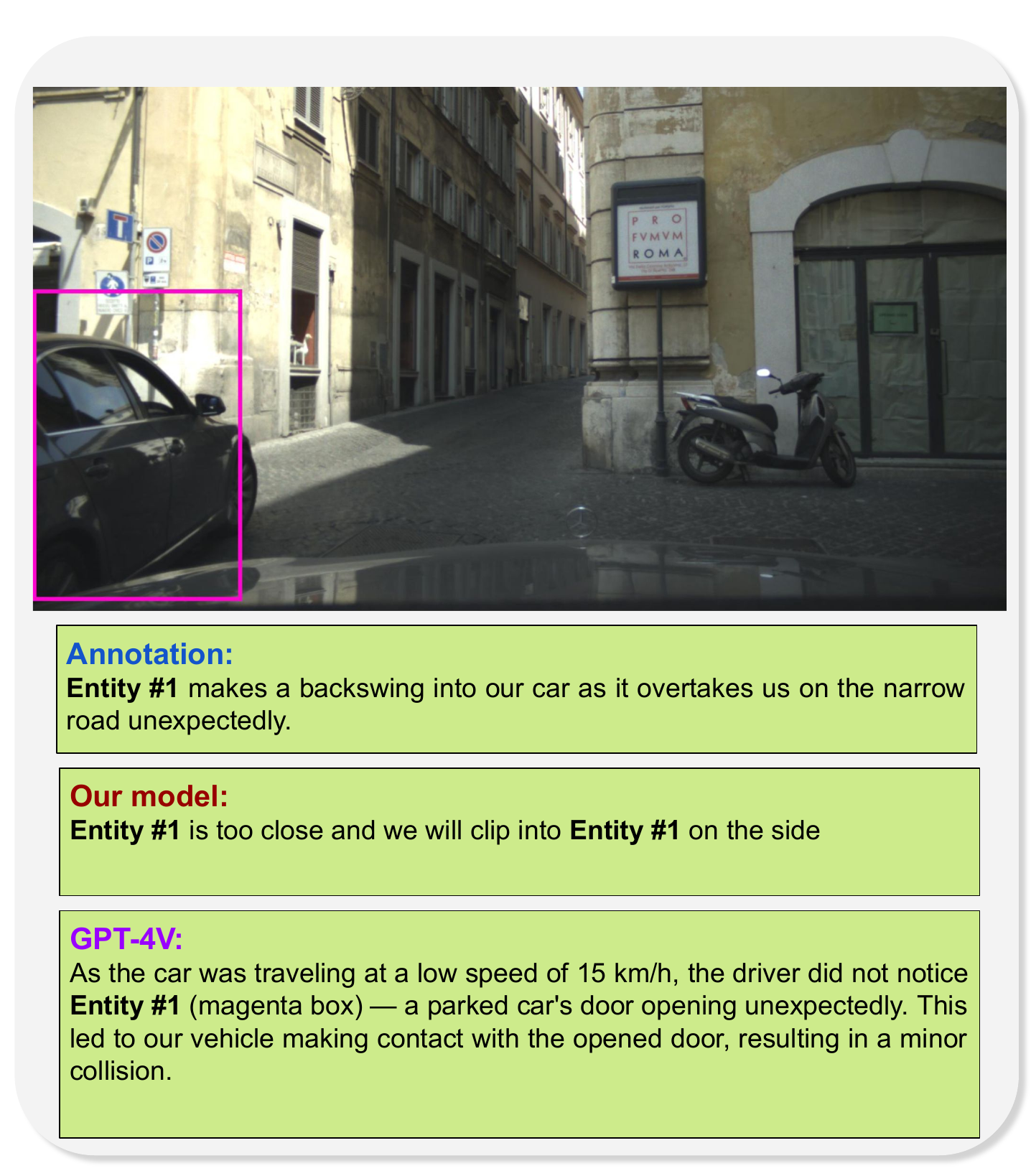}} 
  \caption{Hazard explanation texts generated by our baseline model and GPT-4V: {\em Sideswipe}.}
  \label{fig:hazard_sideswipe}
\end{figure*}

Figure \ref{fig:hazard_sideswipe} presents {\em Sideswipe} hazard type examples. Both models provide reasonable explanations for example (a). 
However, in example (b), both models incorrectly identify the entity as a parked vehicle, highlighting the inherent difficulty in discerning whether vehicles shown in still images are moving or stationary.




\begin{figure*}[t]
  \centering
  \subfloat[]
  {\includegraphics[width=0.48\textwidth, keepaspectratio]{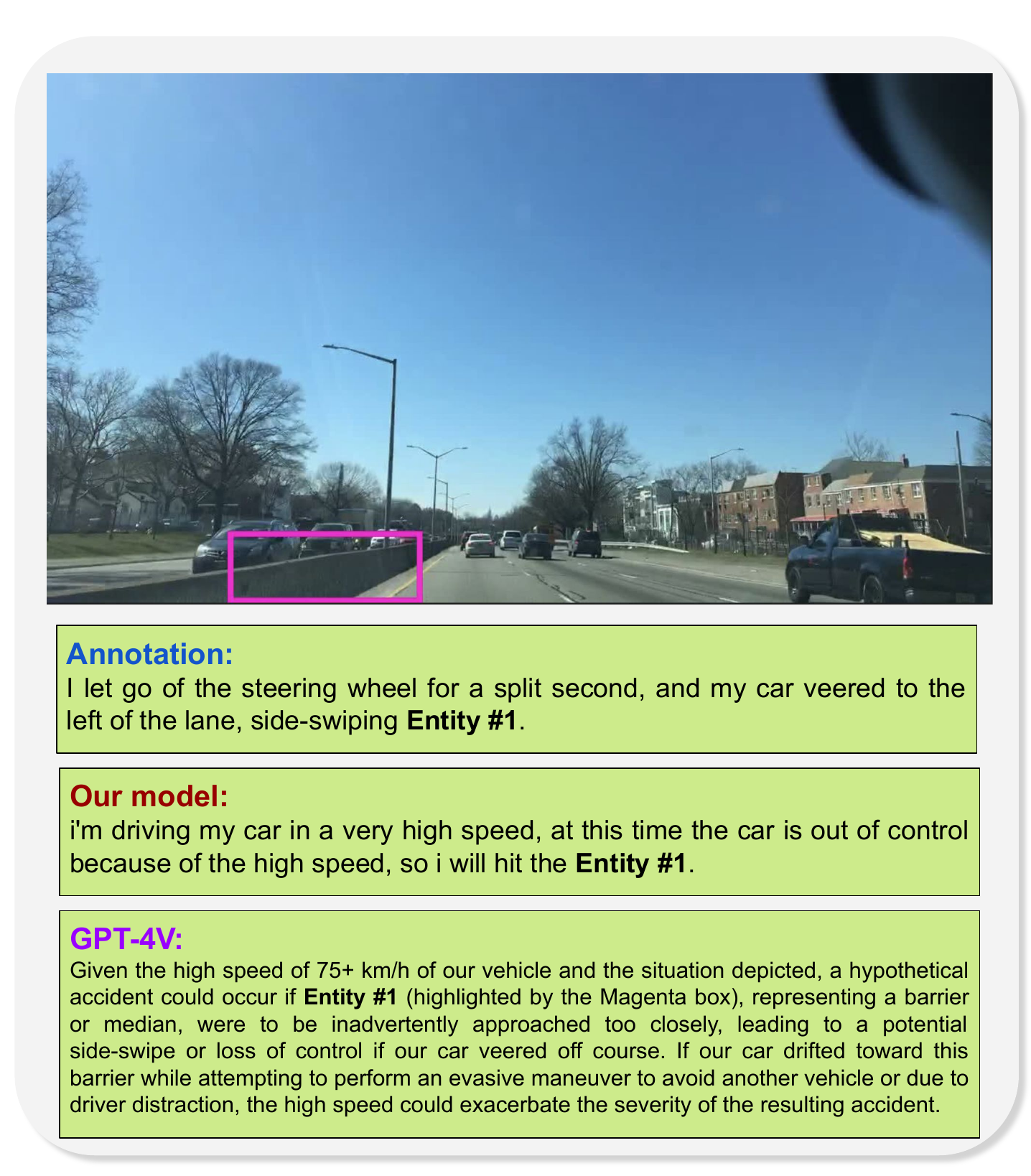}} 
  \subfloat[]
  {\includegraphics[width=0.48\textwidth, keepaspectratio]{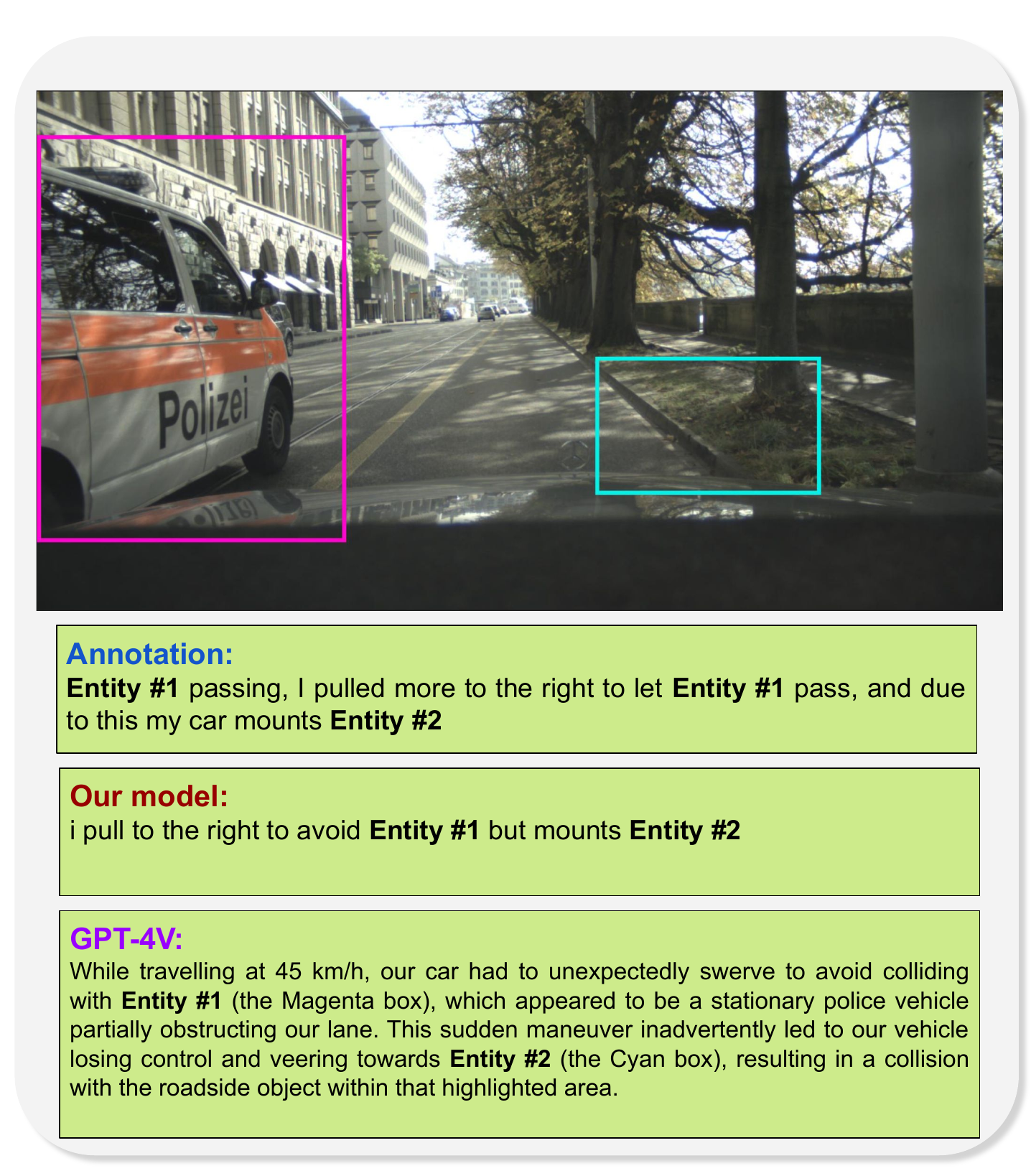}} 
  \caption{Hazard explanation texts generated by our baseline model and GPT-4V: {\em Stationary Object}.}
  \label{fig:hazard_stationary_object}
\end{figure*}

The explanations for the {\em Stationary Object} hazard type examples in Fig.~\ref{fig:hazard_stationary_object}, provided by both models, are logical and align well with the annotations. Remarkably, GPT-4V demonstrates an accurate understanding of the sides of the road. This is surprising, considering that it struggles with more specific entities, such as traffic signs or traffic lights, as demonstrated in previous examples.



\begin{figure*}[t]
  \centering
  \subfloat[]
  {\includegraphics[width=0.48\textwidth, keepaspectratio]{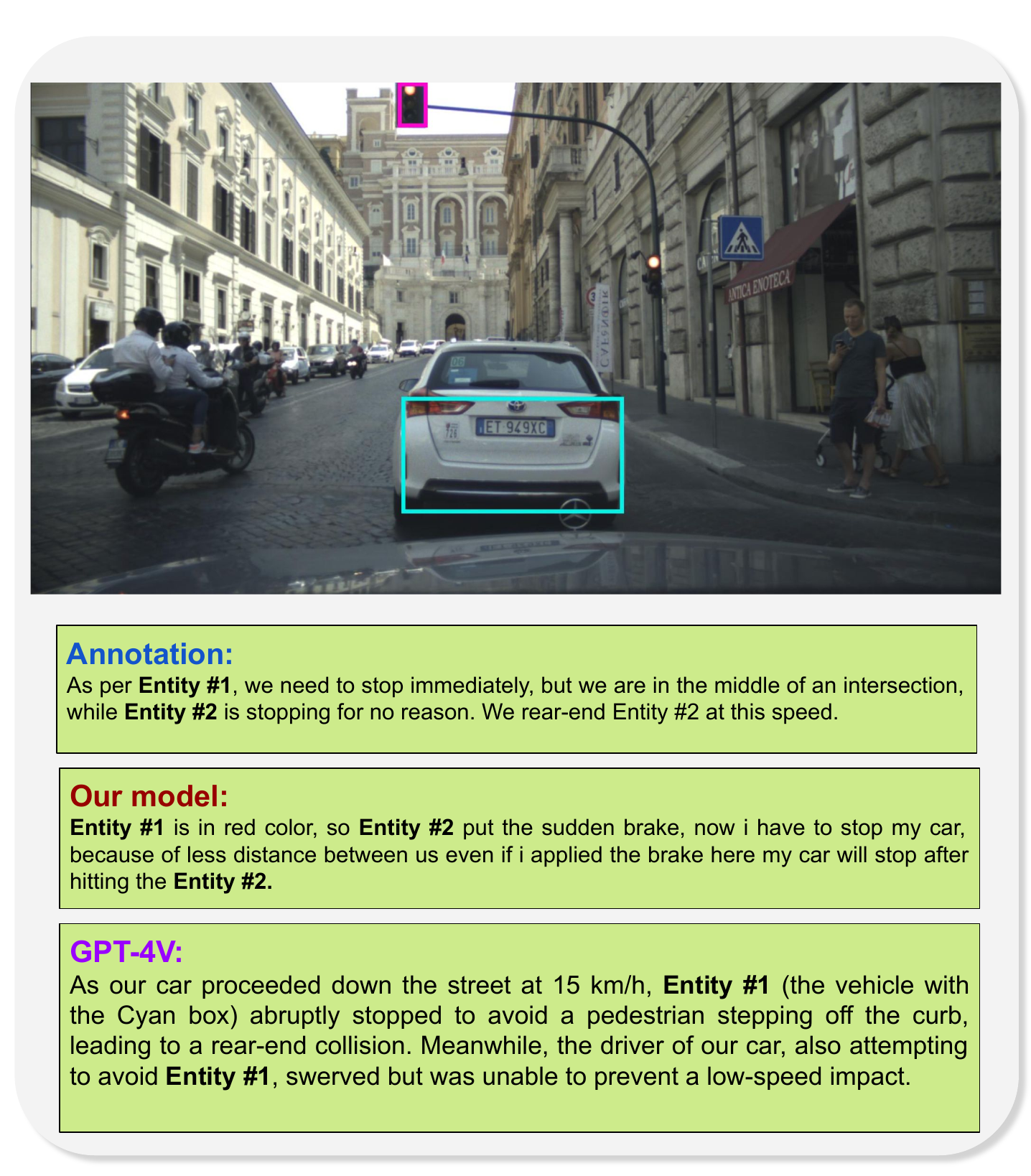}} 
  \subfloat[]
  {\includegraphics[width=0.48\textwidth, keepaspectratio]{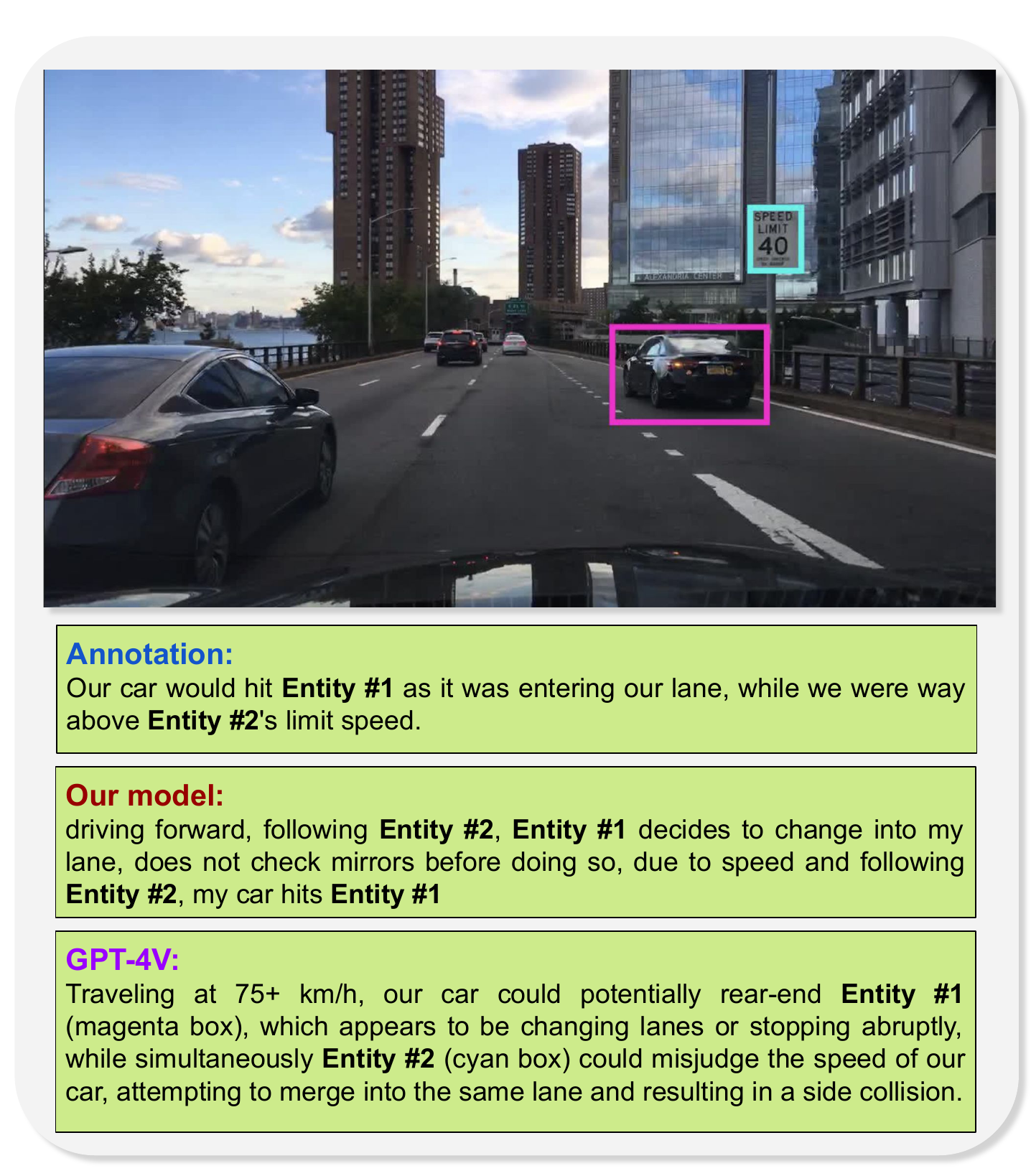}} 
  \caption{Hazard explanation texts generated by our baseline model and GPT-4V: {\em Traffic Signal}.}
  \label{fig:hazard_traffic_signal}
\end{figure*}

Figure \ref{fig:hazard_traffic_signal} illustrates examples of the {\em Traffic Signal} hazard type. Our model successfully formulates hypotheses about accident causes, such as braking or lane changing. However, in example (b), the logic appears flawed; the connection between the speed limit sign (Entity \#$2$) and the actions of another vehicle (Entity \#$1$) is not well-defined. Conversely, GPT-4V does not accurately identify the traffic signals and signs in these examples, leading to implausible hazard explanations.

\begin{figure*}[t]
  \centering
  \subfloat[]
  {\includegraphics[width=0.48\textwidth, keepaspectratio]{figures/generation_examples/example_18.pdf}} 
  \subfloat[]
  {\includegraphics[width=0.48\textwidth, keepaspectratio]{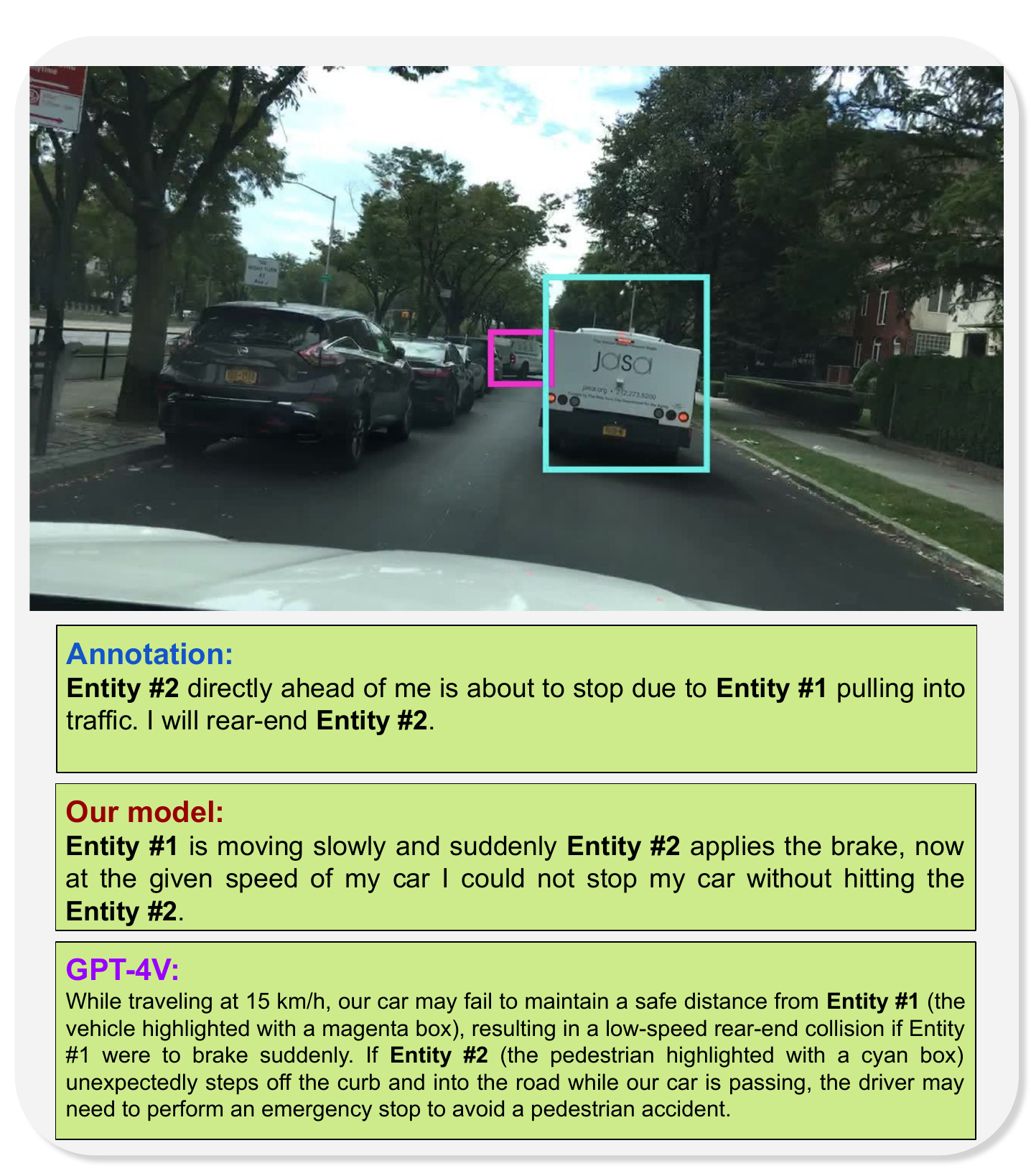}} 
  \caption{Hazard explanation texts generated by our baseline model and GPT-4V: {\em Unexpected Event \& Chain Reaction}.}
  \label{fig:hazard_complex}
\end{figure*}

Figure \ref{fig:hazard_complex} illustrates examples of {\em Unexpected Events} and {\em Chain Reaction} hazard types, where neither model offers fully accurate reasoning. Both models present explanations that do not correspond well with the given annotations. Although our model's explanation seems logical, it fails to clearly state the causal relationship between the events involved. Specifically, in example (b), our model does not clarify that Entity \#$1$ leads to the actions of Entity \#$2$, unlike the annotated explanation where the first entity causes the second to brake suddenly. Additionally, in example (b), GPT-4V incorrectly identifies a white truck as a pedestrian.


The behavior of our model and GPT-4V, as observed in the examples, can be summarized as follows. Our model, which is fine-tuned, excels in following specific instruction formats and produces consistent, patterned outputs, but it tends to oversimplify explanations of hazards. In contrast, GPT-4V is proficient in a descriptive-first, reasoning-based approach to inference. However, it often struggles with accurate entity recognition and full contextual understanding of images, leading to errors in hazard prediction. Notably, GPT-4V frequently fails to recognize traffic signs and signals. Both models exhibit limitations in spatial and temporal recognition of traffic scenes. These observations suggest substantial room and necessity for improvement, implying several directions for future research.

\newpage
\section{License} 
\label{subsec:license}

The image assets from the BDD100K dataset are distributed under the BSD 3-Clause License, while the ECP dataset is governed by the eurocity persons dataset research use license. Our usage of both datasets complies with their respective licenses, and we employ anonymization techniques, such as blurring identifiable faces and license plates, to adhere to regulations governing personal data processing.

The DHPR dataset created in this study is licensed under the Creative Commons Attribution-NonCommercial 4.0 International (CC BY-NC 4.0) license. This license allows others to use, adapt, and distribute the dataset, provided they give appropriate credit to the original creator and do not use the dataset for commercial purposes. 

\begin{figure*}[t]
\includegraphics[width=1.0\textwidth]{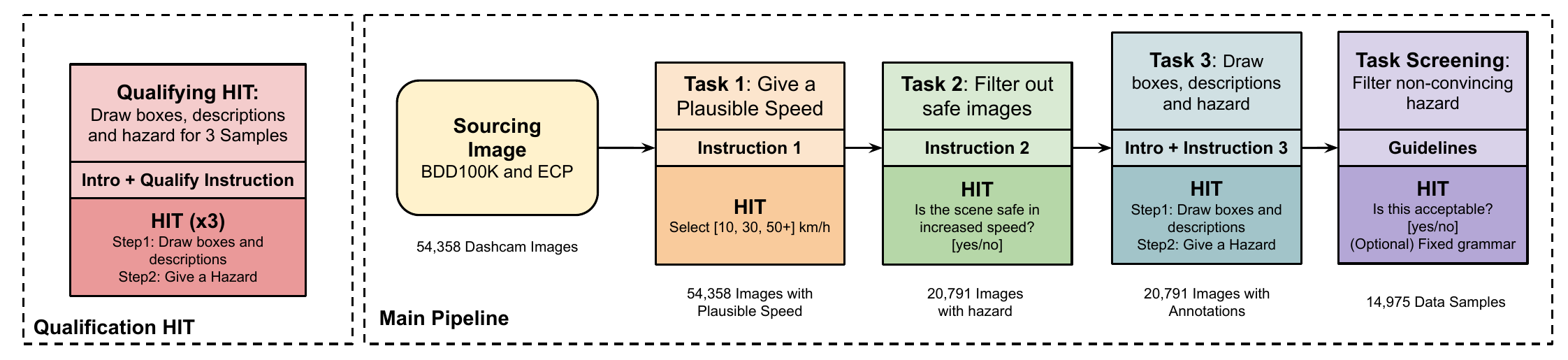}
\caption{Overview of the data collection process}
\label{fig:data-collection-pipeline}
\end{figure*}

\section{Data Collection Process}

\subsection{Overview}

We used Amazon Mechanical Turk to generate the dataset. Figure \ref{fig:data-collection-pipeline} provides an overview of the data collection process. The process involves two steps: the preliminary step for qualifying workers and the main pipeline for creating annotations. Only the workers filtered in this qualification step were invited to participate in the main pipeline (Sec.~\ref{subsec:qualifying-task}). 

The main pipeline consists of three tasks:
\begin{itemize}
    \item {\bf Task 1}: The car's speed is estimated for each input scene image (Sec.~\ref{subsec:task1}); 
    \item {\bf Task 2}: The possibility of an accident occuring is estimated so that risk-free scene images are removed (Sec.~\ref{subsec:task2}); 
    \item {\bf Task 3}: A driving hazard is hypothesized and annotated for an input image (Sec.~\ref{subsec:task3}). 
\end{itemize}

We employ these three steps for the following reasons. In Task 3, workers are asked to annotate a hypothesized hazard for each input image. Since the original images are of accident-free scenes, it can often be challenging to identify hazards, even as hypotheses. If workers are given the freedom to choose whether or not to annotate an image based on its difficulty, we will struggle to collect a sufficient amount of annotations. However, it is also problematic to force workers to identify hazards in risk-free scenes. Therefore, we designed Task 2 to address this issue. In Task 2, we remove scene images with minimal risk and transfer the remaining images to Task 3. To avoid removing too many images, we ask workers to assess the risk of the scene images by assuming a 50\% increase in the car's speed. We designed Task 1 to estimate the normal speed of the car. 

For each task in the qualification step and the main pipeline, we created HITs (Human Intelligence Tasks). All communication was conducted in English. The dataset was collected and evaluated from January through March 2023. Throughout the process, we paid the workers around \$10 USD/hour. To be specific, we paid \$0.02 USD per HIT for multiple choice HITs such as Task 1 and Task 2. For Task 3, we paid \$0.2 USD per HIT, and some workers may finish this within 40 seconds.

We input 54,358 images from two datasets, BDD100K and ECP, to the main pipeline, followed by an additional qualification step (Sec.~\ref{subsec:qualifying-task}). Consequently, we acquired the annotations for a total of 14,975 images, which comprise the final DHPR dataset.

\subsection{Preliminary Step: Qualification/Screening of Workers}
\label{subsec:qualifying-task}

As mentioned above, we utilized a qualifying test to identify competent workers. This test not only serves as an evaluation tool but also provides potential workers with an overview of the tasks discussed earlier. On the initial page, we present essential information in the form of clickable/expandable items within a menu. This includes a description of the qualifying task (Fig.\ref{fig:instruction-rules}), instructions on annotating visual entities (Fig.\ref{fig:mode-selection}), guidelines for writing effective hazard explanations (Fig.\ref{fig:material-outline-examples0}), and examples of exemplary annotations (Fig.\ref{fig:material-outline-examples}).

The qualifying task is designed to mimic Task 3. Its purpose serves three objectives: (i) to assess the workers' understanding of the instructions, (ii) to evaluate their experience and comprehension of driving cars and traffic conditions, and (iii) to gauge their proficiency in providing annotations in the required format. Each worker was administered three questions as part of this task. Examples are presented in Fig.~\ref{fig:qualify-worker-combined}.

To ensure a diverse range of annotations, we invited over 500 workers worldwide to participate in the qualification test. In order to maintain quality control, we specifically targeted workers with a proven track record of approving more than 10,000 HITs and maintaining an approval rate of over 95\%. Following the evaluation of their performance on the test, we manually selected 60 workers who met our criteria. These selected workers were then invited to participate in the main annotation pipeline.

\begin{figure*}[t]
  \centering
\hbox{\includegraphics[width=0.5\textwidth]{HIT_images/qualitfy_task1.pdf} \hfill
\includegraphics[width=0.5\textwidth]{HIT_images/qualify_task2.pdf}}
  \caption{Examples of the qualfifying test}
  \label{fig:qualify-worker-combined}
\end{figure*}

\subsection{Task 1: Estimating the Car's Speed}
\label{subsec:task1}

The workers were instructed to estimate the speed of the car based on the dashcam image of a scene. The web interface for the HIT of the task is shown in Fig.~\ref{fig:plausible_speed}. A total of 54,358 scene images were used, with each assigned to a single worker. As a result of this task, we obtained annotations for all 54,358 images.

\subsection{Task 2: Predicting Accident Possibility to Filter Images}
\label{subsec:task2}

The second task aims to assess the probability of a car being involved in an accident within a few seconds. In order to make Task 3, annotating hypothesized hazards, efficient, it was necessary to eliminate scene images with a very low likelihood of accidents. To achieve this, we introduced an increased speed that was 1.5 times faster than the annotated speed used in the first task. The intention behind this speed increase was to instill a stronger sense of the potential for an accident among the workers, given that the original speed was determined based on their perception of a safe speed. Figure \ref{fig:determine_accident} illustrates the instruction and annotation form provided for this task. Consequently, out of the initial set of 54,358 images, 20,791 images were filtered and subsequently used in the following task.

\subsection{Task 3: Hypothesizing and Annotating a Hazard}
\label{subsec:task3}

The final task involves hypothesizing and annotating a hazard for each of the previously filtered images. This task comprises two parts. The first part involves annotating the visual entities associated with the hazard. Workers are instructed to draw a bounding box around each entity and provide a brief description of it. The second part requires providing a natural language explanation of the hazard, using the term `Entity \#$n$' to refer to the involved entities. Figure \ref{fig:rationale} shows the instruction and annotation forms for this task.

Task 3 is the most time-consuming, accounting for 80\% of the total annotation time. Based on our statistics, each worker may spend up to three minutes per image and can complete a maximum of three hundred HITs per day. In order to enhance productivity and reduce inconsistencies in answers, we have implemented data input validation and a user interface assistant\footnote{
Our data input validation system ensures that submissions meet the following criteria:
at least one box must be drawn; each box should have a corresponding entity description, and vice versa; only one bounding box per entity is allowed; When adding a box, a new entity must be utilized; and the hazard explanation must be at least five words long. If any of these criteria are not met in a submission, a warning prompt will be displayed. 

In addition, the ``Word Assistant Pads'' feature was provided to the workers to minimize the need for typing, as shown in Fig.~\ref{fig:rationale}(b) and (c). It automatically fills in the text prompt input form by clicking buttons. This aid also serves as a reminder to workers regarding the expected content of the input form. Additionally, a brief guideline emphasizing the necessary components of the sentence was provided, including an accident-related entity, its relative position, and the resulting accident. Also, in close proximity to the hazard input form, there are reminders for the specific entities (`Entity \#$1$', `Entity \#$2$', `Entity \#$3$'), as well as preposition words, to discourage the input of noun words for the entity.
}.

\subsection{Post Process: Data Validation}
\label{subsec:datavalidation}

We checked the results during and after Task 3 ourselves, and removed annotations with obvious errors. Additionally, we invited a small number of the most reliable workers to an additional task of eliminating Task 3 annotations that had unsatisfactory quality. The web page design for the HIT is shown in Fig.~\ref{fig:data-validation}. The workers were presented with annotations for each scene image, including bounding boxes, descriptions of visual entities, and hazard explanations. They were asked a binary (yes/no) question regarding the acceptability of the annotations. If necessary, the workers were also requested to correct minor mistakes such as grammatical errors or incorrect word choices. Following this screening step, we obtained annotations for a total of 14,975 scene images.

\begin{figure*}[]
\centering
\includegraphics[width=0.85\textwidth]{HIT_images/qualifying_hit_introduction.pdf}
\caption{Instructions for the qualifying test, which also serves as an introduction to the real tasks in the main annotation pipeline.}

\label{fig:instruction-rules}
\end{figure*}

\begin{figure*}[]
  \centering
  \subfloat[Step 1: Select an entity index]
  {\includegraphics[width=0.85\textwidth]{HIT_images/task3_howto1.pdf}} 
 
  \vspace{0.5cm}
  \subfloat[Step 2: Draw a box and input a brief description]
  {\includegraphics[width=0.85\textwidth]{HIT_images/task3_howto2.pdf}} 
  \caption{Explanation of how to annotate visual entities by using the tools provided. }

  \label{fig:mode-selection}
\end{figure*}

\begin{figure*}[]
\centering
\includegraphics[width=0.8\textwidth]{HIT_images/outline1.pdf}
\caption{Guidelines for writing effective hazard explanations} 
\label{fig:material-outline-examples0}
\end{figure*}

\begin{figure*}[]
\centering
\includegraphics[width=0.8\textwidth]{HIT_images/example1.pdf}
\caption{Examples of exemplary annotations}
\label{fig:material-outline-examples}
\end{figure*}

\begin{figure*}[]
  \centering
  \subfloat[Instructions]
  {\includegraphics[width=\textwidth]{HIT_images/task1_instruction.pdf}} 
  \hfill
  
  \vspace{1cm}
  \subfloat[Annotation form]
  {\includegraphics[width=\textwidth]{HIT_images/task1_annotation_form.pdf}} 
  \caption{(a) Instruction and (b) the anotation form for Task 1, which requests the workers to estimate the car's speed.}
  \label{fig:plausible_speed}
\end{figure*}

\begin{figure*}[]
  \centering
  \subfloat[Instruction]
  {\includegraphics[width=\textwidth]{HIT_images/task2_instruction.pdf}} 
  
  \vspace{0.5cm}
  
  \subfloat[Annotation form]
  {\includegraphics[width=\textwidth]{HIT_images/task2_annotation_form.pdf}} 
  \caption{(a) Instruction and (b) annotation form for Task 2, which is to predict the possibility of an accident for an input image.}
  \label{fig:determine_accident}
\end{figure*}

\begin{figure*}[]
  \centering
  \subfloat[Instruction]
  {\includegraphics[width=\textwidth]{HIT_images/task3_instruction.pdf}} 
  \vspace{0.5cm}
  
  \subfloat[Annotation form for visual entities]
  {\includegraphics[width=\textwidth]{HIT_images/task3_input_form_step1.pdf}} 

  \vspace{0.5cm}

  \subfloat[Annotation form for hazard explanations]
  {\includegraphics[width=\textwidth]{HIT_images/task3_input_form_step2.pdf}} 
  \caption{(a) Instruction, (b) \& (c) annotation forms for Task 3, which is to annotate visual entities involved in a hypothesized hazard and provide an explanation of the hazard.}
  \label{fig:rationale}
\end{figure*}

\clearpage
\begin{figure*}[]
\centering
\includegraphics[width=1\textwidth]{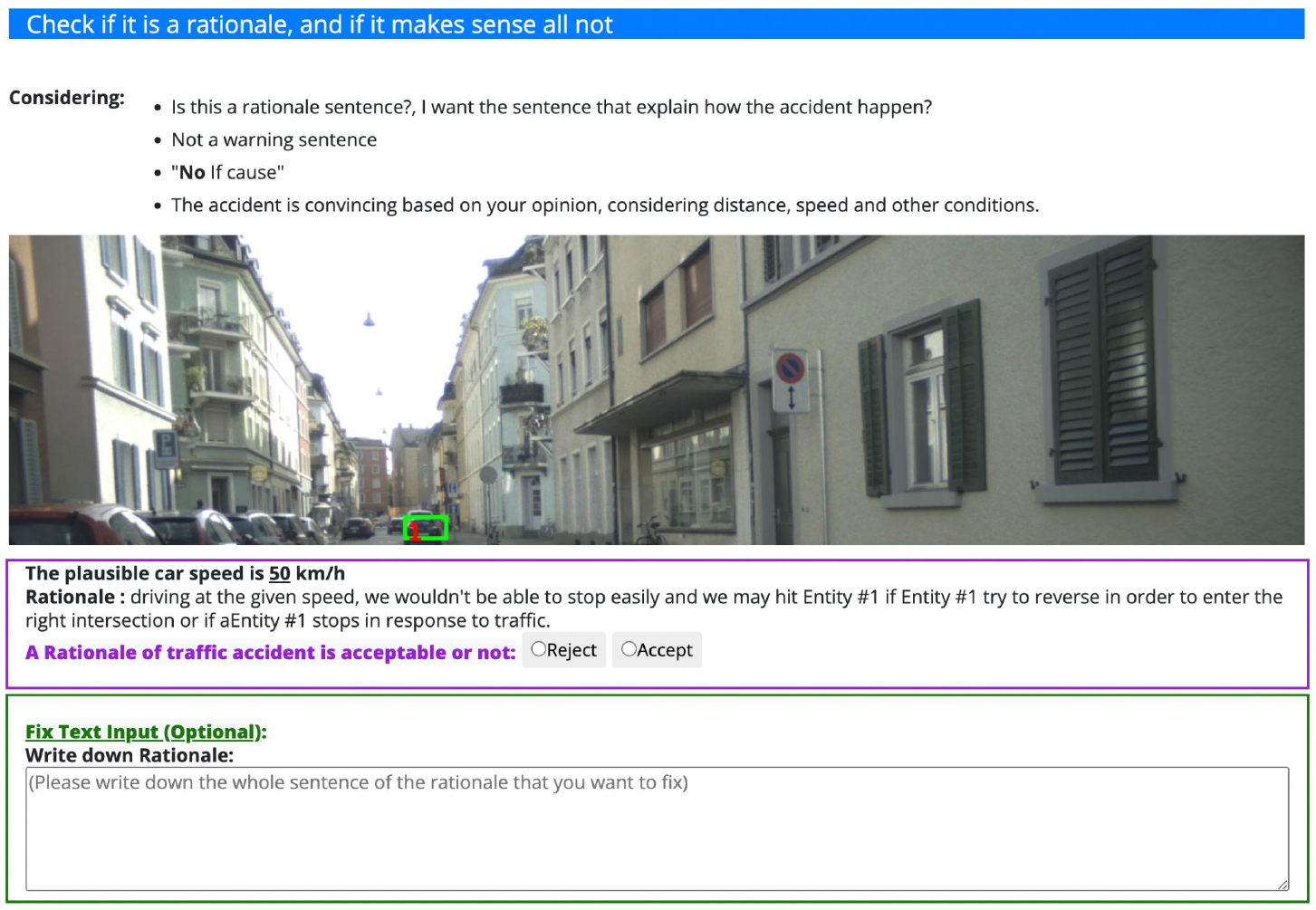}
\caption{Annotation form for the data validation task}
\label{fig:data-validation}
\end{figure*}

\clearpage


\hfill \break